\documentclass{article}

\PassOptionsToPackage{numbers, compress}{natbib}

\usepackage[preprint]{neurips_2025}

\usepackage[utf8]{inputenc} 
\usepackage[T1]{fontenc}    
\usepackage{hyperref}       
\usepackage{url}            
\usepackage{booktabs}       
\usepackage{amsfonts}       
\usepackage{nicefrac}       
\usepackage{microtype}      
\usepackage{amsmath}
\usepackage{graphicx}
\usepackage{caption}  
\usepackage{enumitem}
\usepackage{makecell}
\usepackage{graphicx}
\usepackage{subfigure}
\usepackage{float}
\usepackage{amsmath}
\usepackage{multirow}
\usepackage{colortbl}
\usepackage[dvipsnames]{xcolor} 
\usepackage{fontawesome}

\usepackage{algorithm}
\usepackage{algpseudocode} 
\usepackage{overpic}
\usepackage[utf8]{inputenc} 
\usepackage[T1]{fontenc}    
\usepackage{hyperref}       
\usepackage{url}            
\usepackage{booktabs}       
\usepackage{amsfonts}       
\usepackage{nicefrac}       
\usepackage{microtype}      
\usepackage{amsmath}
\usepackage{amssymb}
\usepackage{enumitem}
\usepackage{makecell}
\usepackage{tikz}  
\usetikzlibrary{shapes.geometric, arrows.meta, calc} 
\usepackage{subcaption}
\usepackage{multirow}
\usepackage{rotating}
\usepackage{comment}
\usepackage{wrapfig}
\usepackage{makecell}

\usepackage{tcolorbox}
\tcbuselibrary{skins}
\tcbuselibrary{breakable} 
\usepackage{lipsum} 
\newcommand{\prompts}[1]{\textit{#1}}

\title{Vision–Language Models as Differentiable Semantic and Spatial Rewards for Text-to-3D Generation}

\author{Weimin Bai$^1$ \quad Yubo Li$^1$ \quad Weijian Luo$^2$ \quad Wenzheng Chen$^1$ \quad He Sun$^1$\textsuperscript{\dag} \\[2pt]
1. Peking University \quad
2. Xiaohongshu Inc \\ [2pt]
\texttt{weiminbai@stu.pku.edu.cn; hesun@pku.edu.cn} \\[4pt]  
    \vspace{0.05em}  
    Project page: \url{https://ai4scientificimaging.org/vlm3d} 
}

\begin{document}

\begingroup
\renewcommand\thefootnote{\dag}  
\footnotetext{Correspondence to \url{hesun@pku.edu.cn}.}
\endgroup

\maketitle

\begin{abstract}

Score Distillation Sampling (SDS) enables high-quality text-to-3D generation by supervising 3D models through the denoising of multi-view 2D renderings, using a pretrained text-to-image diffusion model to align with the input prompt and ensure 3D consistency.
However, existing SDS-based methods face two fundamental limitations: (1) their reliance on CLIP-style text encoders leads to coarse semantic alignment and struggles with fine-grained prompts; and (2) 2D diffusion priors lack explicit 3D spatial constraints, resulting in geometric inconsistencies and inaccurate object relationships in multi-object scenes.
To address these challenges, we propose VLM3D, a novel text-to-3D generation framework that integrates large vision–language models (VLMs) 
into the SDS pipeline as differentiable semantic and spatial priors. Unlike standard text-to-image diffusion priors, VLMs leverage rich language-grounded supervision that enables fine-grained prompt alignment. Moreover, their inherent vision language modeling provides strong spatial understanding, which significantly enhances 3D consistency for single-object generation and improves relational reasoning in multi-object scenes. We instantiate VLM3D based on the open-source Qwen2.5-VL model and evaluate it on the GPTeval3D benchmark.
Experiments across diverse objects and complex scenes show that VLM3D significantly outperforms prior SDS-based methods in semantic fidelity, geometric coherence, and spatial correctness.

\end{abstract}

\section{Introduction}
\label{sec:introduction}

Text-to-3D generation—the task of creating 3D content from natural language descriptions—has drawn significant attention due to its broad applications in vision and graphics\cite{poole2022dreamfusion, wang2023prolificdreamer, chung2023luciddreamer, wang2023score, yu2023text, shi2023mvdream, katzir2023noise}.
A particularly promising direction is Score Distillation Sampling (SDS)\cite{poole2022dreamfusion}, which has rapidly become the dominant paradigm in this field\cite{wang2023prolificdreamer}.
SDS leverages pre-trained text-to-image diffusion models~\cite{ho2020denoising} as ``teachers'' to supervise the optimization of a 3D representation—typically a Neural Radiance Field (NeRF)~\cite{mildenhall2020nerf} or 3D Gaussian Splatting (3DGS)\cite{kerbl2023gaussian, tang2023dreamgaussian}—by denoising multi-view 2D renderings.
This approach effectively distills 2D generative priors into the 3D domain, aligning the 3D representations with the input text prompt\cite{yang2024semantic}, resulting in plausible 3D content\cite{wang2023prolificdreamer, yan2024flow} from even non-experts.

Despite its widespread applications, SDS-based text-to-3D generation still faces critical limitations in producing precisely text-aligned and 3D-consistent assets.
\begin{itemize}[leftmargin=*]
\item First, many popular text-to-image diffusion models rely on CLIP-style text encoders for prompt understanding\cite{radford2021clip, zarei2024understanding, abbasi2025analyzing, zhuang2024magnet}, which provide only weak semantic grounding. As a result, the generated 3D assets often coarsely reflect the prompt and struggle to capture its precise meaning. This limitation becomes particularly pronounced for fine-grained or long prompts, especially those involving multiple objects, intricate spatial relationships, or subtle attributes. 
\item Second, SDS depends entirely on 2D diffusion models as supervisory signals. While effective for 2D image generation, these models lack explicit 3D spatial reasoning, leading to typical failure modes such as geometric inconsistencies across views or the "Janus problem,"\cite{wang2024taming, nath2025deep, hong2023debiasing, seo2024retrieval} where an object appears inconsistently from different angles.
\end{itemize}
%

\begin{figure*}[t]
    \centering
    \includegraphics[width=0.99\textwidth]{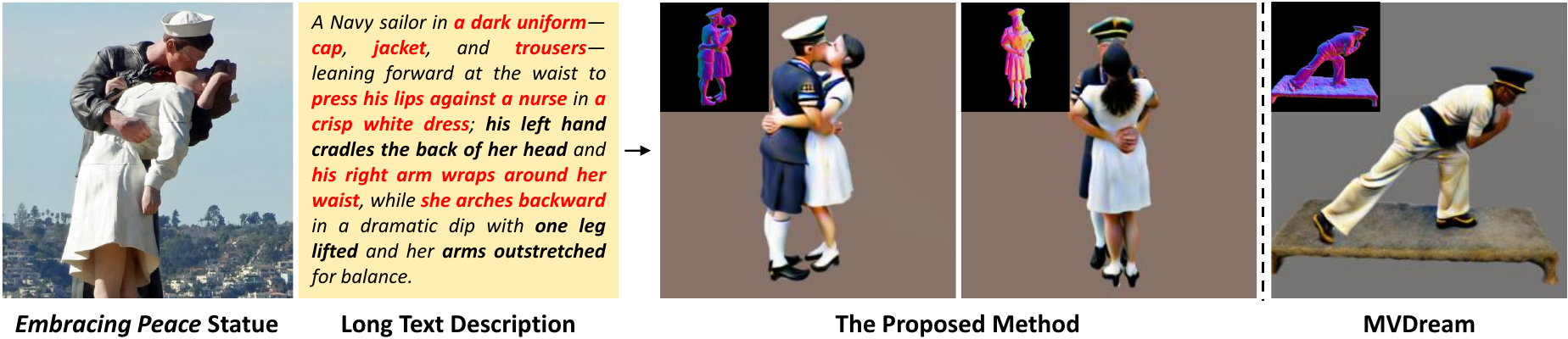}
    \caption{
    \textbf{Reproducing the ``Embracing Peace'' Statue with VLM3D. } {We challenge VLM3D and the baseline MVDream with a detailed description of San Diego’s iconic ``Embracing Peace'' monument, tasking each to generate a 3D asset from text alone. VLM3D accurately recovers both sailor and nurse figures, their attire, and the signature kissing pose, whereas MVDream omits the nurse entirely. Key successfully reproduced details are highlighted in bold red; remaining omissions (bold black) point to opportunities for further improvement.}
    }
    \label{fig:teaser}
    \vspace{-1em}
\end{figure*}

To address these two challenges, we propose VLM3D, a novel text-to-3D generation framework that harnesses Vision–Language Models (VLMs) as powerful semantic and spatial priors inside the SDS pipeline.
Trained on massive image–text datasets, VLMs provide significantly stronger cross-modal semantic understanding than baselines with traditional CLIP models~\cite{radford2021clip, yu2022coca, ma2024sycoca, cheng2024spatialrgpt, hong20233d, tang2024any2point}.
Furthermore, vision-language models' great ability to directly measure the semantic similarity between an image and a long textual description~\cite{jia2021scaling, choi2025goal}, makes them well suited to serve as global, semantically grounded reward signals in the SDS process.
Compared to conventional text-to-image diffusion priors, VLMs offer rich, language-grounded supervision that enables precise prompt alignment~\cite{jiang2024comat, mrini2024fast, wang2025hcma}. This not only supports the generated 3D assets accurately reflecting the overall semantic meaning but also enables fine-grained or long prompts, producing high-quality and detailed 3D models that match complex descriptions.

Moreover, we observe that VLMs implicitly encode strong 3D priors, since the language itself provides rich spatial understanding~\cite{cheng2024spatialrgpt, ma2024spatialpin, chen2024spatialvlm, pan2025metaspatial, ma2025spatialreasoner} that can serve as effective constraints in the creation of 3D content.
By incorporating descriptions that specify 3D intentions—such as spatial relationships or consistency across views—VLMs significantly enhance 3D coherence for single-object generation and improve relational reasoning in multi-object scenes~\cite{zhang2025flatland, ling2025scenethesis, chatterjee2024revision}. This leads to substantial improvements in the structural and perceptual quality of the generated 3D assets.

We instantiate our approach using Qwen2.5-VL 7B~\cite{bai2025qwen2_5vl} as the reward backbone and develop a fully differentiable training pipeline that enables direct optimization of NeRF parameters via VLM-defined semantic rewards. 
Our framework, VLM3D, achieves state-of-the-art performance in text-to-3D generation, significantly outperforming existing SDS-based methods in semantic alignment, geometric consistency, and visual quality. 
To the best of our knowledge, VLM3D  makes the first attempt to shift from implicit score distillation of 2D generative priors to explicit semantic optimization via large vision–language models, creating a principled and generalizable path forward for text-conditioned 3D synthesis and beyond.


\vspace{-3mm}
\section{Related Works}
\label{sec:relatedworks}

\textbf{Diffusion Distillation for Text-to-3D Generation}
Diffusion\cite{ho2020denoising, dhariwal2021diffusion, sohl2015deep} distillation methods transfer the rich priors of pre-trained 2D diffusion models into 3D representations by optimizing a differentiable renderer against a Score Distillation Sampling (SDS) loss. DreamFusion~\cite{poole2022dreamfusion} first demonstrated this concept by distilling the denoising gradients of a large text-to-image diffusion model into a Neural Radiance Field (NeRF)\cite{mildenhall2020nerf} via random-view rendering and iterative gradient guidance, enabling 3D asset creation from text without any 3D training data. Despite its flexibility, DreamFusion often converges slowly, produces modest resolutions, and can suffer from multi-view inconsistencies such as the Janus problem.

Subsequent methods\cite{wang2023prolificdreamer, chung2023luciddreamer, shi2023mvdream, chen2023fantasia3d, lin2023magic3d,  wei2024adversarial} have introduced architectural and algorithmic improvements to address these issues. Magic3D adopts a coarse-to-fine strategy in which a low-resolution NeRF is first optimized under a latent diffusion prior and then converted into a high-resolution mesh for fine-tuning, substantially boosting fidelity and convergence speed \cite{lin2023magic3d}. 
MVDream trains a diffusion model on multi-view image collections, producing 3D-consistent priors that, when distilled, yield stronger cross-view coherence \cite{shi2023mvdream}. 
ProlificDreamer frames the optimization as a particle-based variational inference problem, improving diversity and detail by modeling uncertainty \cite{wang2023prolificdreamer}. 

Alternatively, DreamReward effectively encodes human preferences by integrating a learned 3D reward model into SDS loss through reinforcement learning from human feedback \cite{ye2024dreamreward}. Moreover, complementary refinement techniques—such as Deepmesh (for point cloud to mesh conversion)\cite{zhao2025deepmesh} and DMTet\cite{shen2021deep}—can be employed to further enhance and fine-tune 3D assets generated by SDS. Together, these advances have significantly improved speed, resolution, diversity, and consistency in diffusion-based text-to-3D synthesis. However, because all of these methods rely on 2D text-to-image\cite{nichol2021glide, li2022efficient, zhang2023text} diffusion models that typically employ CLIP-style encoders\cite{radford2021learning, raffel2020exploring} for language guidance, they inherit limitations in fine-grained semantic grounding and explicit 3D spatial reasoning, resulting in residual misalignment and view inconsistencies.

Through these advancements, diffusion distillation approaches have made substantial strides in speed, resolution, consistency, and semantic alignment, establishing a powerful foundation for frameworks like VLM3D that seek to inject explicit semantic and spatial rewards into text-to-3D synthesis.

\textbf{Vision–Language Models}
Vision–Language Models (VLMs) jointly learn from large-scale image–text corpora to produce unified embeddings that capture both visual content and linguistic semantics. Early VLMs, such as CLIP\cite{radford2021clip, jia2021scaling, bao2021beit}, employ a dual-encoder architecture trained via contrastive learning on hundreds of millions of image–text pairs. By maximizing cosine similarity for matched pairs and minimizing it for mismatched ones, CLIP achieves strong zero-shot classification and retrieval performance across diverse vision benchmarks. However, its discriminative training paradigm limits generative flexibility\cite{yu2022coca, li2024erroneous, ramesh2021zero} and fine-grained spatial reasoning\cite{chen2024spatialvlm, qiu2025refining, wang2024sclip, patel2024tripletclip}.

Modern large VLMs extend this foundation by integrating a powerful autoregressive language model with a visual encoder in an end-to-end, generative framework. Methods like BLIP-2~\cite{li2023blip} and MiniGPT-4~\cite{zhu2023minigpt} leverage lightweight query modules or adapters to bridge frozen vision and language backbones\cite{li2023blip, zhu2023minigpt}, then fine-tune on multimodal instruction data to support open-ended tasks such as image captioning, visual question answering, dense grounding, and dialogue\cite{liu2023visual}. More recently, models such as LLaVA~\cite{liu2023visual, 2023llavarlhf} and Qwen-VL~\cite{bai2025qwen2_5vl} have demonstrated advanced spatial grounding capabilities—localizing objects, understanding complex relations, and reasoning over multi-object scenes—while retaining strong semantic alignment. Notably, these models incorporate advanced vision modules—for example, Qwen2.5-VL employs dynamic resolution processing to natively handle variable-size images and absolute time encoding for precise long-range video reasoning —further enhancing their spatiotemporal understanding\cite{bai2025qwen2_5vl, wang2024qwen2}. Because these VLMs provide language-grounded similarity measures and implicitly encode spatial relationships, they serve as ideal reward functions within an SDS pipeline\cite{poole2022dreamfusion, wang2023prolificdreamer, chachy2025rewardsds, tang2023stable}.

\vspace{-3mm}
\section{Preliminary}
\label{sec:preliminary}

\paragraph{2D Text-to-Image Diffusion Models} Diffusion models \cite{ho2020denoising, sohl2015deep, song2020score} define a forward–time SDE that gradually injects noise into a data sample and a corresponding reverse–time SDE that removes noise to generate samples. Concretely, for an image $\mathbf{x}_0$ conditioned on text prompt $y$ (i.e., $\mathbf{x}_0\sim p_{\mathrm{data}}(\cdot\mid y)$ ), the forward SDE is
\[
  d\mathbf{x}_t = f(\mathbf{x}_t, t)\,dt + g(t)\,d\mathbf{w}_t,
\]
and the reverse–time SDE is
\[
  d\mathbf{x}_t = \bigl[f(\mathbf{x}_t, t) - g(t)^2 \nabla_{\mathbf{x}_t}\log p_t(\mathbf{x}_t\mid y)\bigr]\,dt
  + g(t)\,d\bar{\mathbf{w}}_t,
\]
where $t\in[0,T]$, $f$ is the drift, $g$ the diffusion coefficient, $\mathbf{w}_t$ and $\bar{\mathbf{w}}_t$ are forward and reverse Wiener processes, and $p_t(\cdot\mid y)$ is the marginal at time $t$. A neural network $s_\phi(\mathbf{x},y,t)\approx\nabla_{\mathbf{x}}\log p_t(\mathbf{x}\mid y)$ is trained via denoising score matching to approximate the score function:
\[
  \mathcal{L}_{\mathrm{DSM}}
  = \mathbb{E}_{t,\mathbf{x}_0,\boldsymbol{\epsilon}}\!\Bigl[\lambda(t)\,\bigl\|s_\phi(\mathbf{x}_t,y,t)
    + \tfrac{\boldsymbol{\epsilon}}{\sigma(t)}\bigr\|^2\Bigr],
\]
with $\mathbf{x}_t=\mathbf{x}_0+\sigma(t)\,\boldsymbol{\epsilon}$ and $\boldsymbol{\epsilon}\sim\mathcal{N}(\mathbf{0},\mathbf{I})$.

\paragraph{Score Distillation Sampling (SDS)} SDS \cite{poole2022dreamfusion} repurposes a pretrained score network $s_\phi$ to optimize 3D scene parameters $\theta$ (e.g., NeRF weights). Let $I(\theta,v)=\text{Render}(\theta, v)$ be the image produced by a differentiable renderer parameterized by $\theta$ from viewpoint $v$. We set $\mathbf{x}_0 = I(\theta,v)$ and diffuse it to timestep $t$ via
\[
  \mathbf{x}_t = \mathbf{x}_0 + \sigma(t)\,\boldsymbol{\epsilon},\quad 
  \boldsymbol{\epsilon}\sim\mathcal{N}(\mathbf{0},\mathbf{I}),
\]
The SDS loss is then formulated as the weighted KL divergence between the distribution of noised renderings and the text-conditioned diffusion prior:
\begin{equation}\label{eqn:sds_loss_1}
    \mathcal{L}_{\mathrm{SDS}}(\theta) = \mathbb{E}_{t,c}\left[ \omega(t)\, D_{\mathrm{KL}}\Big( q(\boldsymbol{x}_t) \,\big\|\, p(\boldsymbol{x}_t|y) \Big) \right],
\end{equation}
where $w(t)$ is a timestep-dependent weight. Backpropagating through the renderer gives
\[
  \nabla_\theta \mathcal{L}_{\mathrm{SDS}}
  \approx \mathbb{E}_{t,\boldsymbol{\epsilon}}\Bigl[w(t)\bigl(s_\phi(\mathbf{x}_t,y,t)
    + \tfrac{\boldsymbol{\epsilon}}{\sigma(t)}\bigr)\,\frac{\partial I(\theta,v)}{\partial\theta}\Bigr].
\]
This gradient steers $\theta$ so that rendered views become more consistent with the text‐conditioned diffusion prior.

\vspace{-3mm}
\section{Method}
\label{sec:method}
\vspace{-3mm}
We now describe our proposed framework, VLM3D, which integrates a pre-trained VLM into a differentiable text-to-3D synthesis pipeline. In Sec.~\ref{sec:method:p1}, we first detail how the VLM is used as an explicit semantic and spatial reward, followed by our dual-query prompt design that jointly enforces content alignment and geometric consistency in Sec.~\ref{sec:method:p2}. We then describe our overall training objective in Sec.~\ref{sec:method:p3}, which combines this VLM-based reward with the SDS loss using a dynamic weighting strategy to progressively balance coarse alignment and fine detail synthesis during the optimization process.

\subsection{Vision–Language Model as an Explicit Semantic and Spatial Reward}
\label{sec:method:p1}
\vspace{-3mm}

We leverage a pre-trained VLM to provide a fully differentiable reward that measures both semantic fidelity and geometric consistency across multi-view renderings. Concretely, let the set of $N$ images rendered from our 3D representation parameters $\theta$ under viewpoints $v_i$
\begin{equation}
\mathcal{X} = \{x_i\}_{i=1}^N = \bigl\{I(\theta, v_1), \dots, I(\theta, v_N)\bigr\} 
\end{equation}
and the user’s text prompt $\mathbf{y}$ be the two inputs to the large VLM, we constrain the VLM function $Q(\cdot, \cdot)$, via carefully engineered prompts, to output exclusively “Yes” or “No”,
\begin{equation}
Q(\mathbf{y}, \mathcal{X}) \;\mapsto\;\{\text{Yes},\text{No}\}.
\end{equation}
Then we extract the final “Yes” and “No” logits of the VLM’s binary‐classification head, $z_{\text{yes}}$ and $z_{\text{no}}$. The corresponding probabilities are
\begin{equation}
P(\text{Yes}\mid \mathbf{y},\mathcal{X}) = \frac{e^{z_{\text{yes}}}}{e^{z_{\text{yes}}}+e^{z_{\text{no}}}}, \quad 
P(\text{No}\mid \mathbf{y},\mathcal{X}) = \frac{e^{z_{\text{no}}}}{e^{z_{\text{yes}}}+e^{z_{\text{no}}}}.
\end{equation}
We define the VLM reward as the log‐odds of a “Yes” response:
\begin{equation}
\label{eq:vlm_reward}
r_{\text{VLM}}
= \log P(\text{Yes}\mid \mathbf{y},\mathcal{X}) - \log P(\text{No}\mid \mathbf{y},\mathcal{X})
= z_{\text{yes}} - z_{\text{no}}.
\end{equation}
The entire mapping $\theta \to \mathcal{X} \to r_{\text{VLM}}$ is differentiable, so we can backpropagate through the VLM to update $\theta$ directly.

\subsection{VLM Prompt Design}
\label{sec:method:p2}
Our full prompt $\mathbf{y}$ to the VLM is:

\textit{``Carefully evaluate the provided images, which show multiple views of a single 3D object. Does the underlying 3D object, considering all views together, meet all of the following criteria simultaneously? \\
1. Content Match: The object corresponds to the description: \{text description\}. \\
2. Geometric Quality: Based on all views combined, the object appears geometrically sound and consistent. There are no visible signs of major flaws such as multiple faces on one part (Janus-faced issue), broken surfaces, intersecting geometry, or highly unrealistic polygonal facets when considering the object from these different perspectives. \\
Strictly respond with only `Yes' or `No'.''}

Since we want the VLM to guide both semantic alignment and spatial/geometric correctness, we prompt the VLM with two separate “yes or no” queries: one asking whether the rendered views accurately depict the content descriptions, and the other asking whether those views are geometrically consistent and high-fidelity. This dual-query setup ensures that the reward in {Eq.~\ref{eq:vlm_reward}} captures both semantic alignment and spatial coherence.

\begin{figure*}[t]
    \centering
    \includegraphics[width=0.999\textwidth]{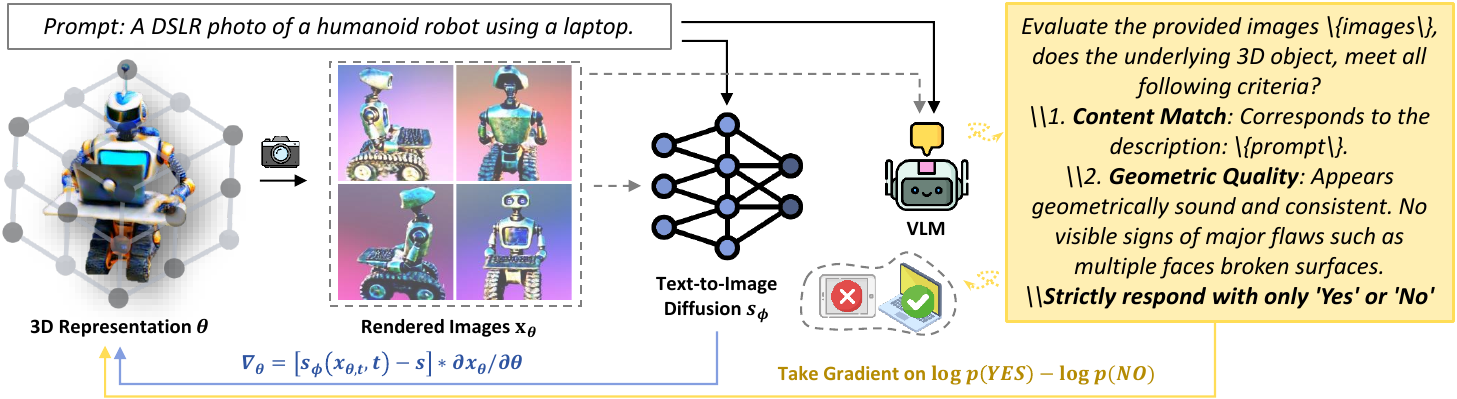}
    \caption{
    \textbf{Overview of VLM3D.} {VLM3D integrates a pretrained vision–language model as a differentiable reward into the SDS pipeline. Its dual‐query prompt—one for content matching and one for geometric consistency and quality—simultaneously enforces semantic fidelity, geometric coherence, and spatial correctness in text‐to‐3D synthesis.}
    }
    \label{fig:overview}
    \vspace{-1em}
\end{figure*}

\subsection{Training Loss and Optimization Strategy}
\label{sec:method:p3}
The full training objective of VLM3D combines the VLM reward with the standard Score Distillation Sampling (SDS) loss~\cite{poole2022dreamfusion}, which regularizes generated views to match the texture and style statistics of powerful pre-trained diffusion models. Specifically, we minimize
\begin{equation}
\label{eq:total_loss}
\mathcal{L}_{\text{total}}
= \mathcal{L}_{\text{SDS}} - \lambda_{\text{VLM}} r_{\text{VLM}}.
\end{equation}
where  $\lambda_{\text{VLM}}$ is the balancing factor between two loss terms.

We adopt a dynamic schedule for $\lambda_{\text{VLM}}$  during VLM3D training. Initially, we set $\lambda_{\text{VLM}}$  high to enforce strong semantic and geometric constraints, ensuring that the coarse shape and layout align with the prompt. We then gradually decay $\lambda_{\text{VLM}}$  so that the SDS loss predominates, refining textures, lighting, and fine details. In the final phase, we fix $\lambda_{\text{VLM}}$ at its reduced value and continue training for several more iterations. Empirically, this annealing schedule accelerates convergence, suppresses view‐inconsistency artifacts, and yields high-fidelity 3D assets that are both semantically precise and visually realistic.

\section{Experiment}
\label{sec:experiment}

We evaluate VLM3D's performance in generating coherent 3D scenes from diverse text prompts. VLM3D is compared with a comprehensive suite of existing zero-shot text-to-3D generative models, identifies its key components enabling accurate 3D geometry, and explores its qualitative capabilities under severe challenges (e.g., long texts, multi-object scenes, rare words). For further implementation details and additional results, please refer to the supplementary material. 

\subsection{Experimental Setup}
\label{sec:exp-setup}

\paragraph{Datasets and Evaluation}  Evaluation is conducted on the public GPTEval3D benchmark \cite{wu2024gpt}, which contains a diverse set of text prompts. In this benchmark, we employ GPT-4o-mini to perform pairwise comparisons between methods, calcualting Elo ratings that emulate human judgments of text alignment, 3D plausibility, and texture–geometry coherence. Additionally, we design custom, challenging prompts for ablation studies and to probe the limitations of our approach.

\paragraph{Baselines} We compare VLM3D against representative score distillation-based text-to-3D methods: DreamFusion~\cite{poole2022dreamfusion}, DreamGaussian~\cite{tang2023dreamgaussian}, Latent-NeRF~\cite{metzer2022latent}, Magic3D~\cite{lin2023magic3d}, and ProlificDreamer~\cite{wang2023prolificdreamer}, MVDream~\cite{shi2023mvdream}. Additionally, we also evaluate against the closely related human-preference reward–based approaches: DreamReward~\cite{ye2024dreamreward} and DreamDPO~\cite{zhou2025dreamdpo}.

\paragraph{Implementation} VLM3D requires minimal hyperparameter tuning; the following parameter set performs robustly across nearly all scenarios. In each iteration, we render eight views from the current 3D model and pass them to the VLM. We linearly anneal the VLM weight \(\lambda_{\text{VLM}}\) from 300 to 1 over the first 10,000 steps, then hold it constant at 1 for an additional 5,000 steps. A key challenge in enabling end-to-end differentiability through VLMs is that their image preprocessors typically detach gradients to accommodate diverse input formats. To overcome this, we reengineer the preprocessor to maintain gradient flow and will release the code.

We adopt MVDream \cite{shi2023mvdream} as our text-to-image diffusion backbone and Qwen2.5-VL 7B \cite{bai2025qwen2_5vl} as the VLM backbone. All experiments run on a single NVIDIA A100 GPU and complete in approximately two hours. Additional implementation details are provided in the supplementary material. 

\begin{figure*}[t]
    \centering
    \setlength{\tabcolsep}{1pt}
    \setlength{\fboxrule}{1pt}
    \resizebox{0.99\textwidth}{!}{
    \begin{tabular}{c}
    \begin{tabular}{cccc|cc}
        \multicolumn{1}{c}{{\footnotesize Dreamfusion}} &
        \multicolumn{1}{c}{{\footnotesize Magic3D}} &
        \multicolumn{1}{c}{{\footnotesize ProlificDreamer}} &
        \multicolumn{1}{c}{{\footnotesize MVDream}} &
        \multicolumn{2}{c}{{\footnotesize \textbf{VLM3D (Ours)}}}
        \\
        \includegraphics[width=0.16\textwidth]{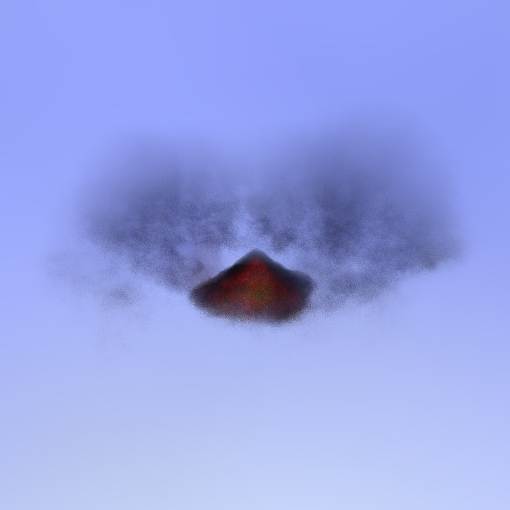} &
        \includegraphics[width=0.16\textwidth]{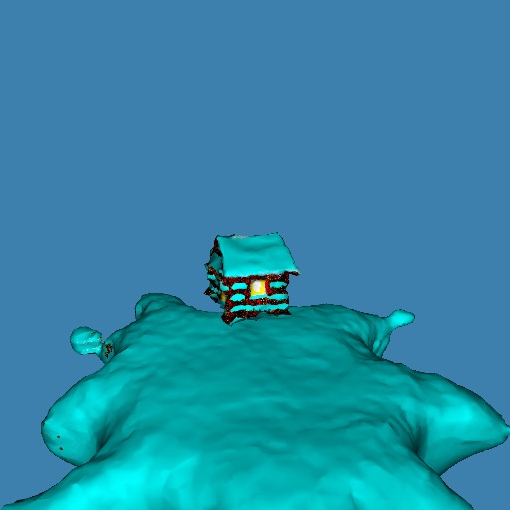} &
        \includegraphics[width=0.16\textwidth]{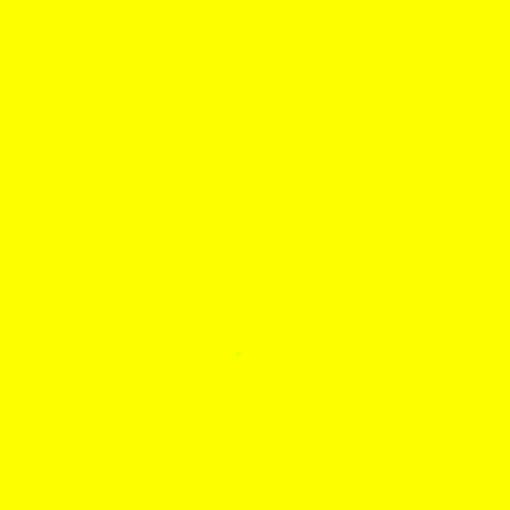} &
        \includegraphics[width=0.16\textwidth]{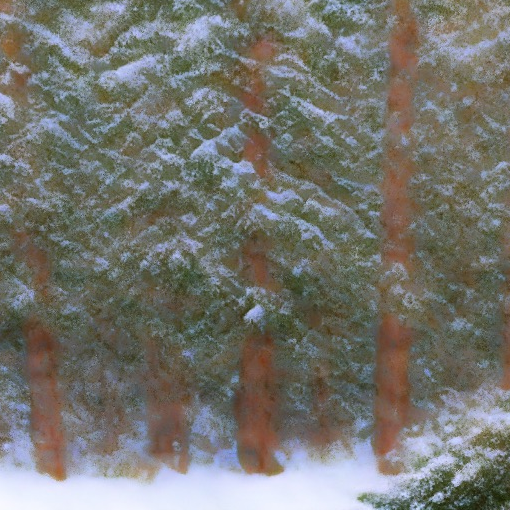} &
        \includegraphics[width=0.16\textwidth]{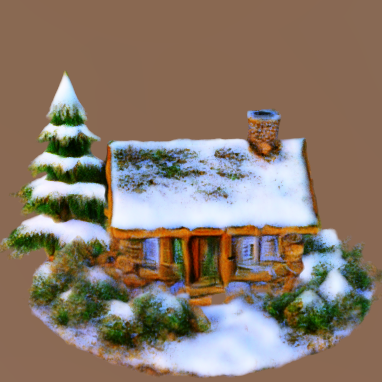} &
        \includegraphics[width=0.16\textwidth]{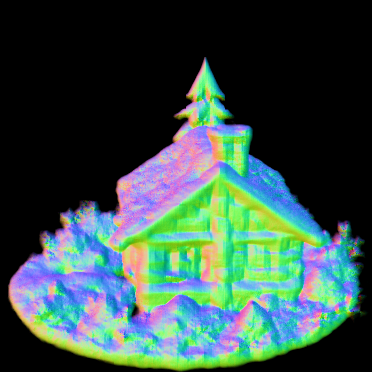} 
        \vspace{-0.1em}
        \\
        \multicolumn{6}{c}{{\prompts{A small, rustic \textcolor{red}{cabin} sits alone in a peaceful, \textcolor{red}{snow-covered forest}}}}
        \vspace{0.2em}
        \\
        \includegraphics[width=0.16\textwidth]{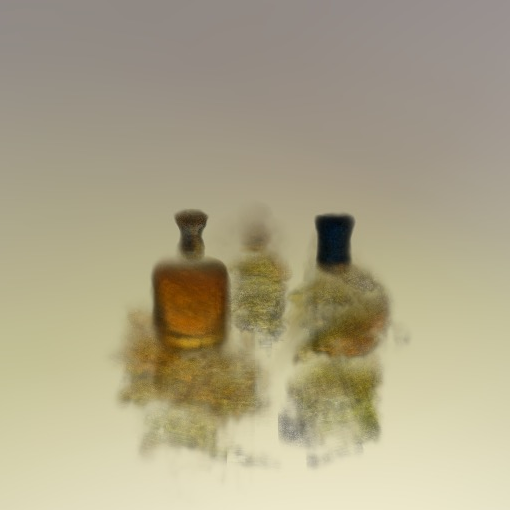} &
        \includegraphics[width=0.16\textwidth]{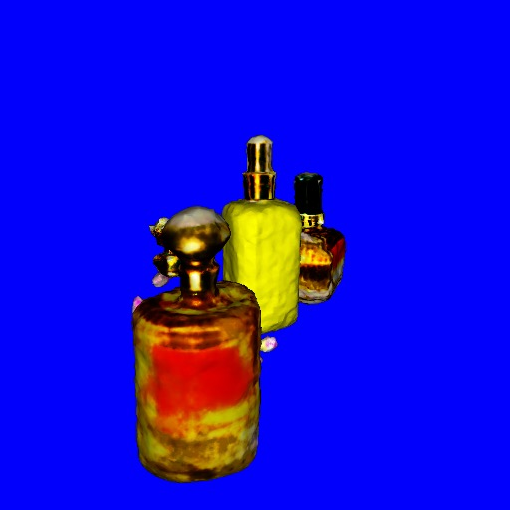} &
        \includegraphics[width=0.16\textwidth]{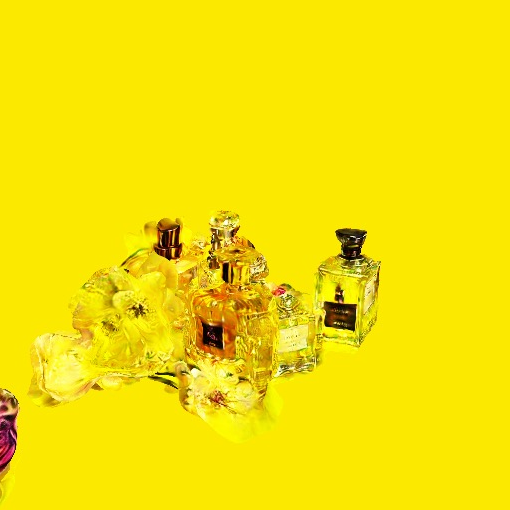} &
        \includegraphics[width=0.16\textwidth]{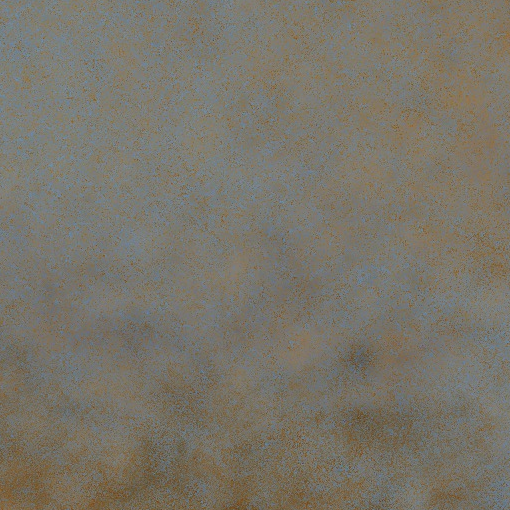} &
        \includegraphics[width=0.16\textwidth]{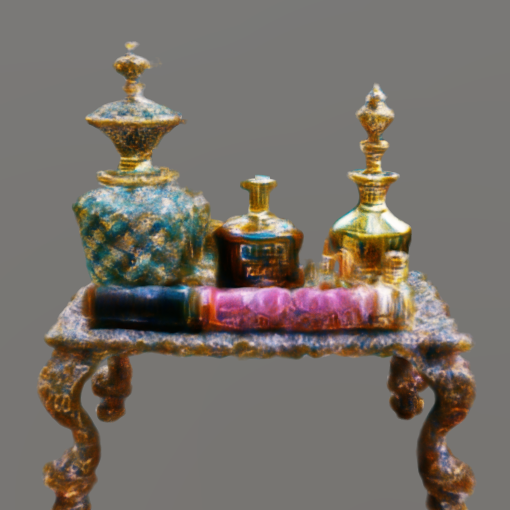} & 
        \includegraphics[width=0.16\textwidth]{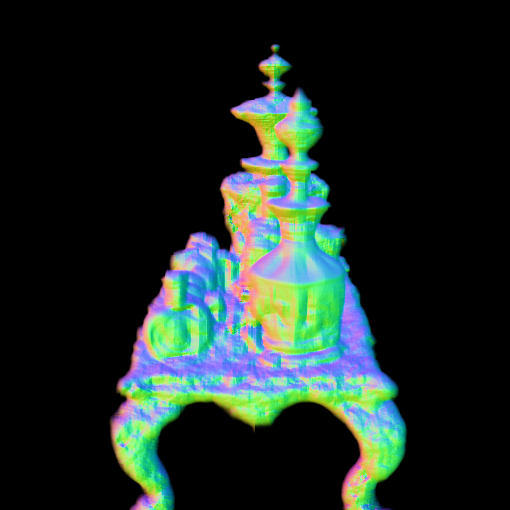}  
        \vspace{-0.1em}
        \\
        \multicolumn{6}{c}{{\prompts{An assortment of \textcolor{red}{vintage, fragrant perfumes} on display}}}
        \vspace{0.2em}
        \\
        \includegraphics[width=0.16\textwidth]{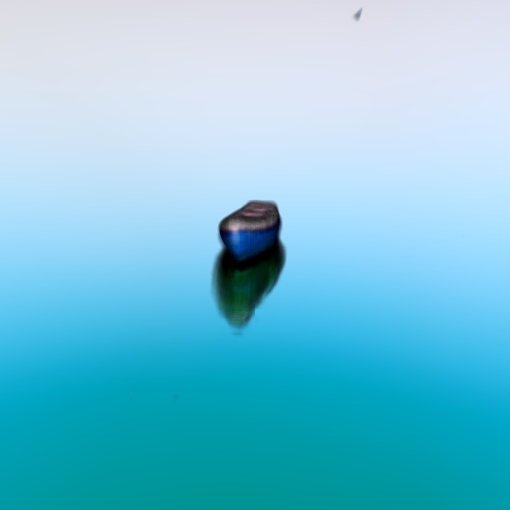} &
        \includegraphics[width=0.16\textwidth]{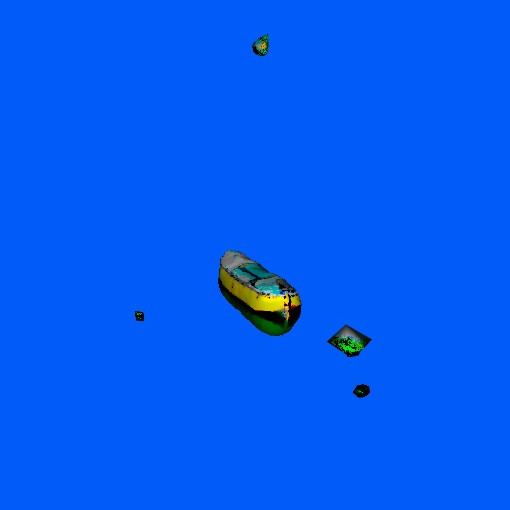} &
        \includegraphics[width=0.16\textwidth]{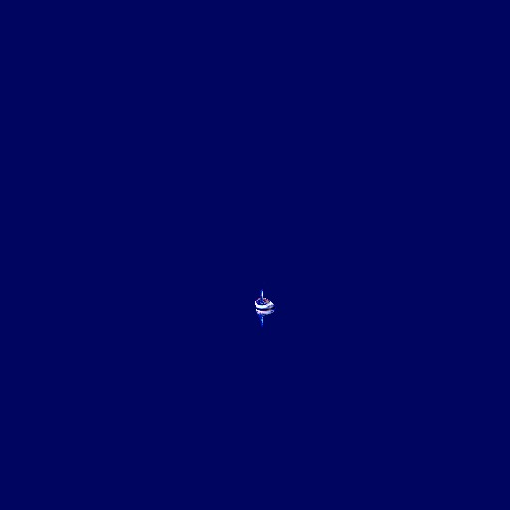} &
        \includegraphics[width=0.16\textwidth]{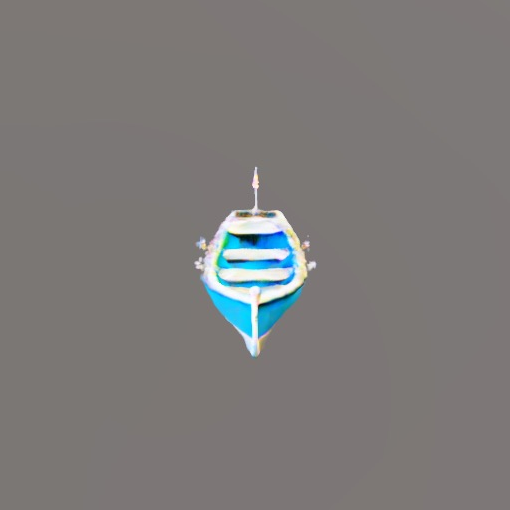} &
        \includegraphics[width=0.16\textwidth]{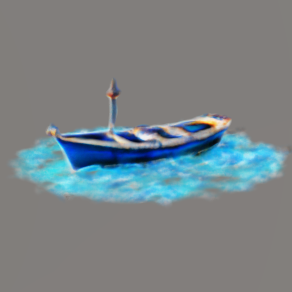} &
        \includegraphics[width=0.16\textwidth]{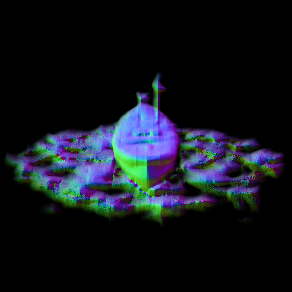}
        \vspace{-0.1em}
        \\
        \multicolumn{6}{c}{{\prompts{A boat \textcolor{red}{floating on calm water}}}}
        \vspace{0.2em}
        \\
        \includegraphics[width=0.16\textwidth]{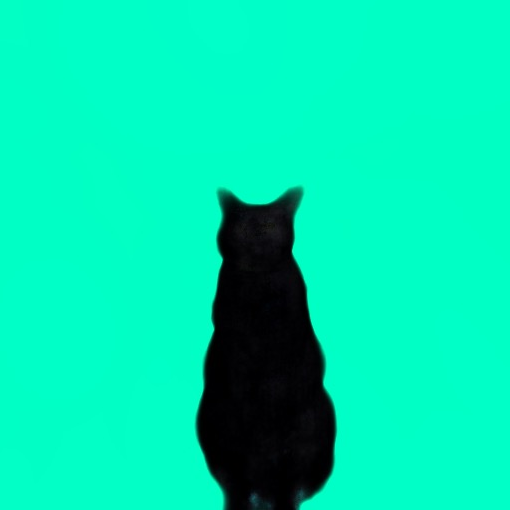} &
        \includegraphics[width=0.16\textwidth]{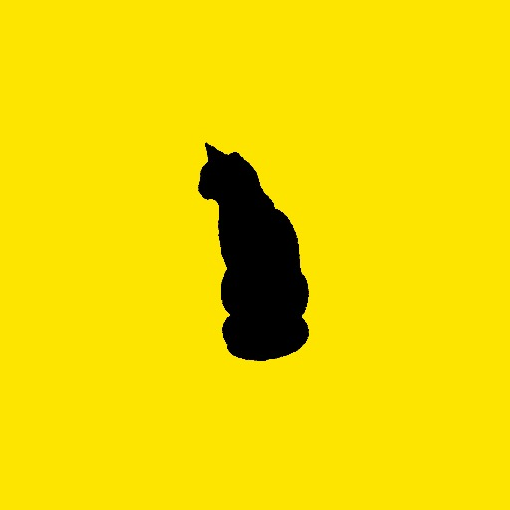} &
        \includegraphics[width=0.16\textwidth]{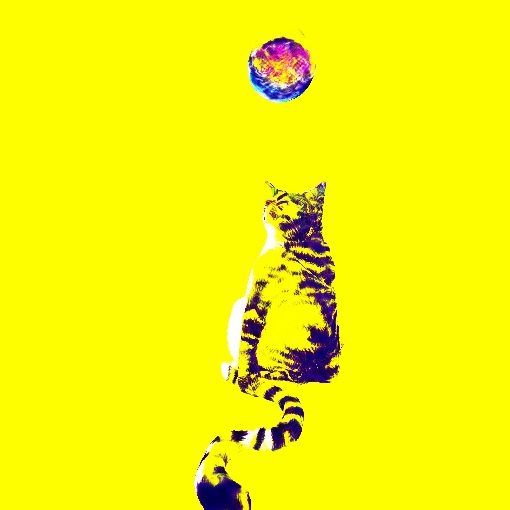} &
        \includegraphics[width=0.16\textwidth]{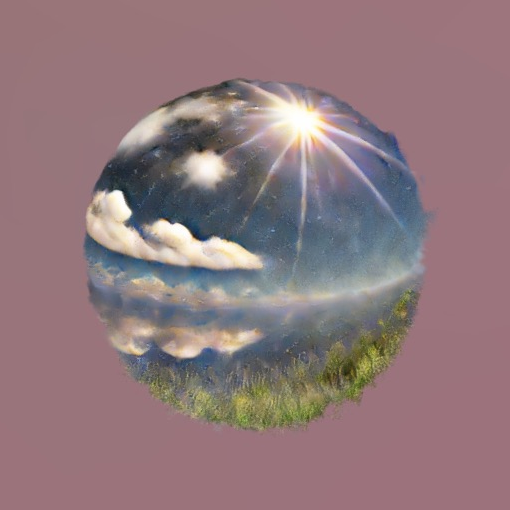} &
        \includegraphics[width=0.16\textwidth]{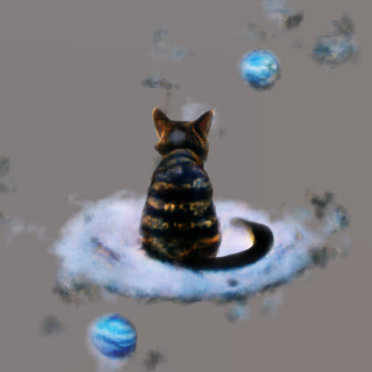} &
        \includegraphics[width=0.16\textwidth]{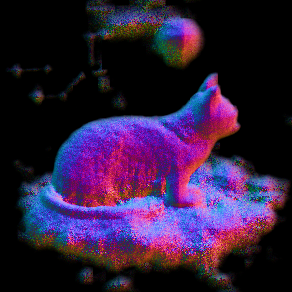}
        \vspace{-0.1em}
        \\
        \multicolumn{6}{c}{{\prompts{A cat pondering \textcolor{red}{the mysteries of the universe}}}}
    \end{tabular}
    \end{tabular}}
    \label{subfig:mesh}
    \vspace{-0.1em}
    \caption{
    \textbf{ Comparison of VLM3D and Baseline Methods on the GPTEval3D Benchmark~\cite{wu2024gpt}.} VLM3D outperforms baseline methods in geometric consistency, 3D plausibility, texture richness, and text alignment. Baselines frequently miss subtle semantic cues (highlighted in red), whereas VLM3D faithfully captures these details, yielding more accurate and visually coherent 3D assets. We present a rendered image and a normal map from an alternate viewpoint produced by VLM3D.
    }
    \label{fig:3d_comparison}
    \vspace{-0.3cm}
\end{figure*}

\begin{figure*}[t]
    \centering
    \setlength{\tabcolsep}{1pt}
    \setlength{\fboxrule}{1pt}
    \resizebox{0.99\textwidth}{!}{
    \begin{tabular}{c}
    \begin{tabular}{cc|cc|ccc}
        \multicolumn{2}{c}{{DreamReward~\cite{ye2024dreamreward}}} &
        \multicolumn{2}{c}{{DreamDPO~\cite{zhou2025dreamdpo}}} &
        \multicolumn{3}{c}{{\textbf{VLM3D (Ours)}}}
        \\
        \includegraphics[width=0.16\textwidth]{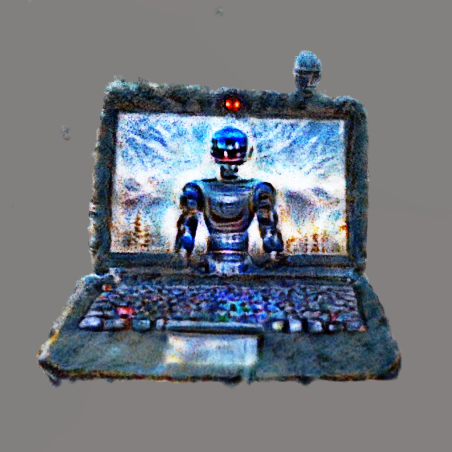} &
        \includegraphics[width=0.16\textwidth]{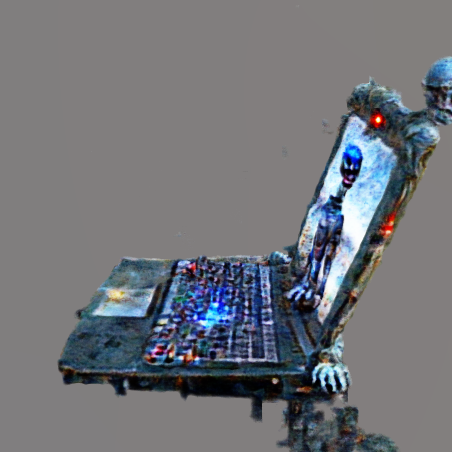} &
        \includegraphics[width=0.16\textwidth]{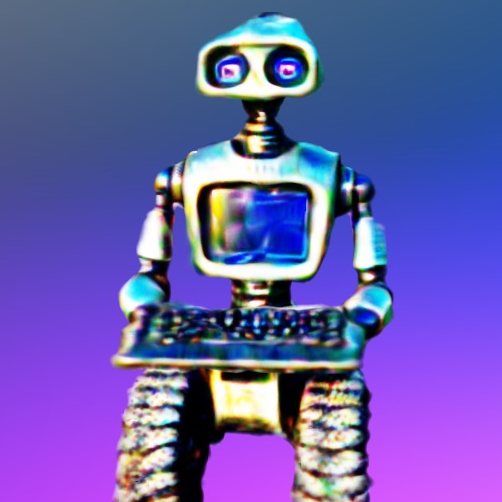} &
        \includegraphics[width=0.16\textwidth]{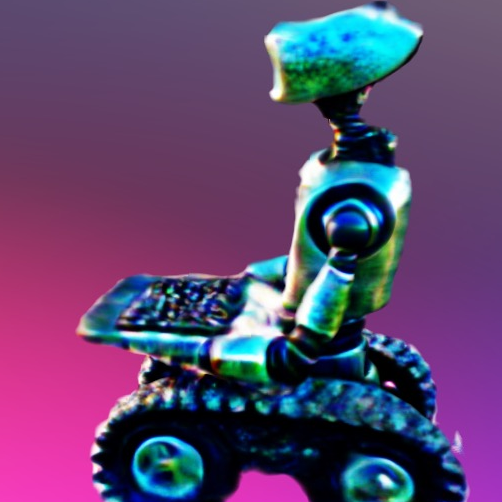} &
        \includegraphics[width=0.16\textwidth]{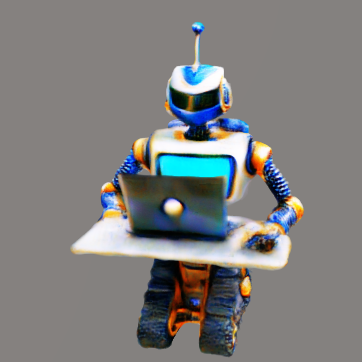} &
        \includegraphics[width=0.16\textwidth]{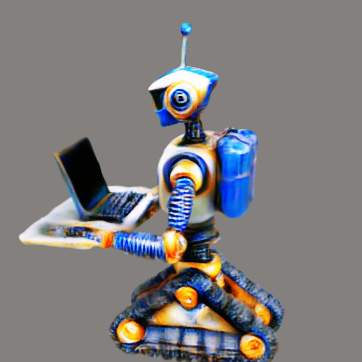} &
        \includegraphics[width=0.16\textwidth]{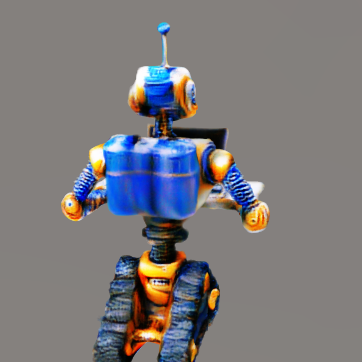}  
        \vspace{-0.1em}
        \\
        \multicolumn{7}{c}{{\prompts{A DSLR photo of a humanoid robot \textcolor{red}{using a laptop}}}}
        \vspace{0.2em}
        \\
        \includegraphics[width=0.16\textwidth]{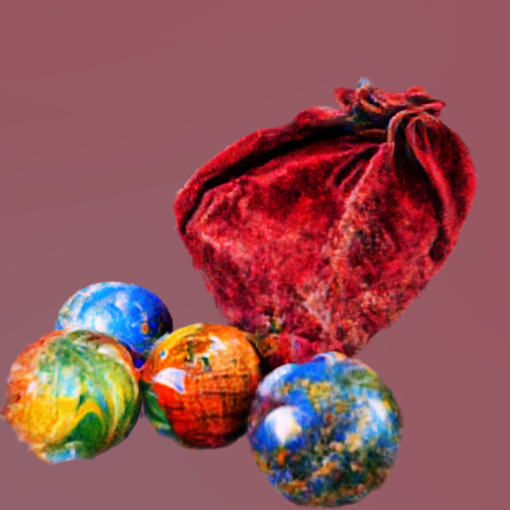} &
        \includegraphics[width=0.16\textwidth]{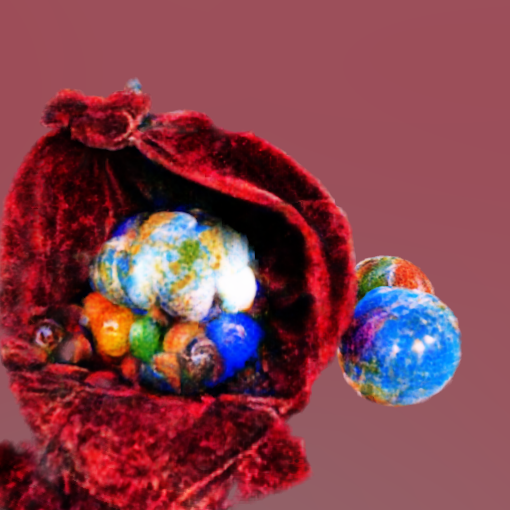} &
        \includegraphics[width=0.16\textwidth]{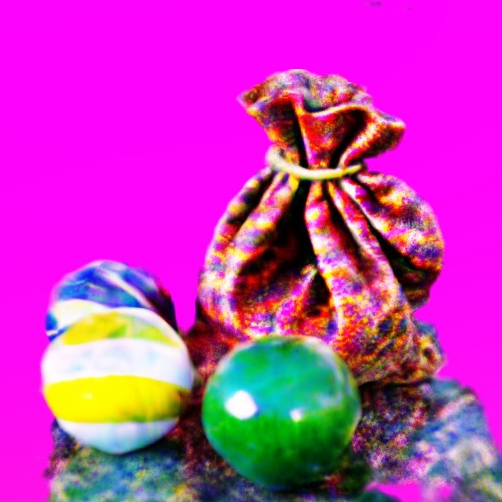} &
        \includegraphics[width=0.16\textwidth]{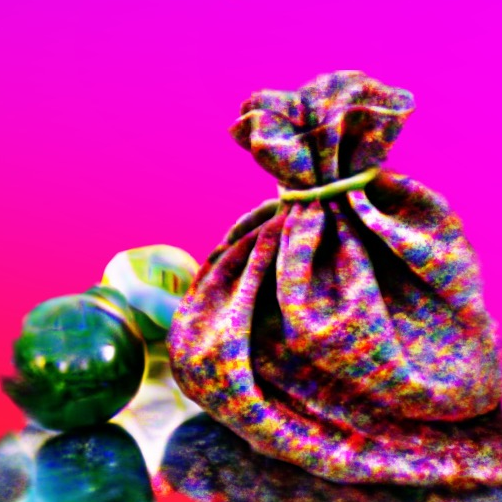} &
        \includegraphics[width=0.16\textwidth]{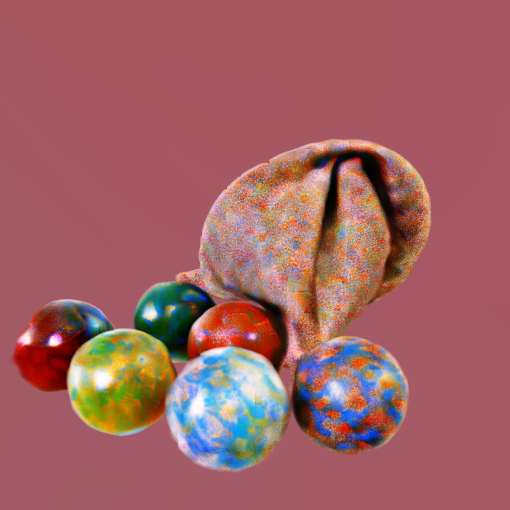} &
        \includegraphics[width=0.16\textwidth]{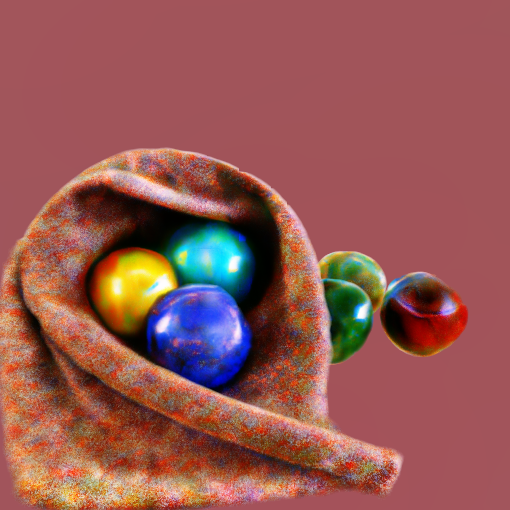} &
        \includegraphics[width=0.16\textwidth]{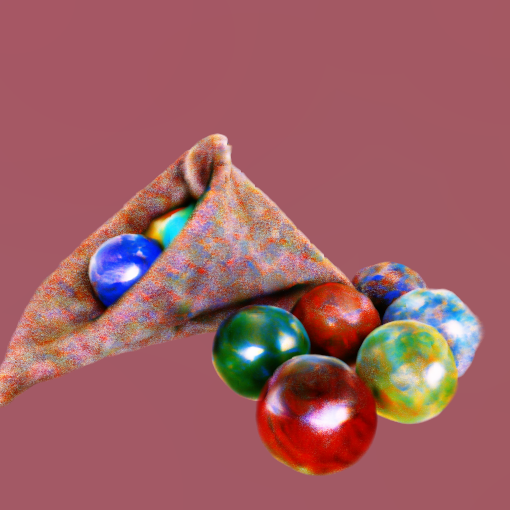}  
        \vspace{-0.1em}
        \\
        \multicolumn{7}{c}{{\prompts{A bunch of colorful marbles \textcolor{red}{spilling out} of a red {velvet} bag}}}
        \vspace{0.2em}
        \\
        \includegraphics[width=0.16\textwidth]{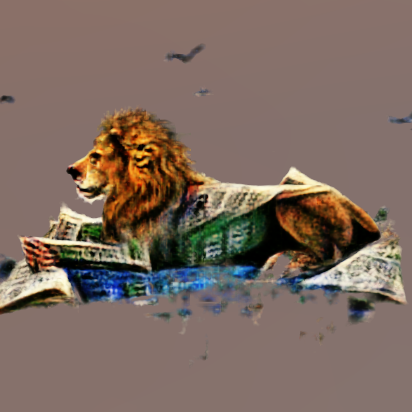} &
        \includegraphics[width=0.16\textwidth]{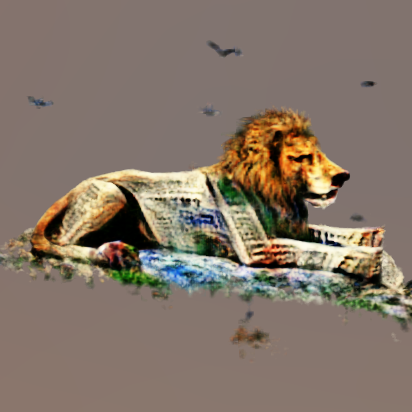} &
        \includegraphics[width=0.16\textwidth]{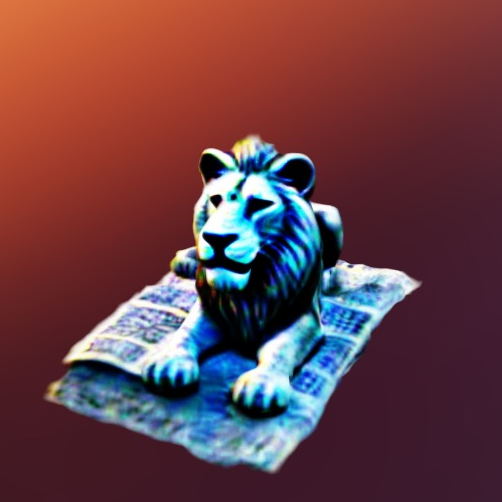} &
        \includegraphics[width=0.16\textwidth]{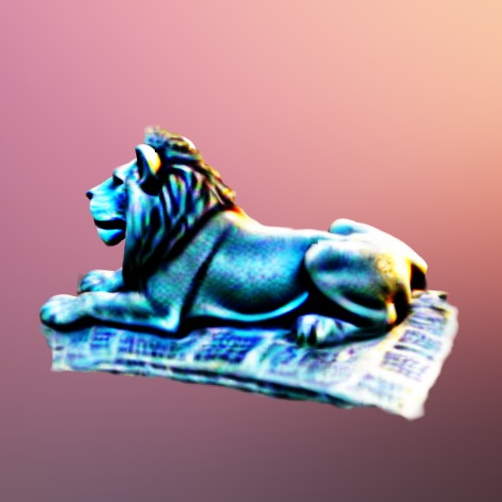} &
        \includegraphics[width=0.16\textwidth]{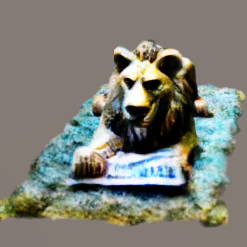} &
        \includegraphics[width=0.16\textwidth]{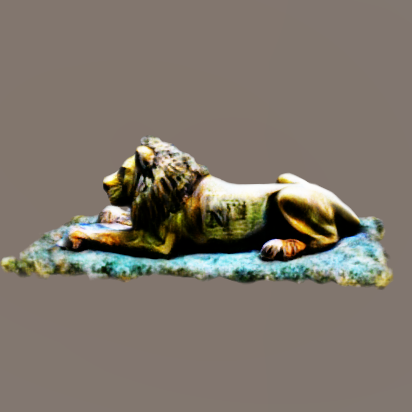} &
        \includegraphics[width=0.16\textwidth]{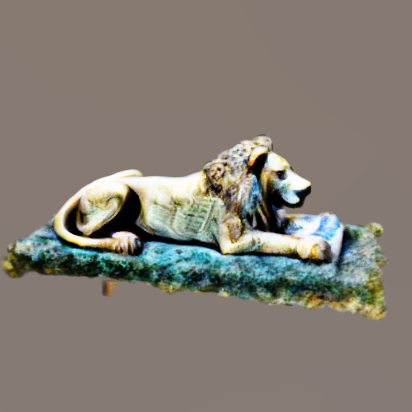}  
        \vspace{-0.1em}
        \\
        \multicolumn{7}{c}{{\prompts{A DSLR photo of a lion \textcolor{red}{reading the newspaper}}}}
        \\
        \includegraphics[width=0.16\textwidth]{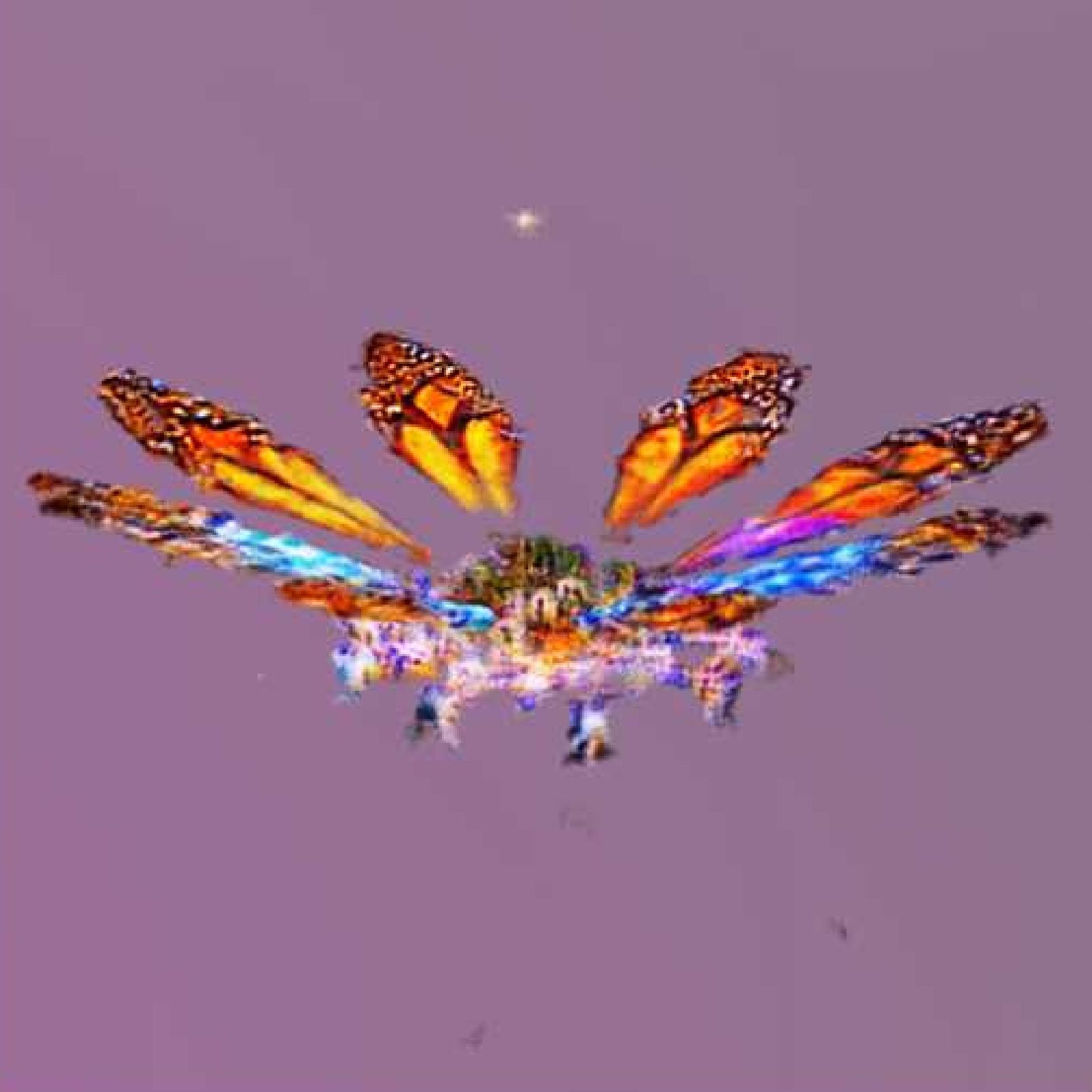} &
        \includegraphics[width=0.16\textwidth]{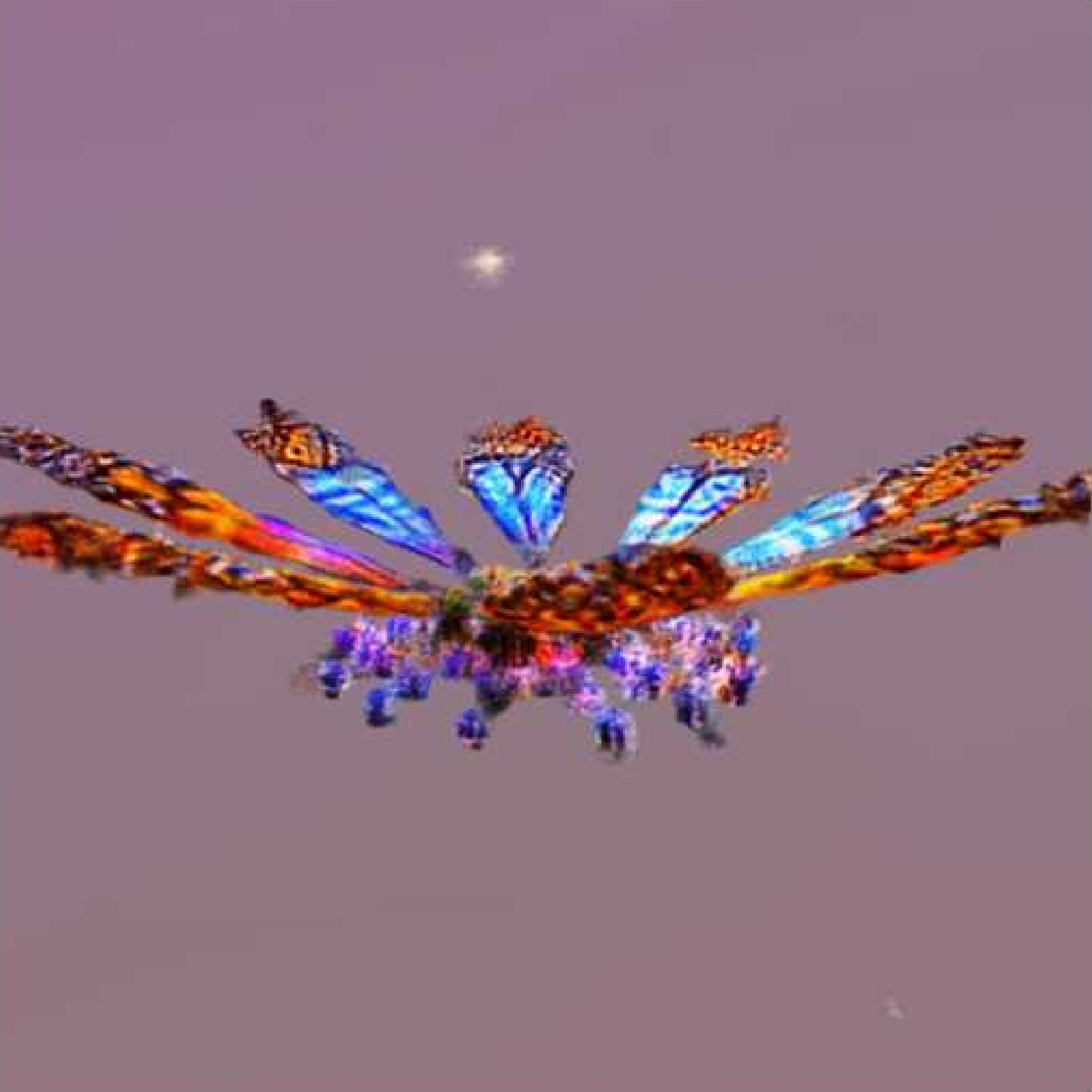} &
        \includegraphics[width=0.16\textwidth]{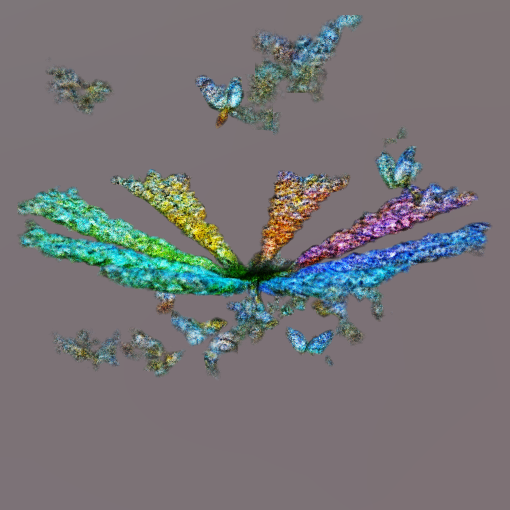} &
        \includegraphics[width=0.16\textwidth]{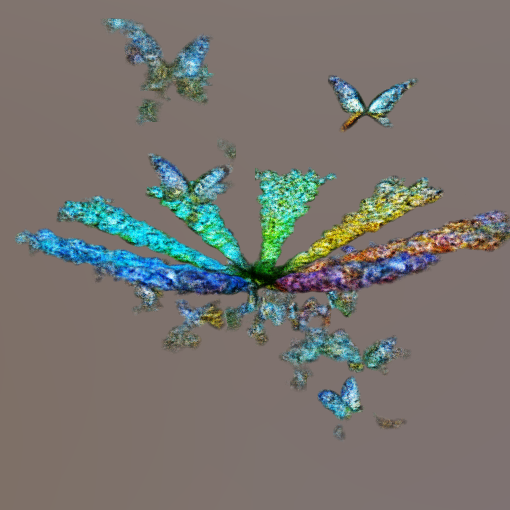} &
        \includegraphics[width=0.16\textwidth]{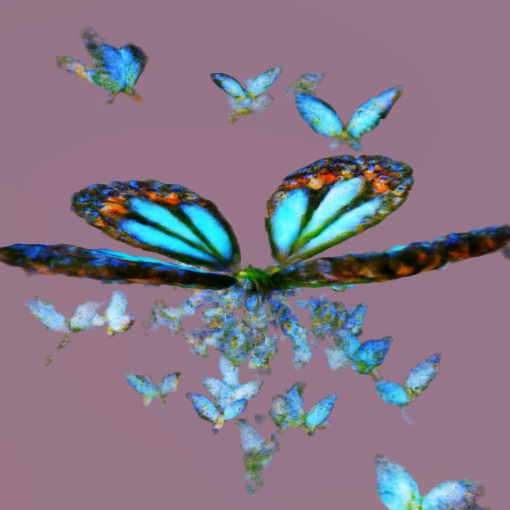} &
        \includegraphics[width=0.16\textwidth]{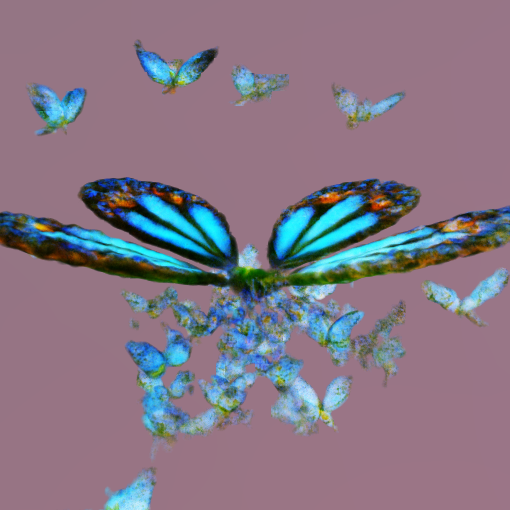} &
        \includegraphics[width=0.16\textwidth]{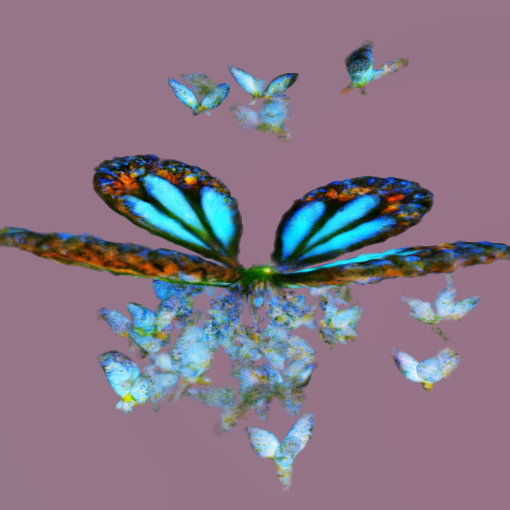} 
        \vspace{-0.1em}
        \\
        \multicolumn{7}{c}{{\prompts{A mesmerizing dance performed by \textcolor{red}{a kaleidoscope of butterflies}}}}
        \\
        \includegraphics[width=0.16\textwidth]{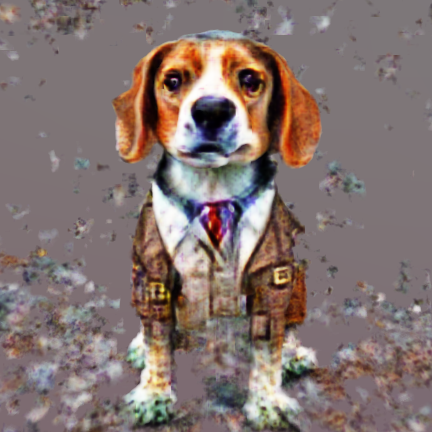} &
        \includegraphics[width=0.16\textwidth]{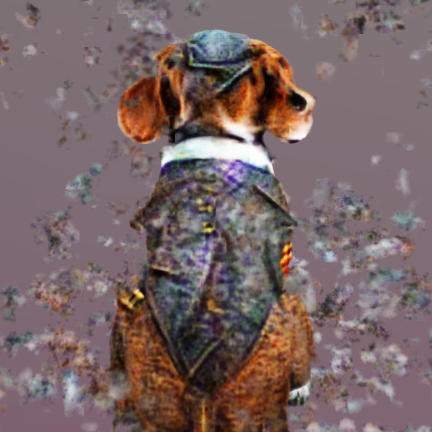} &
        \includegraphics[width=0.16\textwidth]{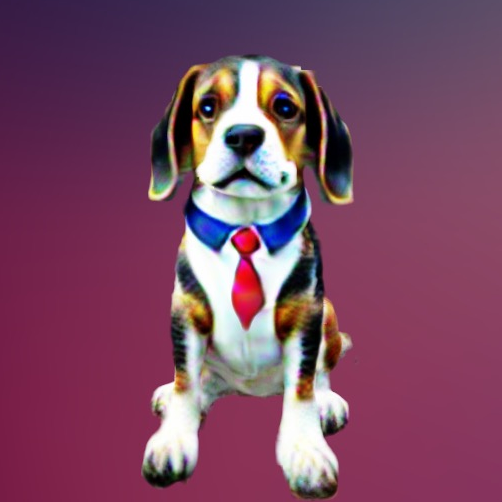} &
        \includegraphics[width=0.16\textwidth]{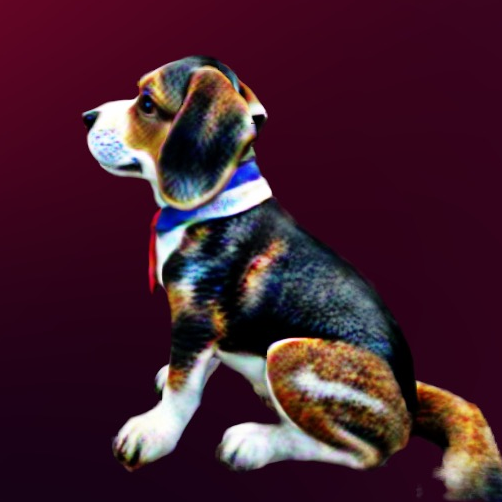} &
        \includegraphics[width=0.16\textwidth]{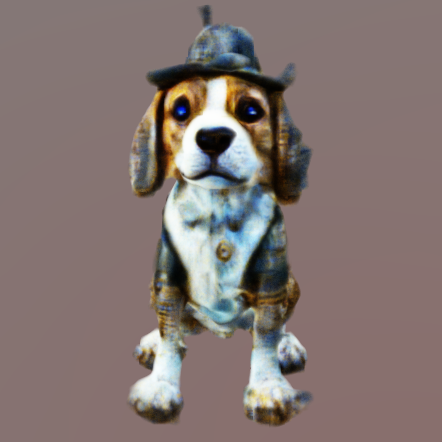} &
        \includegraphics[width=0.16\textwidth]{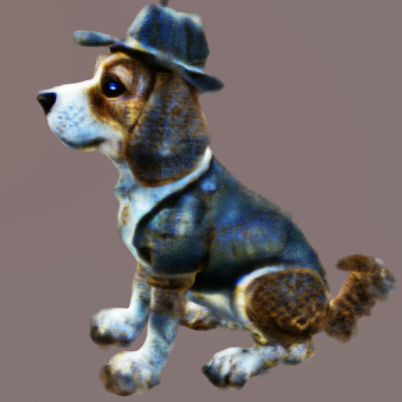} &
        \includegraphics[width=0.16\textwidth]{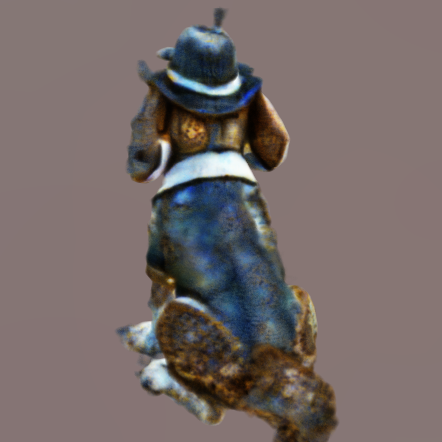} 
        \vspace{-0.1em}
        \\
        \multicolumn{7}{c}{{\prompts{A beagle in a \textcolor{red}{detective’s outfit}}}}
        \vspace{0.2em}
        \\
        \includegraphics[width=0.16\textwidth]{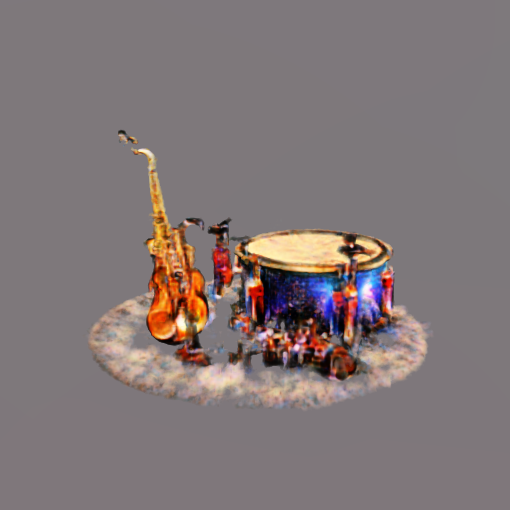} &
        \includegraphics[width=0.16\textwidth]{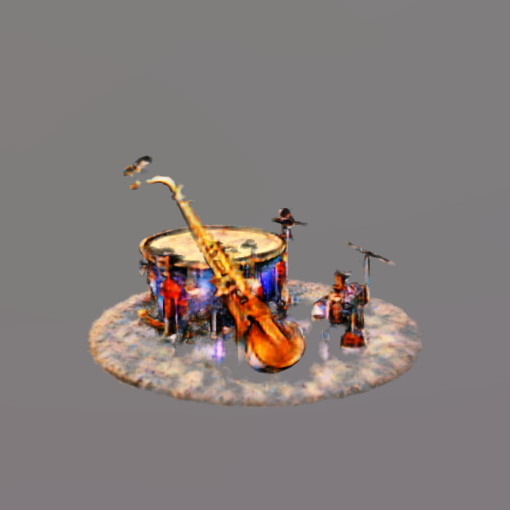} &
        \includegraphics[width=0.16\textwidth]{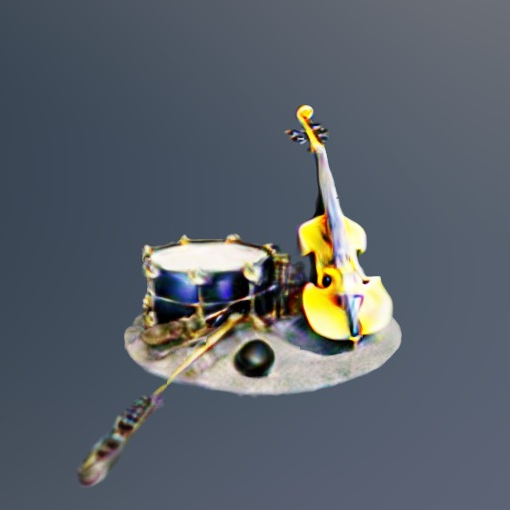} &
        \includegraphics[width=0.16\textwidth]{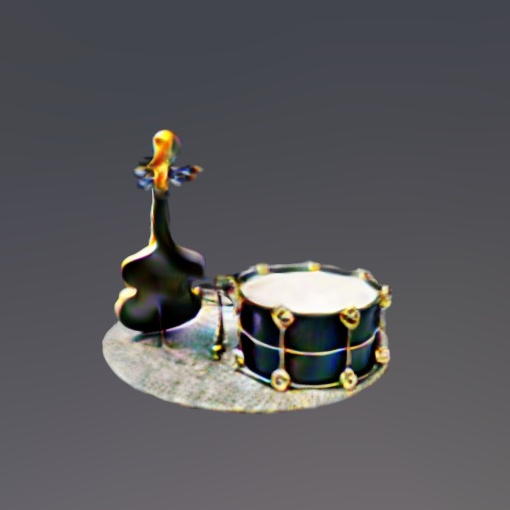} &
        \includegraphics[width=0.16\textwidth]{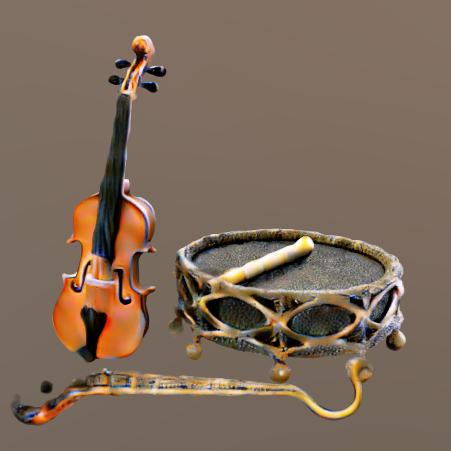} &
        \includegraphics[width=0.16\textwidth]{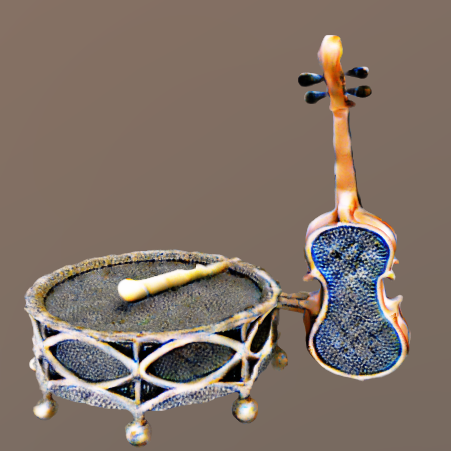} &
        \includegraphics[width=0.16\textwidth]{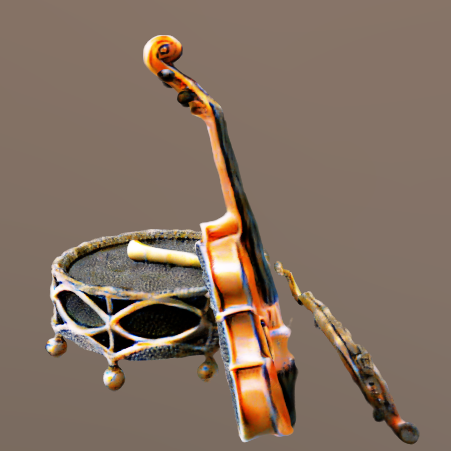} 
        \vspace{-0.1em}
        \\
        \multicolumn{7}{c}{{\prompts{An ensemble of hollow, irregularly shaped musical instruments, including \textcolor{red}{a saxophone, a violin, and a drum}}}} \\
        \multicolumn{7}{c}{{\prompts{resting on a stage before a jazz concert}}}
    \end{tabular}
    \end{tabular}}
    \label{subfig:mesh}
    \vspace{-0.1em}
    \caption{
    \textbf{Comparison of VLM3D with DreamReward~\cite{ye2024dreamreward} and DreamDPO~\cite{zhou2025dreamdpo}.} VLM3D outperforms these methods in semantic fidelity while retaining high perceptual quality. Although DreamReward and DreamDPO enhance texture and detail via differentiable human‐preference rewards or non‐differentiable preference‐guided optimization, they still miss fine‐grained concepts (highlighted in red) that VLM3D accurately captures.
    }
    \label{fig:3d_comparison_with_reward}
     \vspace{-0.3cm}
\end{figure*}

\subsection{Quantitative Evaluation on the GPTEval3D Benchmark}
Table~\ref{tab:quantitative_comparisons} reports performance of VLM3D against baselines on six metrics: Text–Asset Alignment, 3D Plausibility, Texture Details, Geometry Details, Texture–Geometry Coherence, and Overall Score. VLM3D—which integrates explicit VLM rewards with SDS—outperforms all baselines on every metric. In particular, it achieves the top Text–Asset Alignment score, confirming that VLM guidance substantially enhances semantic fidelity. It also leads in 3D Plausibility and perceptual quality (both texture and geometry). Overall, VLM3D’s aggregate score surpasses the strongest baseline by a large margin, demonstrating the complementary benefits of combining SDS and VLM-based feedback.

\vspace{-3mm}
\subsection{Qualitative Comparison with Baseline Methods}
\vspace{-3mm}

Fig.~\ref{fig:3d_comparison} shows example 3D assets generated by seven representative baselines for several prompts. Compared to standard SDS methods and their variants, as shown in Fig.~\ref{fig:3d_comparison}, VLM3D produces cleaner geometry and richer detail, as highlighted by the normal‐map overlays. It also accurately captures subtle descriptive cues—such as ``snow-covered forest'', ``vintage, fragrant perfumes'', ``floating on water'', and ``mysteries of the universe''—marked in red. Similar improvements hold when comparing against recent reward-based methods such as DreamReward and DreamDPO. Although those methods boost perceptual quality, as shown in Fig.~\ref{fig:3d_comparison_with_reward}, they still lag behind in prompt fidelity. In contrast, VLM3D more effectively captures object interactions—rendering actions like ``using a laptop,'' ``spilling out,'' and ``reading the newspaper''—and accurately models rare concepts such as ``kaleidoscope'' and ``detective’s outfit.'' It also excels at complex multi-object scenes, as shown by the jazz concert example with multiple instruments, including ``a saxophone, a violin, and a drum''.

\begin{table*}[t]
  \centering
  \caption{
  \textbf{Quantitative Results on 110 Prompts from the GPTEval3D Benchmark~\cite{wu2024gpt}.} We compute all six GPTEval3D metrics—text alignment, 3D plausibility, texture–geometry coherence, geometry details,  texture details, and overall score—to comprehensively evaluate 3D generation quality. VLM3D achieves the highest score on every metric, demonstrating its superior performance.
  }
  \resizebox{\textwidth}{!}{
    \begin{tabular}{ccccccc}
    \toprule
    \multirow{2}[4]{*}{Method} & \multicolumn{6}{c}{Prompts from GPTEval3D} \\
\cmidrule{2-7}          & Alignment & Plausibility & T-G Coherency. & Geo Details & Tex Details & Overall \\
    \midrule
    DreamFusion\cite{poole2022dreamfusion} & 1000.0  & 1000.0  & 1000.0  & 1000.0  & 1000.0  & 1000.0 \\
    DreamGaussian\cite{tang2023dreamgaussian} & 1100.6 & 953.6 & 1158.6 & 1126.2 & 1130.8 & 951.4 \\
    Fantasia3D\cite{chen2023fantasia3d} & 1067.9 & 891.9 & 1006.0 & 1109.3 & 1027.5 & 933.5 \\
    Instant3D\cite{li2023instant3d} & 1200.0  & 1087.6 & 1152.7 & 1152.0 & 1181.3 & 1097.8 \\
    Latent-NeRF\cite{metzer2022latent} & 1222.3 & 1144.8 & 1156.7 & 1180.5 & 1160.8 & 1178.7 \\
    Magic3D\cite{lin2023magic3d} & 1152.3 & 1000.8 & 1084.4 & 1178.1 & 1084.6 & 961.7 \\
    ProlificDreamer\cite{wang2023prolificdreamer} & 1261.8 & 1058.7 & 1152.0 & 1246.4 & 1180.6 & 1012.5 \\
    MVDream\cite{shi2023mvdream} & 1270.5 & 1147.5 & 1250.6 & 1324.9 & 1255.5 & 1097.7 \\
    DreamReward\footnotemark[1]\cite{ye2024dreamreward} & 1287.5 & 1195.0 & 1254.4 & 1295.5 & 1261.6 & 1193.3 \\
    DreamDPO\cite{zhou2025dreamdpo} & 1298.9 & 1171.9 & 1276.4 & 1373.2 & 1296.9 & 1203.1 \\
    \midrule
    VLM3D (Ours) & \textbf{1365.5} & \textbf{1293.7} & \textbf{1365.4} & \textbf{1419.0} & \textbf{1368.7} & \textbf{1268.6} \\
    \bottomrule
    \end{tabular}%
    }
  \label{tab:quantitative_comparisons}%
\end{table*}%
\footnotetext[1]{Our metrics differ from those reported in the original DreamReward paper because GPT-4V has been deprecated in GPTEval3D, so we instead use GPT-4o-mini.}

\begin{figure*}[h]
    \centering
    \setlength{\tabcolsep}{1pt}
    \setlength{\fboxrule}{1pt}
    \resizebox{0.99\textwidth}{!}{
    \begin{tabular}{c}
    \begin{tabular}{ccc|ccc}
        \multicolumn{1}{c}{{MVDream~\cite{shi2023mvdream}}} &
        \multicolumn{2}{c|}{{\textbf{VLM3D (Ours)}}} &
        \multicolumn{1}{c}{{MVDream~\cite{shi2023mvdream}}} &
        \multicolumn{2}{c}{{\textbf{VLM3D (Ours)}}}
        \\
        \includegraphics[width=0.22\textwidth]{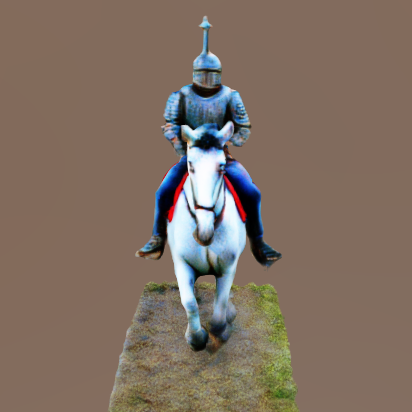} &
        \includegraphics[width=0.22\textwidth]{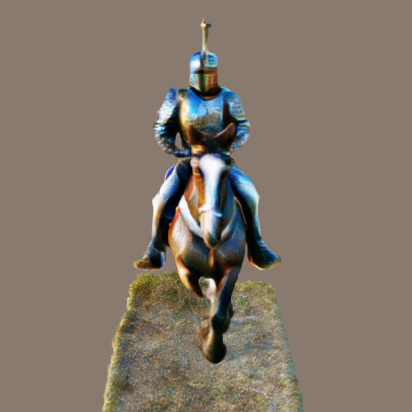} &
        \includegraphics[width=0.22\textwidth]{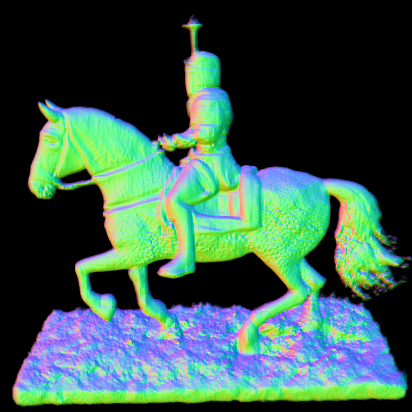} &
        \includegraphics[width=0.22\textwidth]{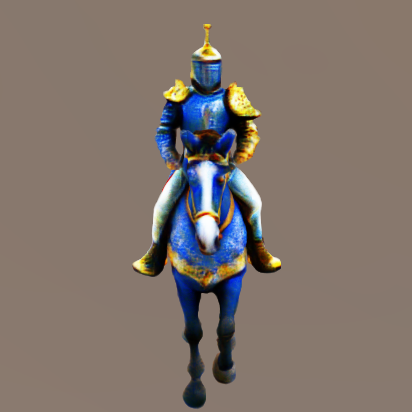} &
        \includegraphics[width=0.22\textwidth]{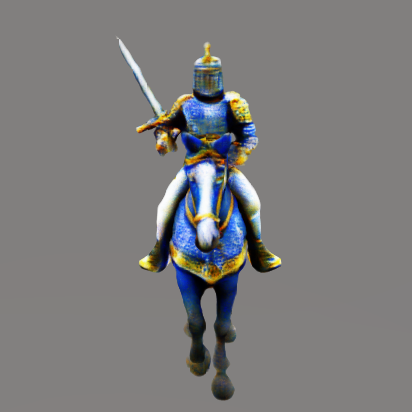} &
        \includegraphics[width=0.22\textwidth]{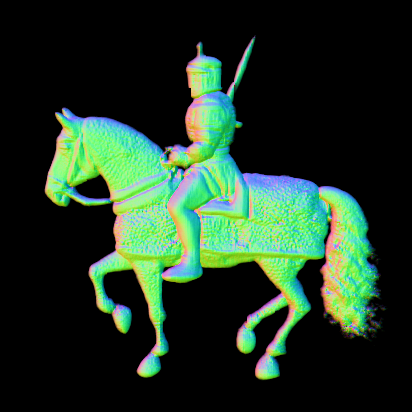} 
        \vspace{-0.1em}
        \\
        \multicolumn{3}{c|}{{\prompts{A knight riding a horse}}} &
        \multicolumn{3}{c}{{\prompts{A knight riding a horse \textcolor{red}{and holding a sword}}}}
        \vspace{0.2em}
        \\
        \includegraphics[width=0.22\textwidth]{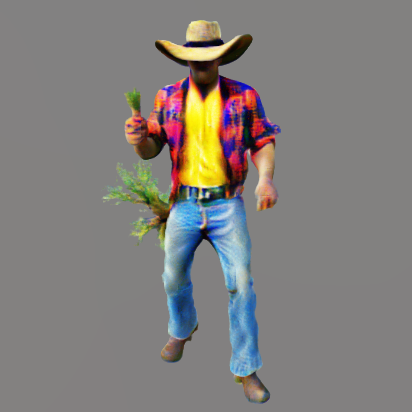} &
        \includegraphics[width=0.22\textwidth]{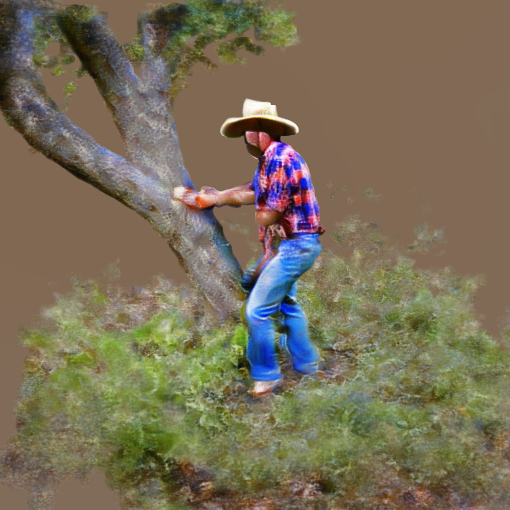} &
        \includegraphics[width=0.22\textwidth]{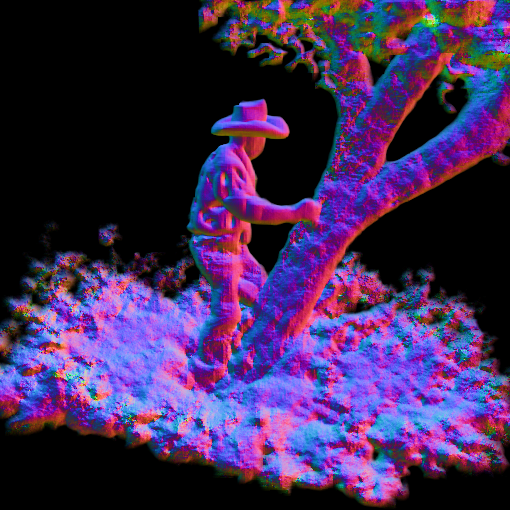} &
        \includegraphics[width=0.22\textwidth]{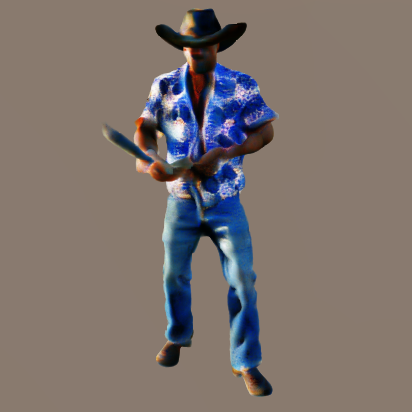} &
        \includegraphics[width=0.22\textwidth]{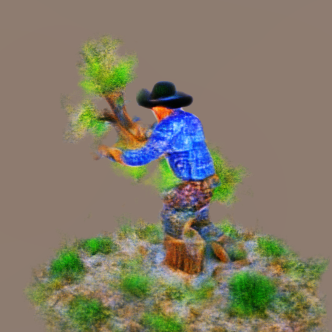} &
        \includegraphics[width=0.22\textwidth]{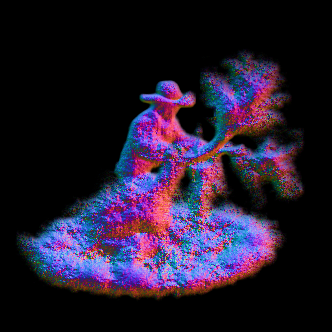}  
        \vspace{-0.1em}
        \\
        \multicolumn{3}{c|}{{\prompts{A man wearing a leather cowboy hat, a colorful Hawaiian shirt}}} &
        \multicolumn{3}{c}{{\prompts{A man wearing a leather cowboy hat, a colorful Hawaiian shirt}}} \\
        \multicolumn{3}{c|}{{\prompts{with bold \textcolor{red}{red} hues, and old jeans is chopping}}} &
        \multicolumn{3}{c}{{\prompts{with bold \textcolor{red}{blue} hues, and old jeans is chopping}}} \\
        \multicolumn{3}{c|}{{\prompts{through \textcolor{red}{the trunk of a tree}, demonstrating strength}}} &
        \multicolumn{3}{c}{{\prompts{through \textcolor{red}{the trunk of a tree}, demonstrating strength}}} \\
        \includegraphics[width=0.22\textwidth]{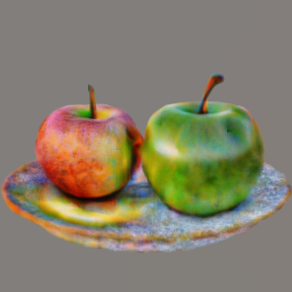} &
        \includegraphics[width=0.22\textwidth]{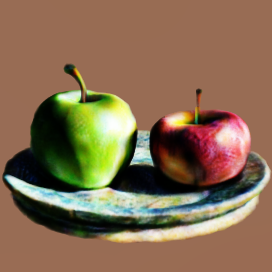} &
        \includegraphics[width=0.22\textwidth]{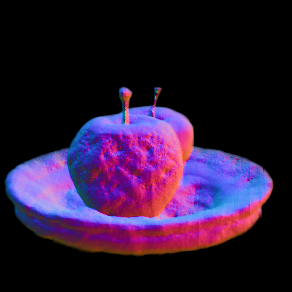} &
        \includegraphics[width=0.22\textwidth]{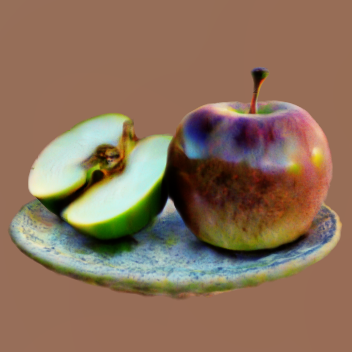} &
        \includegraphics[width=0.22\textwidth]{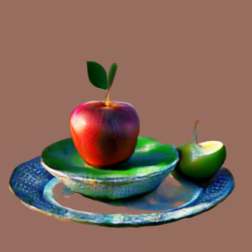} &
        \includegraphics[width=0.22\textwidth]{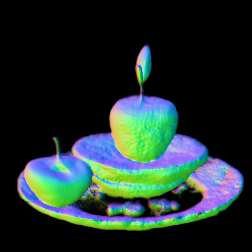} 
        \vspace{-0.1em}
        \\
        \multicolumn{3}{c|}{{\prompts{A red apple and a green apple \textcolor{red}{in} a plate}}} &
        \multicolumn{3}{c}{{\prompts{A red apple is \textcolor{red}{inside} a plate and a green apple is \textcolor{red}{beside} the plate}}}
        \\
    \end{tabular}
    \end{tabular}}
    \label{subfig:mesh}
    \vspace{-0.1em}
    \caption{
    \textbf{Sensitivity Analysis to Text Perturbations.} We compare VLM3D and MVDream on pairs of prompts that differ by a single concept (highlighted in red). VLM3D accurately adds or removes objects (first row), changes clothing color (second row), and updates spatial relations (third row), demonstrating its better semantic understanding than baselines.
    }
    \label{fig:analysis_textalignment}
    \vspace{-0.5em}
\end{figure*}

\subsection{More Analyses and Ablation Studies}
In this section, we dissect VLM3D’s behavior and quantify how key design choices affect text-to-3D generation performance.
\paragraph{Sensitivity to Prompt Perturbations} 
We assess VLM3D’s semantic fidelity by comparing it to MVDream on prompts that differ by a single concept. In the first experiment of Fig.~\ref{fig:analysis_textalignment}, we issue ``a knight holding a sword'' versus ``a knight not holding a sword.'' VLM3D precisely follows the instruction—adding or removing the sword—while MVDream always omits it. Next, we modify a farmer scene by changing the clothing color and specifying ``chopping through a tree trunk.'' VLM3D accurately reflects both the new color and the action, whereas MVDream fails to capture details of the action. Remarkably, VLM3D also exhibits superior spatial reasoning: when prompted to place a green apple ``in'' versus ``beside'' a plate, VLM3D honors each spatial relation, but MVDream places the apple inside the plate in both cases. These results demonstrate VLM3D’s strong semantic alignment and its sensitivity to even subtle prompt variations.

\begin{figure}[t]
    \centering
    \setlength{\tabcolsep}{1pt}
    \setlength{\fboxrule}{1pt}
    \resizebox{0.99\textwidth}{!}{
    \begin{tabular}{c}
    \begin{tabular}{cccccccc}
        & \multicolumn{1}{c}{{SDS~\cite{poole2022dreamfusion}}} &
        \multicolumn{2}{c}{{VLM3D (w/o Geometry Query)}} & 
        \multicolumn{2}{c}{{VLM3D (w/ Single View Input)}} &
        \multicolumn{2}{c}{{\textbf{VLM3D*}}}
        \\
        \begin{turn}{90} \,\,\small{Stable Diffusion~\cite{rombach2022high}} \end{turn} &
        \includegraphics[width=0.22\textwidth]{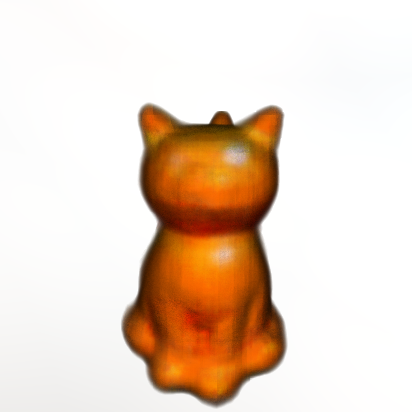} &
        \includegraphics[width=0.22\textwidth]{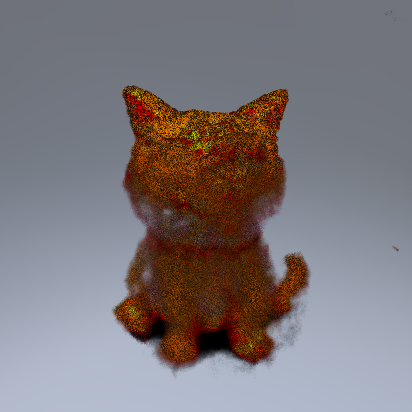} &
        \includegraphics[width=0.22\textwidth]{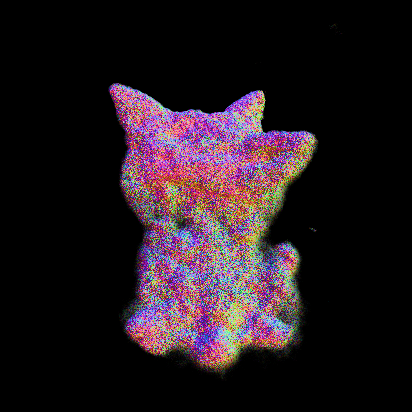} &
        \includegraphics[width=0.22\textwidth]{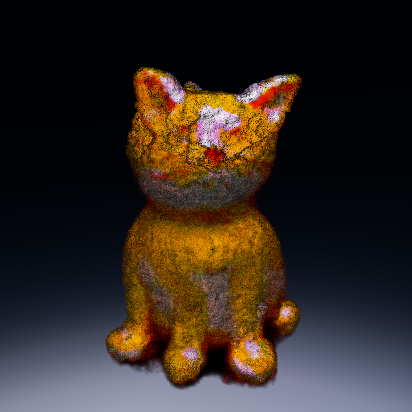} &
        \includegraphics[width=0.22\textwidth]{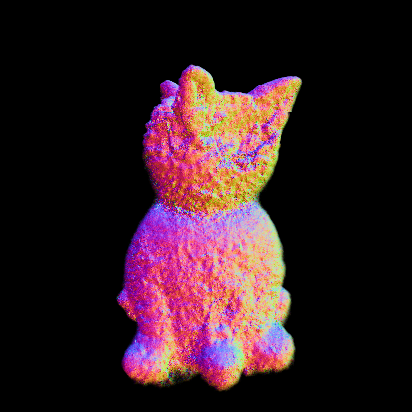} &
        \includegraphics[width=0.22\textwidth]{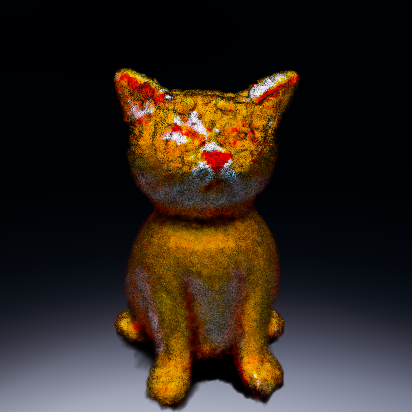} &
        \includegraphics[width=0.22\textwidth]{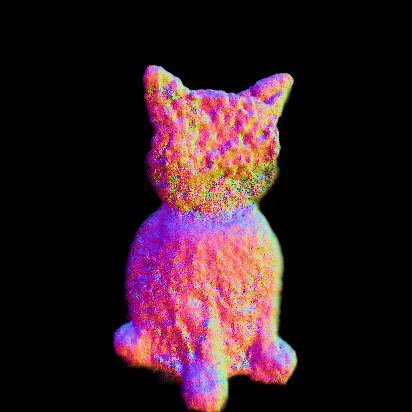} 
        \vspace{-0.1em}
        \\
        \multicolumn{8}{c}{{\prompts{An orange tabby cat shaped cookie jar}}}
        \\
        \begin{turn}{90} \,\,\,\,\,\small{MVDiffusion~\cite{shi2023mvdream}} \end{turn} &
        \includegraphics[width=0.22\textwidth]{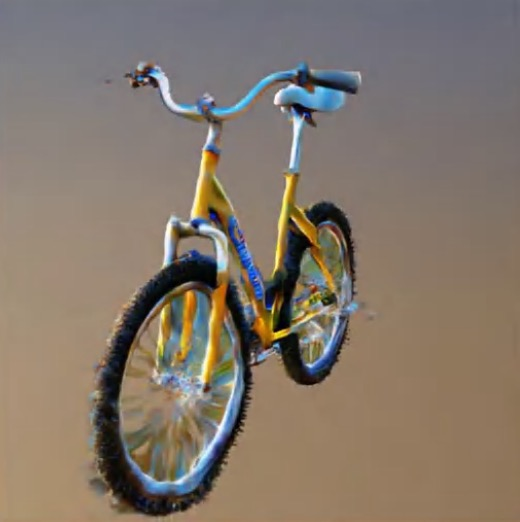} &
        \includegraphics[width=0.22\textwidth]{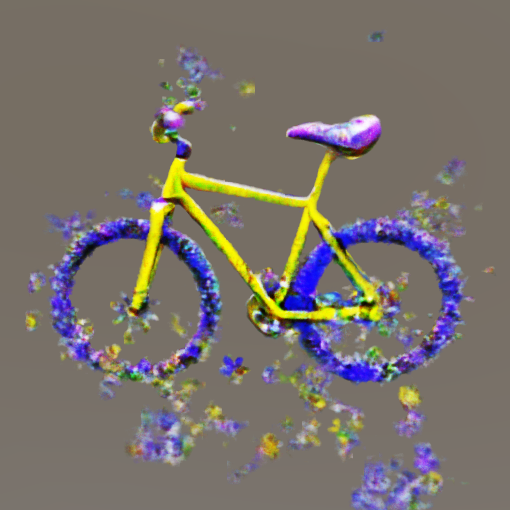} &
        \includegraphics[width=0.22\textwidth]{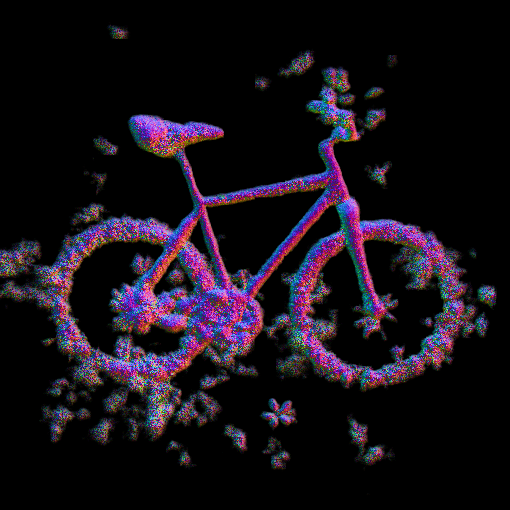} &
        \includegraphics[width=0.22\textwidth]{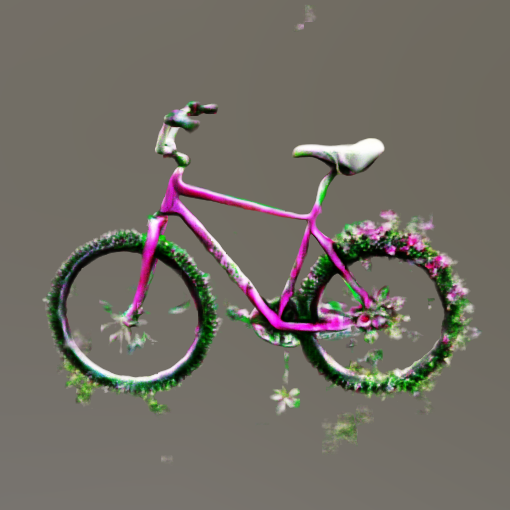} &
        \includegraphics[width=0.22\textwidth]{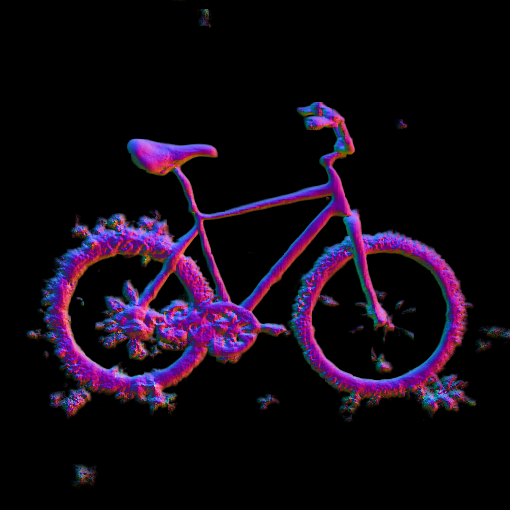} &
        \includegraphics[width=0.22\textwidth]{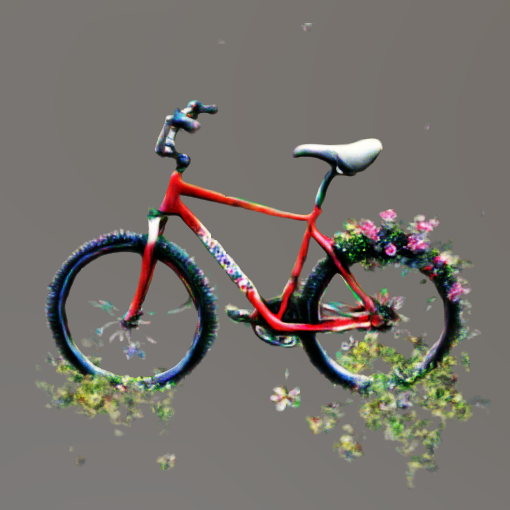} &
        \includegraphics[width=0.22\textwidth]{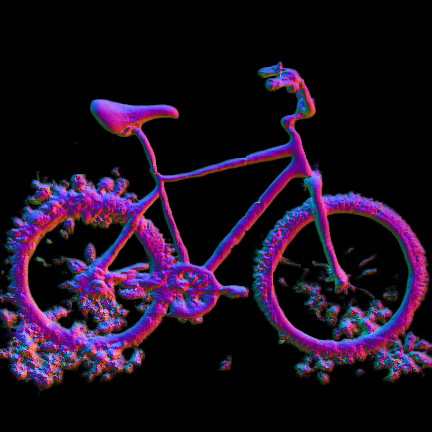} 
        \vspace{-0.1em}
        \\
        \multicolumn{8}{c}{{\prompts{A bicycle that leaves a trail of flowers}}}
    \end{tabular}
    \end{tabular}}
    \label{subfig:mesh}
    \vspace{-0.1em}
    \caption{
    \textbf{Ablation of Geometric Query and Multi-View Input.} We assess the impact of (a) removing the explicit geometry-consistency query from the VLM prompt and (b) using a single view instead of multi-view images. Omitting either component degrades 3D quality—leading to Janus-face artifacts, floating parts, and fractured surfaces. Each row uses a different diffusion backbone: the top employs Stable Diffusion 2-1 \cite{rombach2022high}, while the bottom uses MVDream \cite{shi2023mvdream}.
    }
    \label{fig:ablation}
     \vspace{-0.3cm}
\end{figure}

\paragraph{Ablation: Geometric Query in VLM Prompt}  
We evaluate the effect of including an explicit query for geometric consistency and quality in the VLM prompt (see Sec.~\ref{sec:conclusion}). Without this query, VLM3D exhibits classic multi‐face (Janus) artifacts on a cat prompt, and generates bicycles with floating parts and fractured surfaces (Fig.~\ref{fig:ablation}). Including the geometric query substantially mitigates these errors, confirming that explicit geometric guidance is critical.

\paragraph{Ablation: Multi‐View vs.\ Single‐View Inputs}  
We further test whether multiple views are necessary for spatial coherence. When the VLM receives only a single rendered image instead of the full view set, VLM3D again suffers from Janus artifacts (Fig.~\ref{fig:ablation}). This ablation underscores the importance of multi‐view input in enforcing 3D consistency.

\section{Conclusion}
\label{sec:conclusion}
In this paper, we presented VLM3D, a novel framework that injects large VLMs as explicit, differentiable semantic and spatial rewards into the SDS pipeline for text-to-3D generation. By designing dual yes/no queries—one for content fidelity and one for geometric constraints—and backpropagating the resulting VLM log-odds reward, VLM3D achieves precise alignment to complex prompts, resolves view-consistency artifacts, and produces high-fidelity textures and geometry. We demonstrated these gains across a diverse suite of baselines on the GPTEval3D benchmark, achieving state-of-the-art improvements in semantic alignment, 3D plausibility, and texture–geometry coherence.

\paragraph{Limitations and Future Directions}  
Despite these advances, VLM3D still struggles with very long or highly detailed prompts. For instance (Fig.~\ref{fig:teaser}), when tasked with generating the renowned “Embracing Peace” statue in San Diego purely from its text description, VLM3D correctly reproduces the kissing pose and dual figures with correct clothing—outperforming MVDream, which omits the nurse entirely—but still misses finer details such as the nurse’s lifted leg and outstretched arm. Interestingly, a standalone VLM can readily describe these details when presented with a photograph of the statue, indicating that our current reward formulation does not fully leverage the VLM’s rich visual understanding.

Building on these insights, we see several promising paths to further advance large VLMs for text‐conditioned 3D generation. First, disentangling semantic and geometric feedback—separate VLM heads for content and geometry queries—might provide finer control over prompt fidelity and 3D consistency. Second, advanced prompt engineering, such as hierarchical or detail‐focused cues, can ensure the VLM reward faithfully captures nuanced attributes and fine‐grained descriptions. 

\bibliographystyle{IEEEtran}
\bibliography{neurips_2025}

\newpage
\appendix


\section{Implementation Details}
\label{appendix-implementation-details}
In this section, we first provide a detailed implementation of our proposed method (Section~\ref{sec:appendix-details-vlm3d}), followed by implementation details for the baseline methods (Section~\ref{sec:appendix-details-baselines}).
To facilitate evaluation and reproducibility, we provide a compressed package containing the source code and videos in the attachment.
We will also publish in a public repository.

\subsection{VLM3D}
\label{sec:appendix-details-vlm3d}

\paragraph{Pseudo-code for VLM3D}
A detailed pseudo-code for VLM3D is presented in algorithm~\ref{alg:vlm3d}.

\paragraph{Hyperparameters}
The optimization of the 3D representations, NeRFs in this work, typically takes $10,000$ to $15,000$ update steps, with a learning rate set to $1 \times 10^{-3}$. In our text-to-image diffusion model, we employ a classifier-free guidance (CFG) loss with a scale of $50$. While common sparsity and opacity regularizations are omitted, we retain orientation regularization, annealing its weight from $10$ to $1000$ over the initial $5000$ training iterations.

Among the hyperparameters, the weight assigned to the Vision-Language Model (VLM) reward is particularly critical. As detailed in the manuscript, this weight is generally annealed from a high initial value to a lower one during training; here, we elaborate on this annealing strategy. For challenging prompts---those characterized by significant length, multiple objects, complex spatial relationships, or infrequent terminology---we set the initial VLM reward weight to a value within the range of $[300, 800]$. 
This higher initial weight empowers the VLM to comprehensively interpret these intricate prompts, thereby establishing a robust foundation for generation. 
Conversely, for simpler prompts that are readily processed by standard Stable Diffusion-based methods, such a high initial weight is not strictly necessary, although it remains effective. 
Overall, this annealing schedule for the VLM reward weight consistently yields 3D assets with high-quality geometry and strong prompt alignment, steadily outperforming SDS-based methods, especially on more demanding prompts.

\paragraph{Backbones of Diffusion Models and VLMs}
For our text-to-image diffusion backbone, we evaluated two models: Stable Diffusion v2.1~\cite{rombach2022high} and the fine-tuned Stable Diffusion v2.1-base-4view~\cite{shi2023mvdream}. The latter demonstrated superior robustness and capability in 3D generation tasks, and was therefore adopted as our primary diffusion model. Similarly, for the Vision-Language Model (VLM) component, we assessed three prominent open-source options: Qwen2.5-VL (7B)~\cite{bai2025qwen2_5vl}, PaliGemma (3B)~\cite{beyer2024paligemma}, and IDEFICS. Our preliminary analysis indicated that Qwen2.5-VL (7B) delivered the most compelling performance, leading to its selection as our VLM backbone.



\paragraph{Differentiable Image Processors for VLMs}
VLM3D shares foundational principles with Reinforcement Learning from Human Feedback (RLHF)~\cite{christiano2017deep, ouyang2022training}, a paradigm widely recognized for its power in aligning models with reward models.
This perceived superiority of RLHF typically stems from its use of explicit reward modeling and its inherent capacity for exploration during the differentiable learning process.

Despite its potential, the practical application of RLHF or similar gradient-based reward optimization for VLMs encounters significant obstacles. 
Even among prominent open-source VLMs, a common architectural design involves various image preprocessors that detach the gradient flow through the visual components. 
\textit{We identify that a key challenge in enabling end-to-end differentiability through VLMs is that their image preprocessors typically detach gradients to accommodate diverse input formats, often by converting image data to NumPy arrays for intermediate processing steps.} 
This detachment cuts off the pathway for backpropagating reward signals derived from the VLM, a critical requirement for end-to-end gradient-based optimization techniques that leverage VLM feedback.

\textbf{A core technical contribution of this work is the establishment of a fully differentiable pathway through the VLM, which is crucial for our VLM3D framework.} 
To overcome the detachment issue in the image processors, we have re-engineered this module by redesigning its internal operations to exclusively utilize Torch tensors, thereby maintaining an uninterrupted gradient flow.
By ensuring continuous gradient flow, we enable an end-to-end differentiable forward process. 
This crucial modification allows the VLM's rich semantic and spatial understanding to be translated into well-defined, differentiable reward signals, directly informing and refining the generation of 3D assets. 
We have attached the re-engineered preprocessor code in the supplementary material.

\subsection{Baselines}
\label{sec:appendix-details-baselines}

\paragraph{MVDream~\cite{shi2023mvdream}}
For the MVDream baseline, we utilize the official codebase provided by~\cite{shi2023mvdream}. The text-to-image diffusion model employed is the fine-tuned Stable Diffusion v2.1-base-4view, also introduced in~\cite{shi2023mvdream}. We adhere to the original hyperparameter configurations.

\paragraph{DreamReward~\cite{ye2024dreamreward}}
Our implementation of DreamReward~\cite{ye2024dreamreward} is based on the official source code. We employ Stable Diffusion v2.1~\cite{rombach2022high} as the text-to-image diffusion backbone, complemented by the official Reward3D Scorer~\footnote{The Reward3D weights can be found at \url{https://huggingface.co/yejunliang23/Reward3D}} serving as the 3D reward model. All hyperparameters are kept consistent with the original implementation.

\paragraph{DreamDPO~\cite{zhou2025dreamdpo}}
Considering that the official code for DreamDPO~\cite{zhou2025dreamdpo} has not yet been publicly released, we directly report the results presented in the original paper for comparative analysis.

\paragraph{Others}
For results of other established methods, including DreamFusion~\cite{poole2022dreamfusion}, DreamGaussian~\cite{tang2023dreamgaussian}, Latent-NeRF~\cite{metzer2022latent}, Magic3D~\cite{lin2023magic3d}, and ProlificDreamer~\cite{wang2023prolificdreamer}, our experiments are primarily derived from the threestudio project~\footnote{The threestudio can be found at \url{https://github.com/threestudio-project/threestudio}}. Some results were also sourced from the GPTEval3D repository~\footnote{The benchmark GPTEval3D can be found at \url{https://github.com/3DTopia/GPTEval3D}}.


\begin{algorithm}[t]
\caption{Pseudo-code for VLM3D}
\label{alg:vlm3d}
\begin{algorithmic}[1]
\State \textbf{Input:} Text-to-image diffusion model $s_{\phi}$, VLM $Q(\cdot, \cdot)$, Text prompt $y$, Number of views $N$
\State \textbf{Input:} Learning rate $\eta$ for 3D representation $\theta$, Annealing schedule $\lambda_{VLM}$
\State
\State \textbf{Initialization:} Random A 3D representation (e.g., NeRF) parameterized by $\theta_0$
\State
\While{not converged}
    \State Sample $N$ camera viewpoints $\{v_i\}_{i=1}^N$
    \State Render $N$ images $\mathcal{X} = \{x_i\}_{i=1}^N = \{I(\theta, v_i)\}_{i=1}^N$ 
    \State
    \State {\textbf{VLM Reward}}
    \State \hspace{\algorithmicindent} Query VLM: $Q(y, \mathcal{X}) \mapsto \{\text{Yes}, \text{No}\}$ using the designed prompt
    \State \hspace{\algorithmicindent} Extract "Yes" ($z_{yes}$) and "No" ($z_{no}$) logits from VLM
    \State \hspace{\algorithmicindent} Calculate VLM reward: $r_{VLM} = z_{yes} - z_{no}$
    \State
    \State {\textbf{SDS Loss}}
    \State \hspace{\algorithmicindent} Sample timestep $t \sim \text{Uniform}(0, T)$ and noise $\epsilon \sim \mathcal{N}(0, I)$
    \State \hspace{\algorithmicindent} Compute noisy image $x_t = x_0 + \sigma(t)\epsilon$
    \State \hspace{\algorithmicindent} Estimate score $s_{\phi}(x_t, y, t)$
    \State \hspace{\algorithmicindent} SDS gradient: $\nabla_{\theta}\mathcal{L}_{SDS} \approx w(t)(s_{\phi}(x_t, y, t) + \frac{\epsilon}{\sigma(t)}) \frac{\partial I(\theta, v_i)}{\partial \theta}$
    \State
    \State {\textbf{Total Loss and Parameter Update}}
    \State \hspace{\algorithmicindent} Retrieve current VLM weight $\lambda_{VLM}$ from annealing schedule
    \State \hspace{\algorithmicindent} Compute total loss gradient: $\nabla_{\theta}\mathcal{L}_{total} = \nabla_{\theta}\mathcal{L}_{SDS} - \lambda_{VLM} \nabla_{\theta}r_{VLM}$
    \State \hspace{\algorithmicindent} Update parameters: $\theta \leftarrow \theta - \eta \nabla_{\theta}\mathcal{L}_{total}$.
\EndWhile
\end{algorithmic}
\end{algorithm}

\section{Details of VLM prompt design}

The design of the prompt provided to the Vision-Language Model (VLM) is a critical factor influencing the quality and relevance of the VLM reward for 3D generation. 
To identify the most effective prompt for our VLM3D framework, we experimented with various prompt formulations. 
Below, we present three representative examples from our exploration, including our finally selected prompt, and discuss the insights behind our choice.

\begin{tcolorbox}[
    enhanced,
    colback=Gray!20, 
    colframe=Gray,    
    coltitle=Black,
    fonttitle=\bfseries\sffamily,
    title=Example Prompt A (Focus: Content Alignment Only),
    breakable,
    boxsep=1mm,
]
Carefully evaluate the provided images, which show multiple views of a single 3D object. Does the underlying 3D object, considering all views together, correspond to the description: '\texttt{[text\_description\_here]}'? \par
Strictly respond with only 'Yes' or 'No'.
\end{tcolorbox}
\vspace{0.5cm}

\begin{tcolorbox}[
    enhanced,
    colback=Gray!20,
    colframe=Gray,
    coltitle=Black,
    fonttitle=\bfseries\sffamily,
    title=Example Prompt B (Focus: Simplified Criteria),
    breakable,
    boxsep=1mm,
]
Carefully evaluate the 3D object shown in the multiple input image views based on the following criteria:
\begin{enumerate}[nosep, leftmargin=*] 
    \item \textbf{Content Match:} Does the object strongly match the text description: \texttt{[text\_description\_here]}?
    \item \textbf{Visual Plausibility:} Does the object appear visually coherent and plausible across all views, without major jarring artifacts or inconsistencies?
\end{enumerate}
Considering both criteria, answer 'Yes' if both are reasonably met. Otherwise, answer 'No'.
\end{tcolorbox}
\vspace{0.5cm}

\begin{tcolorbox}[
    enhanced,
    colback=Green!25,
    colframe=ForestGreen,
    coltitle=White,
    fonttitle=\bfseries\sffamily,
    title=Selected VLM Prompt (Our Optimal Formulation),
    breakable,
    boxsep=1mm,
]

Carefully evaluate the provided images, which show multiple views of a single 3D object. Does the underlying 3D object, considering all views together, meet all of the following criteria simultaneously?
\begin{enumerate}[nosep, leftmargin=*]
    \item \textbf{Content Match:} The object corresponds to the description: \texttt{[text\_description\_here]}.
    \item \textbf{Geometric Quality:} Based on all views combined, the object appears geometrically sound and consistent. There are no visible signs of major flaws such as multiple faces on one part (Janus-faced issue), broken surfaces, intersecting geometry, or highly unrealistic polygonal facets when considering the object from these different perspectives.
\end{enumerate}
Strictly respond with only 'Yes' or 'No'.
\end{tcolorbox}

\paragraph{Discussion of Prompt Selection}
Our empirical evaluations demonstrated that the "Selected VLM Prompt (Our Optimal Formulation)" yielded the best results for 3D generation. This prompt's effectiveness stems from several key attributes when compared to other variants, such as "Example Prompt A" and "Example Prompt B".

\begin{figure}[t]
    \centering
    \setlength{\tabcolsep}{1pt}
    \setlength{\fboxrule}{1pt}
    \resizebox{0.99\textwidth}{!}{
    \begin{tabular}{c}
    \begin{tabular}{c|ccc}
        MVDream~\cite{shi2023mvdream} & VLM3D (w/ Prompt A) & VLM3D (w/ Prompt B) & VLM3D (w/ Optimal Prompt) \\
        \begin{overpic}[width=0.32\linewidth]{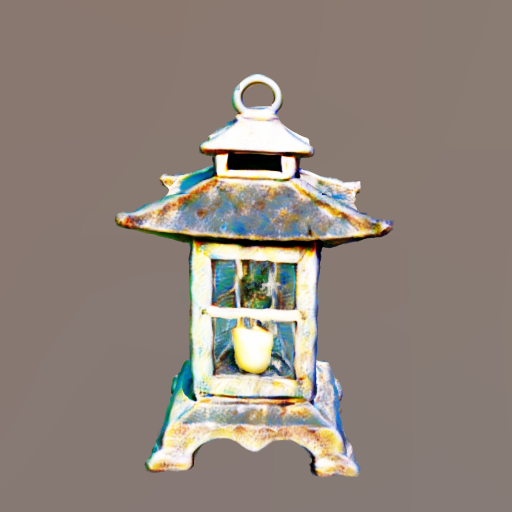}
	   \put(68,68){{\includegraphics[width=0.1\linewidth]{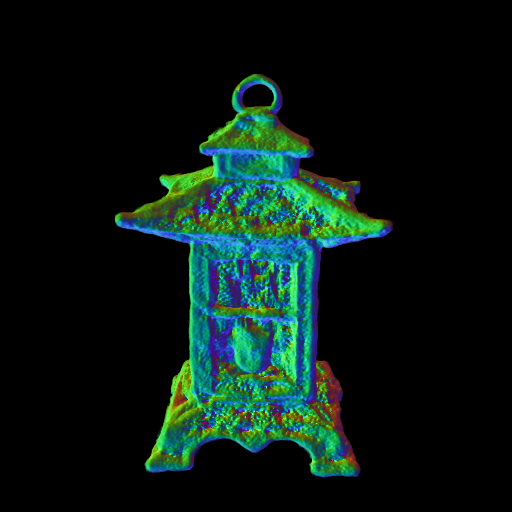}}}
	\end{overpic} &
         \begin{overpic}[width=0.32\linewidth]{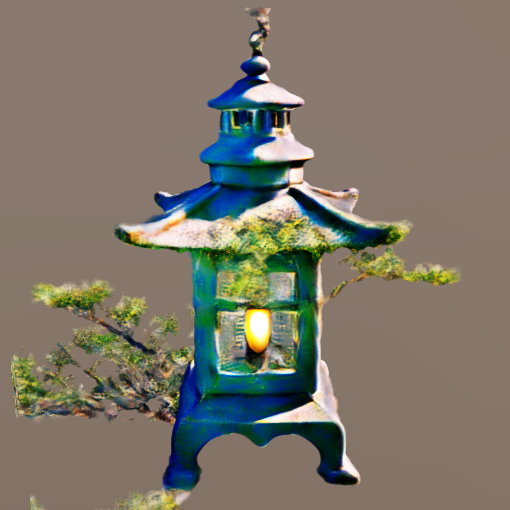}
	   \put(68,68){{\includegraphics[width=0.1\linewidth]{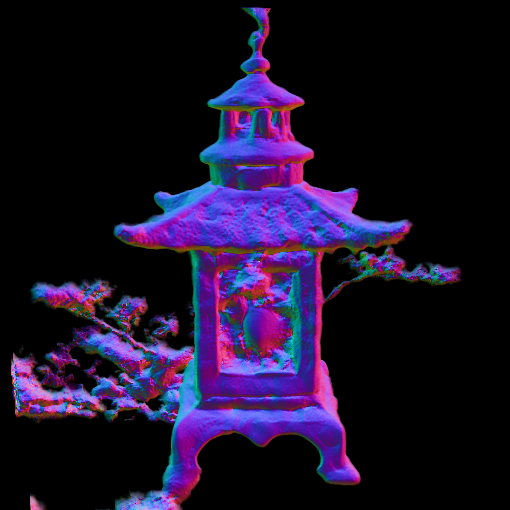}}}
	\end{overpic} &
        \begin{overpic}[width=0.32\linewidth]{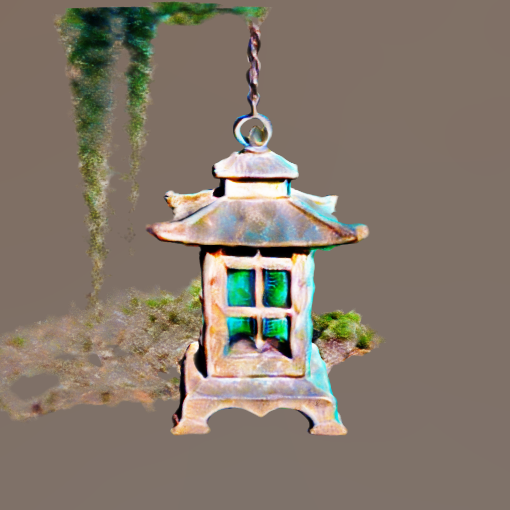}
	   \put(68,68){{\includegraphics[width=0.1\linewidth]{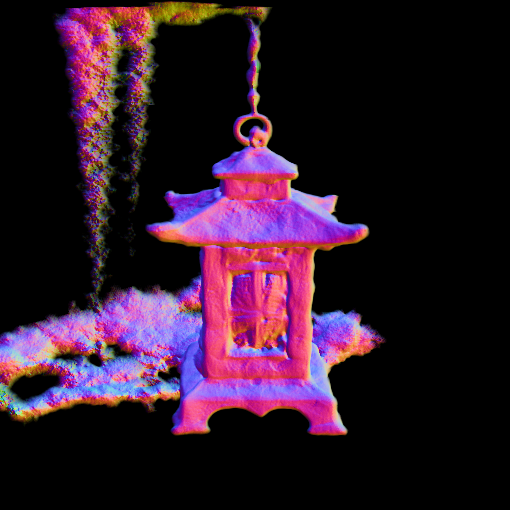}}}
	\end{overpic} &
        \begin{overpic}[width=0.32\linewidth]{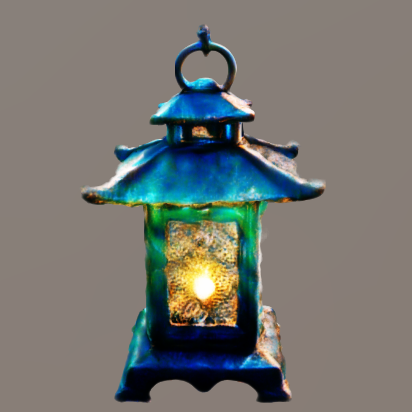}
						\put(68,68){{\includegraphics[width=0.1\linewidth]{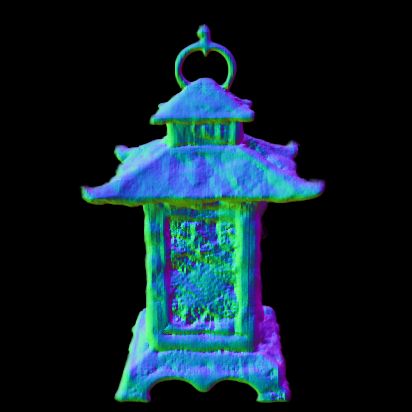}}}
					\end{overpic}
        \vspace{-0.1em}
        \\
        \multicolumn{4}{c}{{\prompts{An old, layered, asymmetrical lantern, with a \textcolor{red}{patina copper finish} and translucent panes that flicker with}}} \\
        \multicolumn{4}{c}{{\prompts{\textcolor{red}{bioluminescent light} from \textcolor{red}{cultured algae within}}}}
    \end{tabular}
    \end{tabular}}
    \vspace{-0.1em}
    \caption{
    \textbf{Effect of different VLM prompt designs.} 
    }
    \label{fig:appendix-compare-prompts}
     \vspace{-0.3cm}
\end{figure}

The chosen prompt excels due to its comprehensive yet clear criteria. It explicitly requires the VLM to assess both {Content Match} and  {Geometric Quality} simultaneously, considering all views collectively. 
This dual-query structure is crucial, as focusing solely on content alignment, like in "Example Prompt A," often leads to 3D assets that match the text semantically but suffer from significant geometric flaws. 
The inclusion of an explicit geometric quality check, as detailed in our selected prompt and supported by ablation studies (see Fig.~\ref{fig:ablation} and Fig.~\ref{fig:appendix-compare-prompts}), is critical for ensuring 3D consistency and plausibility.

Furthermore, the "Geometric Quality" criterion in our selected prompt is highly specific, enumerating common failure modes like "multiple faces on one part, broken surfaces, intersecting geometry, or highly unrealistic polygonal facets." This level of detail provides clearer guidance to the VLM compared to more abstract phrasing. For instance, "Example Prompt B" uses the term "Visual Plausibility" and asks if the criteria are "reasonably met." While aiming for a similar goal, "Visual Plausibility" can be more subjective and may not as effectively penalize specific 3D inconsistencies as the detailed "Geometric Quality" checklist. 
The explicit instruction to consider the object "from these different perspectives" also reinforces the need for multi-view consistency.

Finally, the unambiguous instruction to "Strictly respond with only 'Yes' or 'No'" ensures a clean, binary signal for reward calculation, simplifying the integration of VLM feedback into our optimization pipeline. 
In contrast, prompts that might give less constrained responses could complicate the derivation of a differentiable reward. 

{The balance of comprehensiveness, specificity in defining undesirable artifacts, and a clear output format made our selected prompt the most robust and effective choice among the candidates evaluated.}

\section{Additional Results}
\label{appendix-additional-results}
\begin{figure}[t]
    \centering
    \setlength{\tabcolsep}{1pt}
    \setlength{\fboxrule}{1pt}
    \resizebox{0.99\textwidth}{!}{
    \begin{tabular}{c}
    \begin{tabular}{ccccccc}
        \begin{turn}{90} \,\,\,\,\,\,\,\,\,\small{MVDream~\cite{shi2023mvdream}} \end{turn} &
        \includegraphics[width=0.22\textwidth]{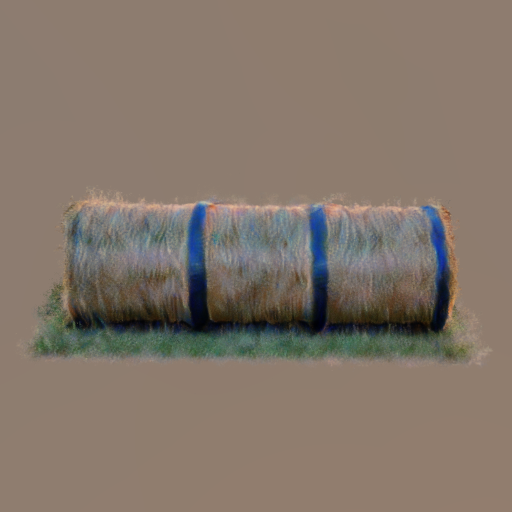} &
        \includegraphics[width=0.22\textwidth]{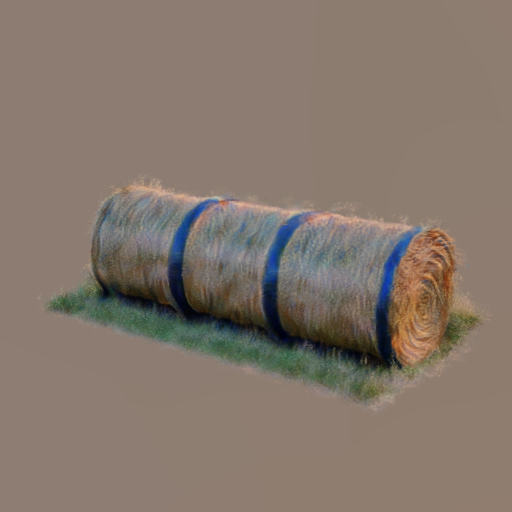} &
        \includegraphics[width=0.22\textwidth]{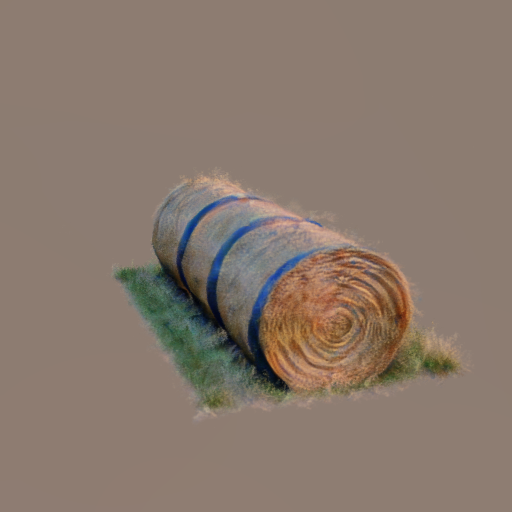} &
        \includegraphics[width=0.22\textwidth]{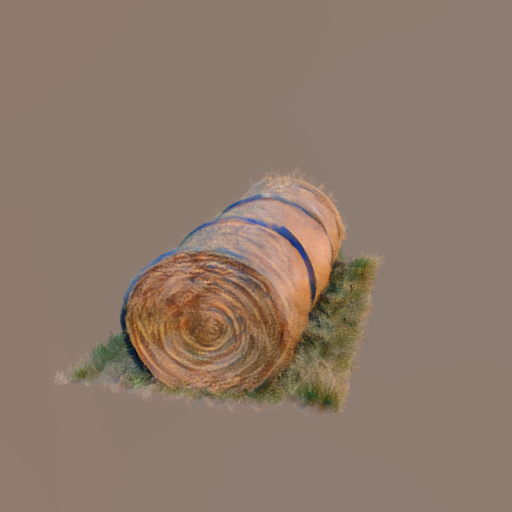} &
        \includegraphics[width=0.22\textwidth]{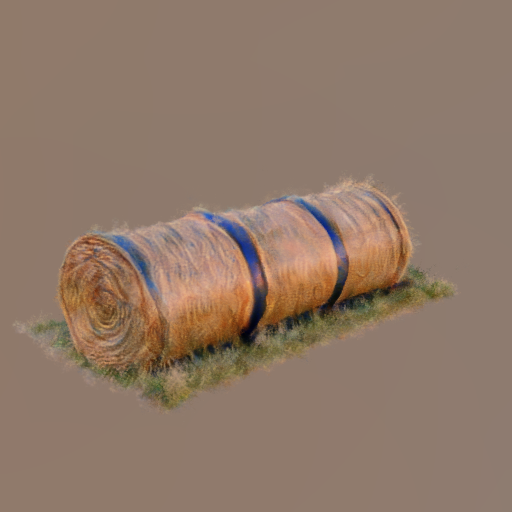} &
        \includegraphics[width=0.22\textwidth]{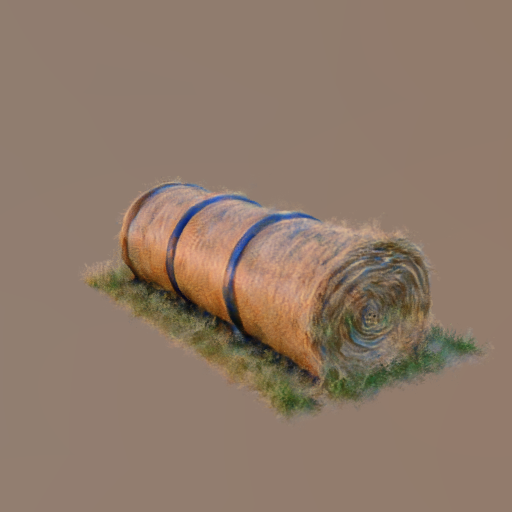} 
        \vspace{-0.1em}
        \\
        \begin{turn}{90} \,\,\,\,\,\,\,\,\small{\textbf{VLM3D (ours)}} \end{turn} &
        \includegraphics[width=0.22\textwidth]{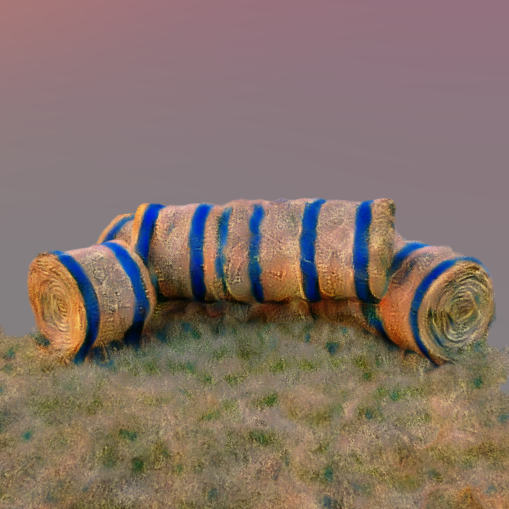} &
        \includegraphics[width=0.22\textwidth]{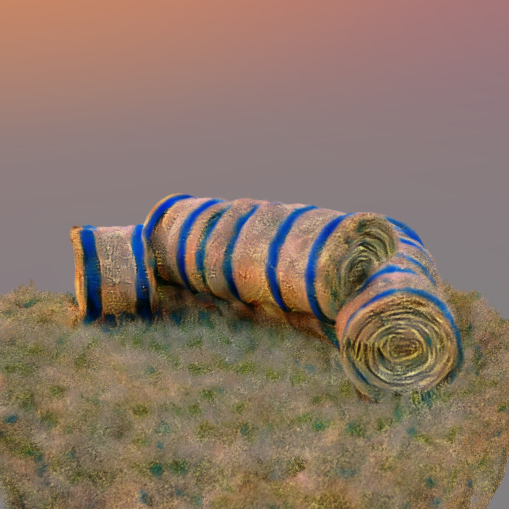} &
        \includegraphics[width=0.22\textwidth]{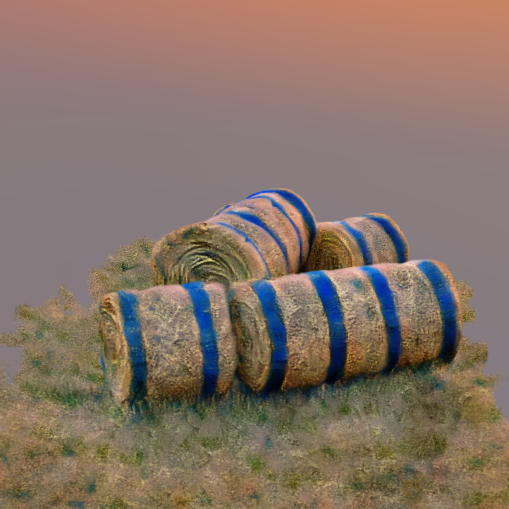} &
        \includegraphics[width=0.22\textwidth]{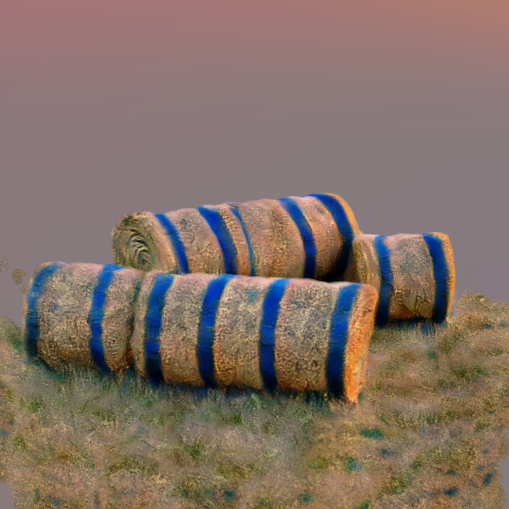} &
        \includegraphics[width=0.22\textwidth]{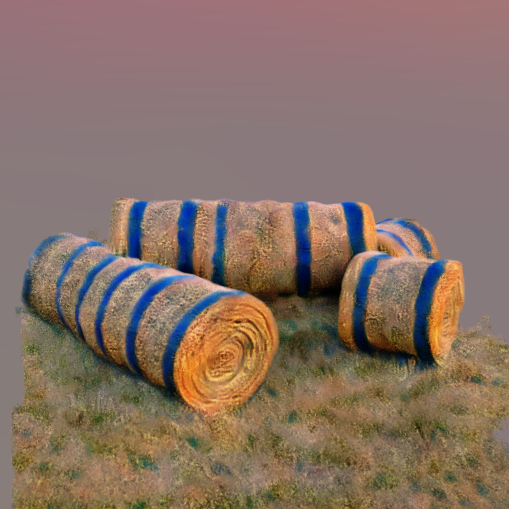} &
        \includegraphics[width=0.22\textwidth]{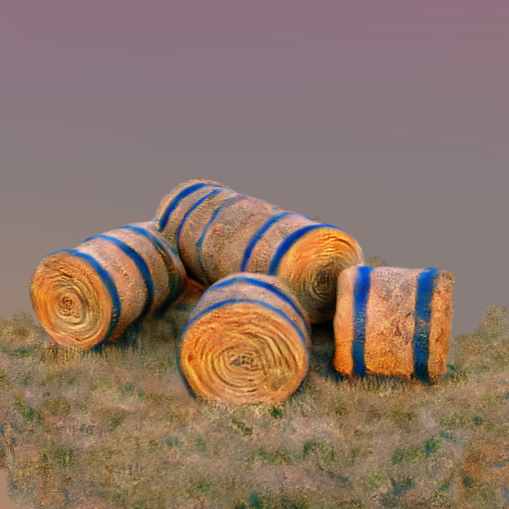} 
        \vspace{-0.1em}
        \\
        \multicolumn{7}{c}{{\prompts{\textcolor{red}{Several} large, solid, symmetrical hay bales, with a rough, golden texture, scattered across}}} \\
        \multicolumn{7}{c}{{\prompts{\textcolor{red}{a rural, open field}, with the setting sun casting long shadows}}}
        \\
        \begin{turn}{90} \,\,\,\,\,\,\,\,\,\small{MVDream~\cite{shi2023mvdream}} \end{turn} &
        \includegraphics[width=0.22\textwidth]{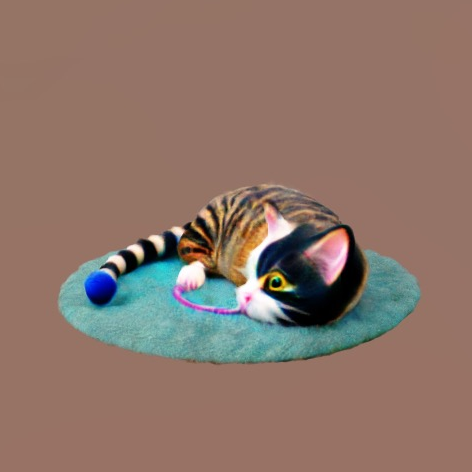} &
        \includegraphics[width=0.22\textwidth]{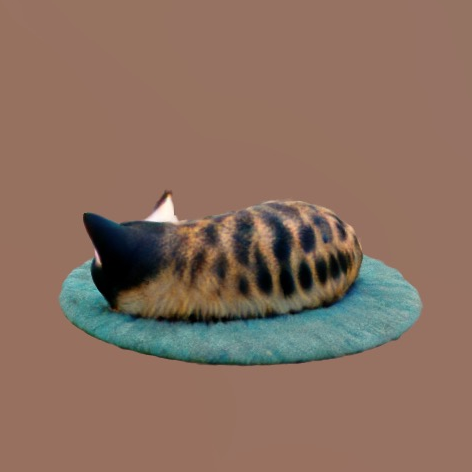} &
        \includegraphics[width=0.22\textwidth]{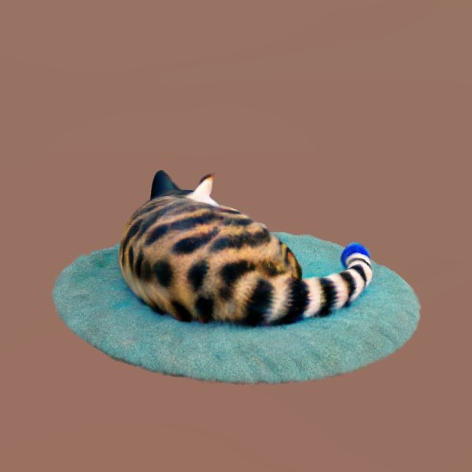} &
        \includegraphics[width=0.22\textwidth]{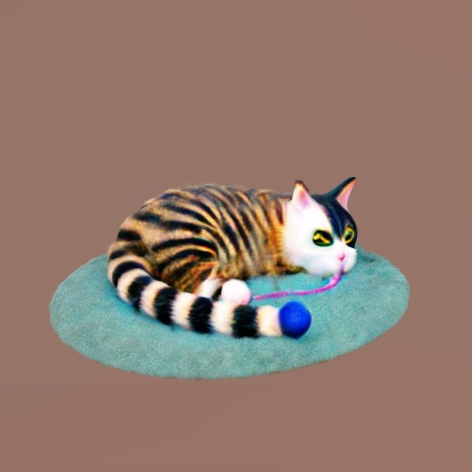} &
        \includegraphics[width=0.22\textwidth]{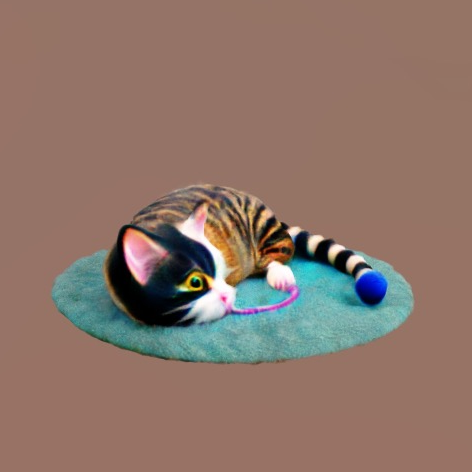} &
        \includegraphics[width=0.22\textwidth]{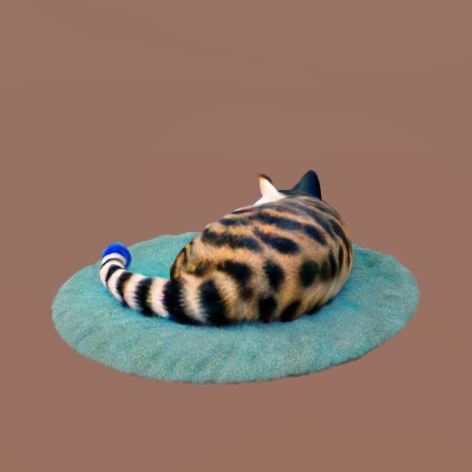} 
        \vspace{-0.1em}
        \\
        \begin{turn}{90} \,\,\,\,\,\,\,\,\small{\textbf{VLM3D (ours)}} \end{turn} &
        \includegraphics[width=0.22\textwidth]{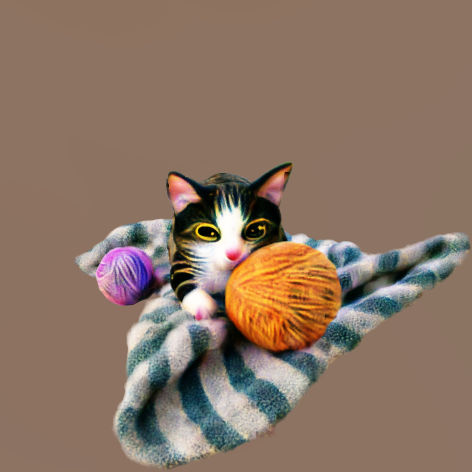} &
        \includegraphics[width=0.22\textwidth]{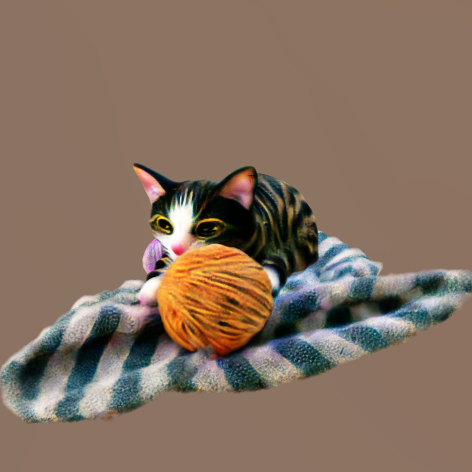} &
        \includegraphics[width=0.22\textwidth]{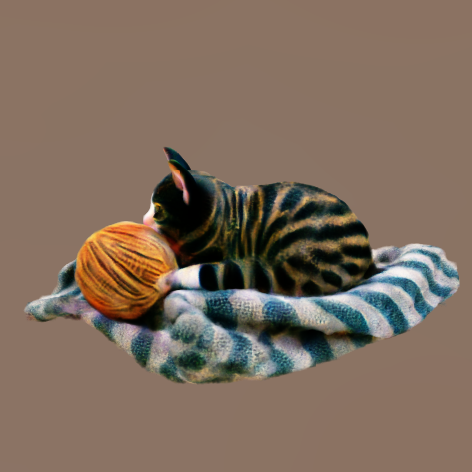} &
        \includegraphics[width=0.22\textwidth]{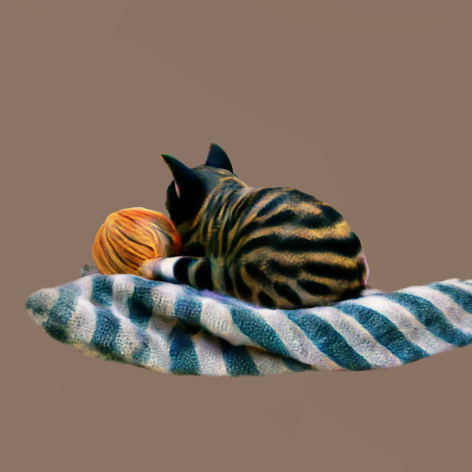} &
        \includegraphics[width=0.22\textwidth]{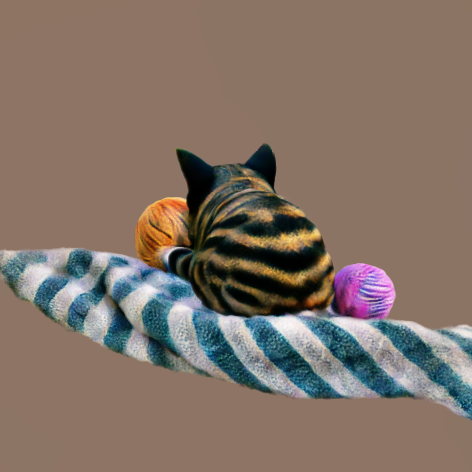} &
        \includegraphics[width=0.22\textwidth]{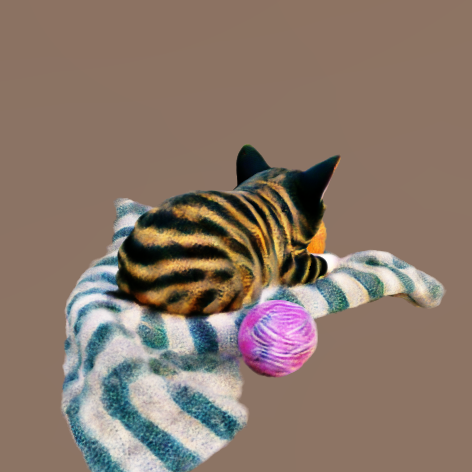} 
        \vspace{-0.1em}
        \\
        \multicolumn{7}{c}{{\prompts{A cat lying on its side \textcolor{red}{batting at a ball of yarn}}}}
        \\
        \begin{turn}{90} \,\,\,\,\,\,\,\,\,\small{MVDream~\cite{shi2023mvdream}} \end{turn} &
        \includegraphics[width=0.22\textwidth]{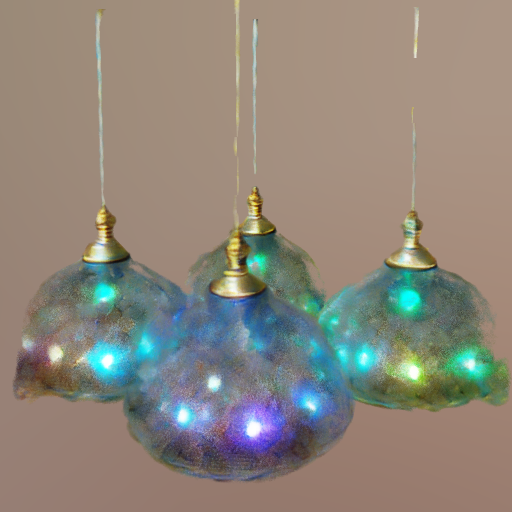} &
        \includegraphics[width=0.22\textwidth]{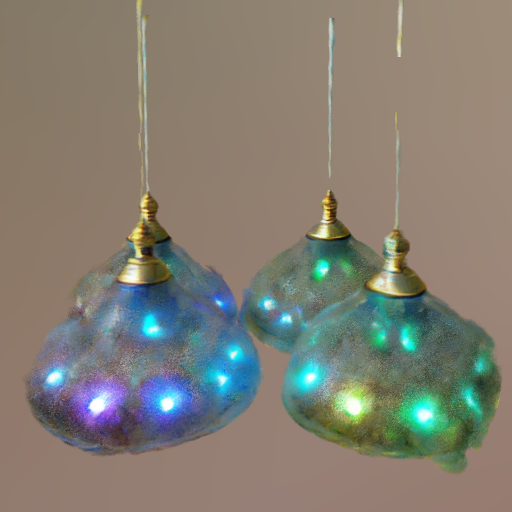} &
        \includegraphics[width=0.22\textwidth]{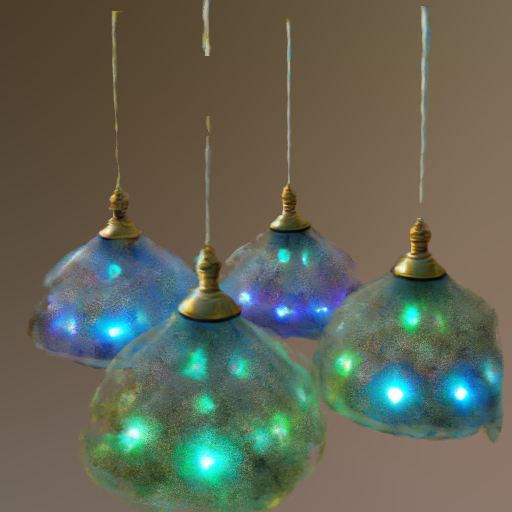} &
        \includegraphics[width=0.22\textwidth]{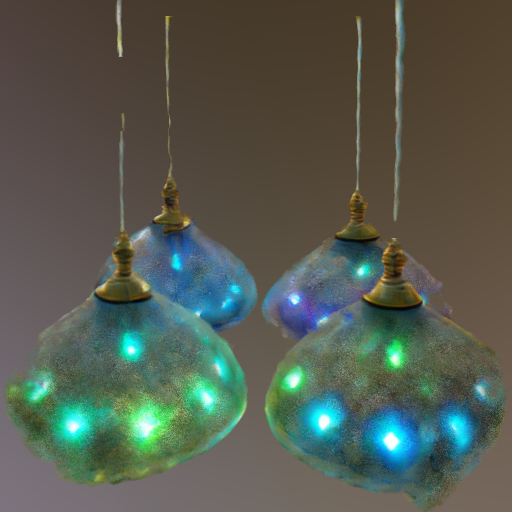} &
        \includegraphics[width=0.22\textwidth]{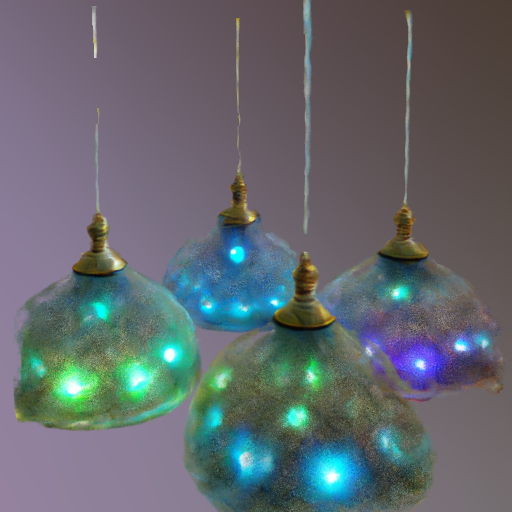} &
        \includegraphics[width=0.22\textwidth]{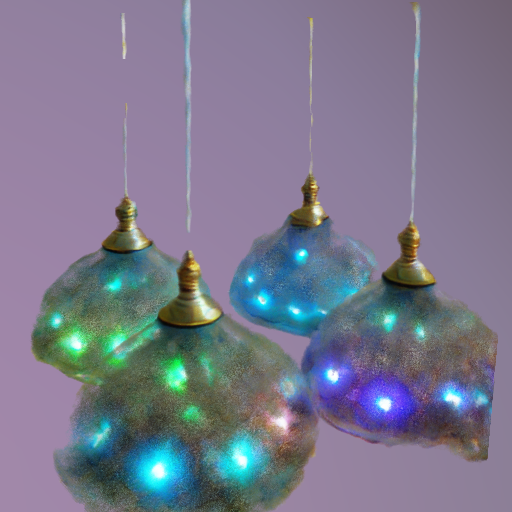} 
        \vspace{-0.1em}
        \\
        \begin{turn}{90} \,\,\,\,\,\,\,\,\small{\textbf{VLM3D (ours)}} \end{turn} &
        \includegraphics[width=0.22\textwidth]{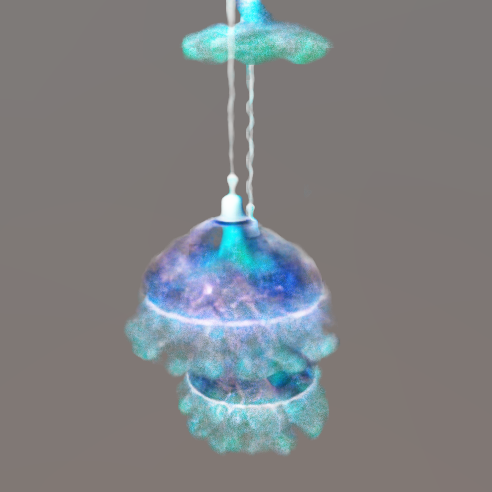} &
        \includegraphics[width=0.22\textwidth]{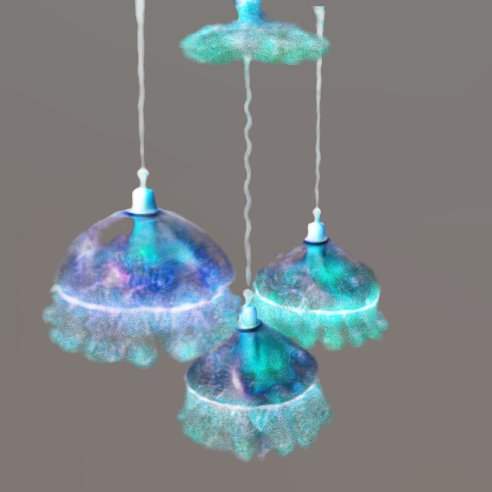} &
        \includegraphics[width=0.22\textwidth]{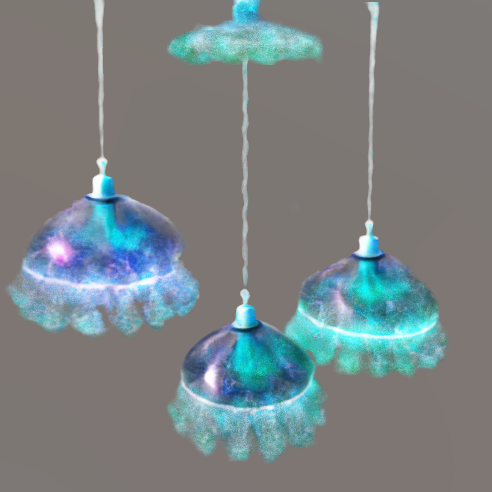} &
        \includegraphics[width=0.22\textwidth]{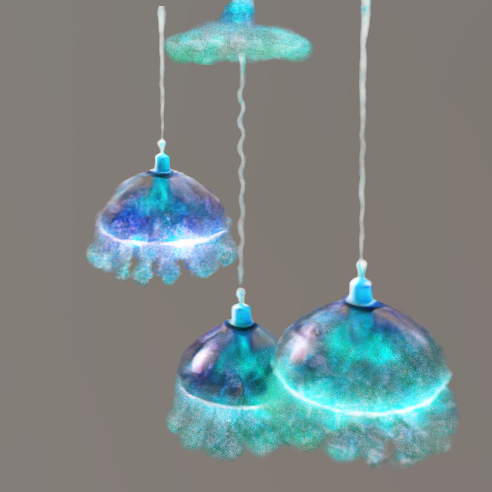} &
        \includegraphics[width=0.22\textwidth]{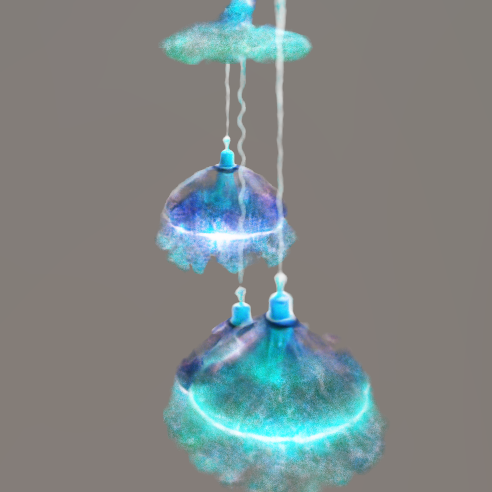} &
        \includegraphics[width=0.22\textwidth]{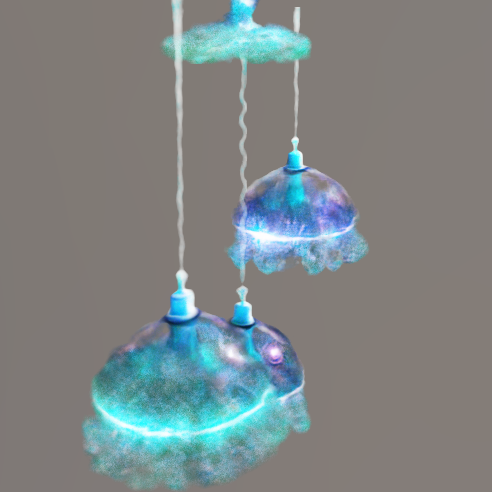} 
        \vspace{-0.1em}
        \\
        \multicolumn{7}{c}{{\prompts{An ensemble of \textcolor{red}{jellyfish-like} hanging lamps, pulsing with soft bioluminescence}}}
        \\
        \begin{turn}{90} \,\,\,\,\,\,\,\,\,\small{MVDream~\cite{shi2023mvdream}} \end{turn} &
        \includegraphics[width=0.22\textwidth]{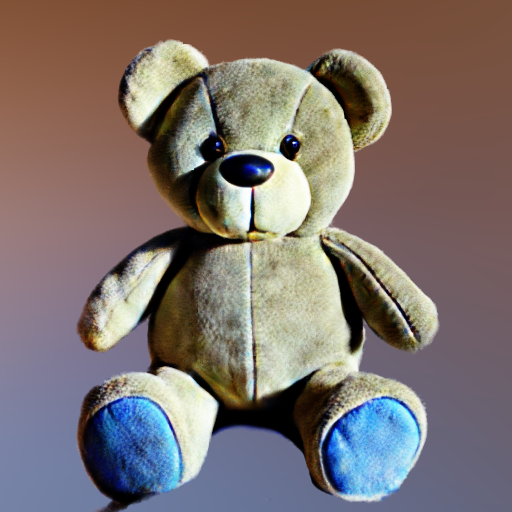} &
        \includegraphics[width=0.22\textwidth]{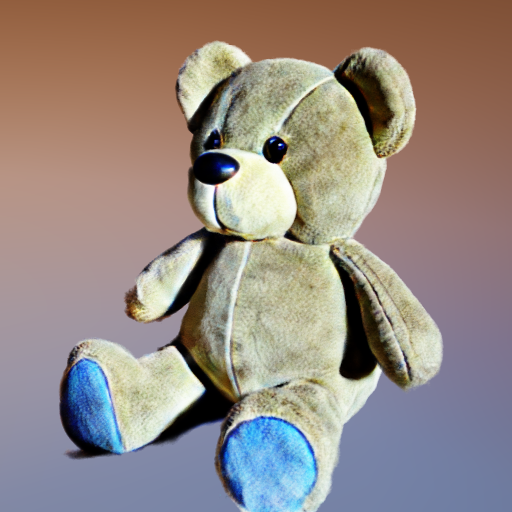} &
        \includegraphics[width=0.22\textwidth]{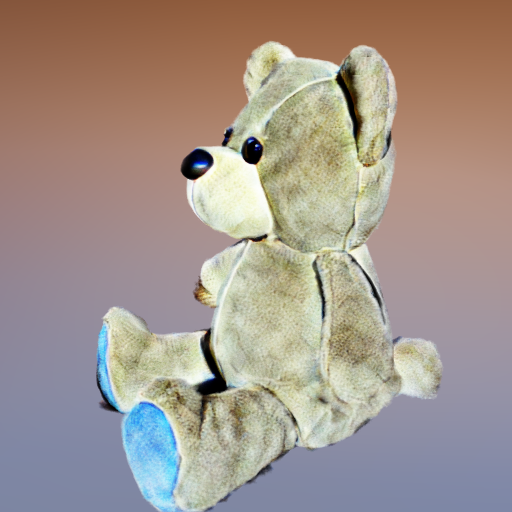} &
        \includegraphics[width=0.22\textwidth]{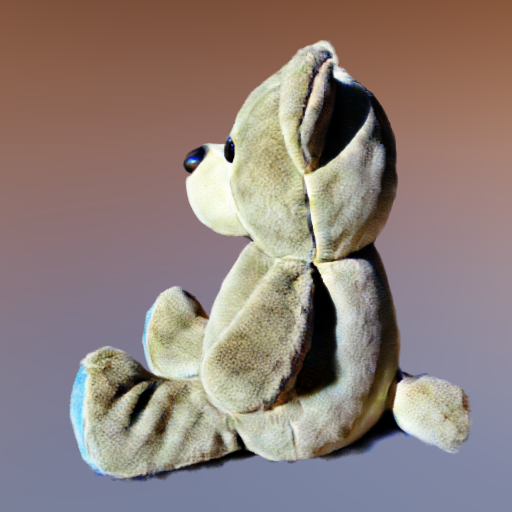} &
        \includegraphics[width=0.22\textwidth]{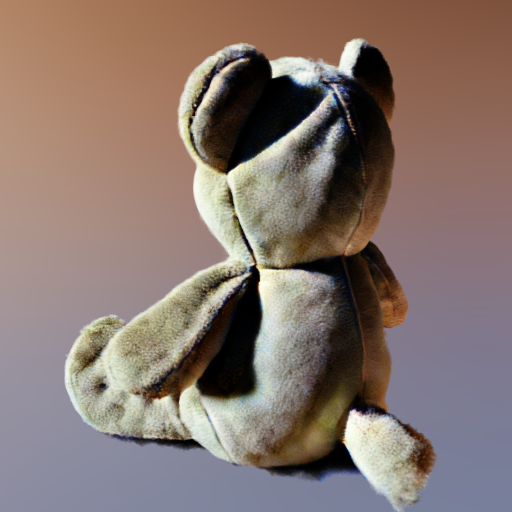} &
        \includegraphics[width=0.22\textwidth]{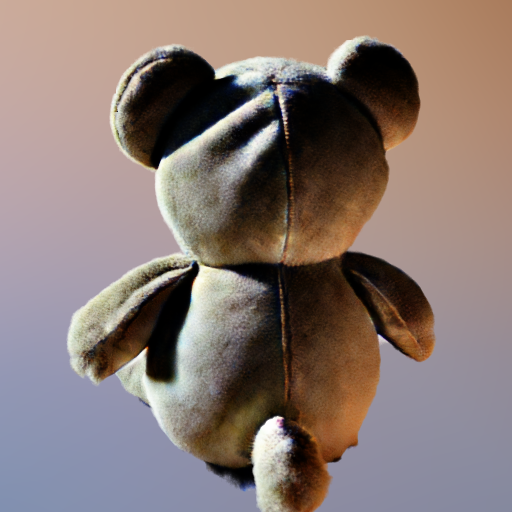}
        \vspace{-0.1em}
        \\
        \begin{turn}{90} \,\,\,\,\,\,\,\,\small{\textbf{VLM3D (ours)}} \end{turn} &
        \includegraphics[width=0.22\textwidth]{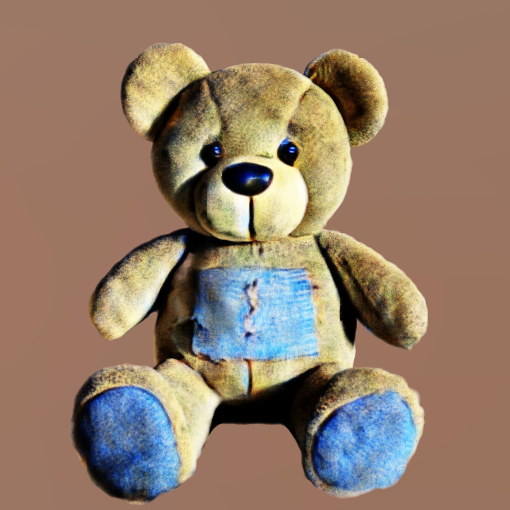} &
        \includegraphics[width=0.22\textwidth]{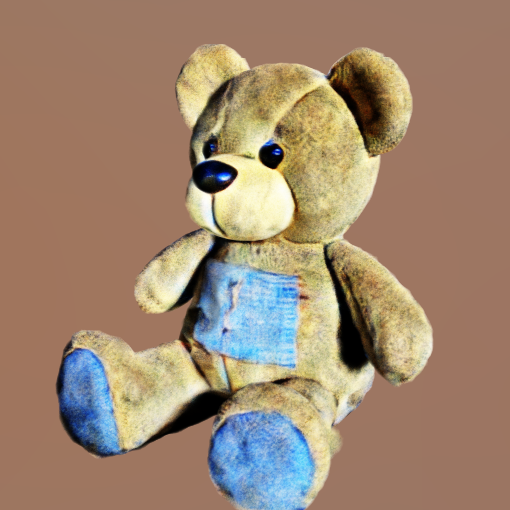} &
        \includegraphics[width=0.22\textwidth]{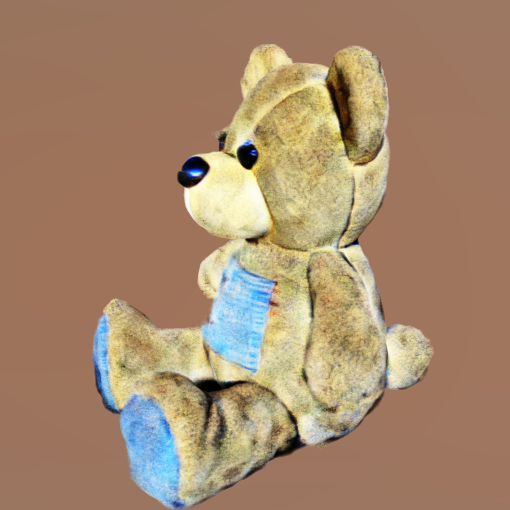} &
        \includegraphics[width=0.22\textwidth]{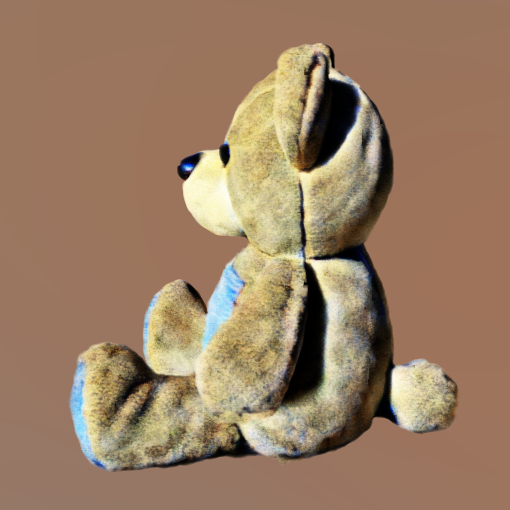} &
        \includegraphics[width=0.22\textwidth]{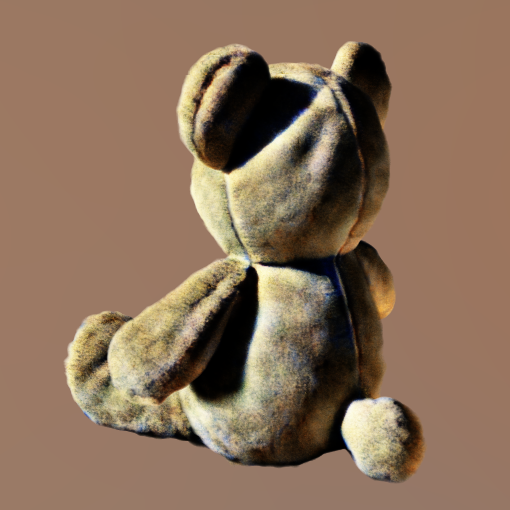} &
        \includegraphics[width=0.22\textwidth]{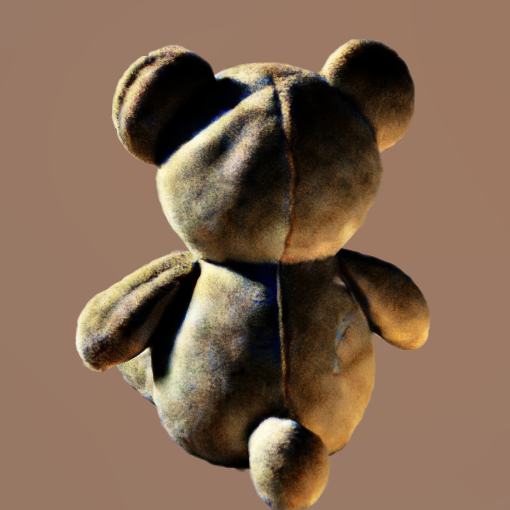}
        \vspace{-0.1em}
        \\
        \multicolumn{7}{c}{{\prompts{A plush teddy bear, sitting alone with \textcolor{red}{a slight tear} in its seam}}}
    \end{tabular}
    \end{tabular}}
    \vspace{-0.1em}
    \caption{
    \textbf{Additional results generated by our VLM3D.} 
    }
    \label{fig:appendix-additional-results1}
     \vspace{-0.3cm}
\end{figure}

\begin{figure}[t]
    \centering
    \setlength{\tabcolsep}{1pt}
    \setlength{\fboxrule}{1pt}
    \resizebox{0.9999\textwidth}{!}{
    \begin{tabular}{c}
    \begin{tabular}{ccccccc}
        \begin{turn}{90} \,\,\,\,\,\,\,\,\,{MVDream~\cite{shi2023mvdream}} \end{turn} &
        \includegraphics[width=0.22\textwidth]{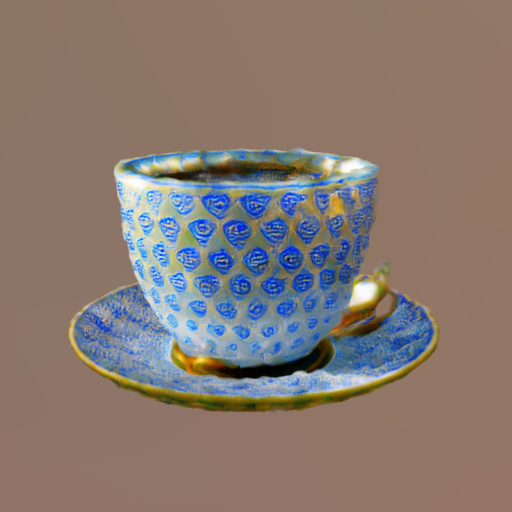} &
        \includegraphics[width=0.22\textwidth]{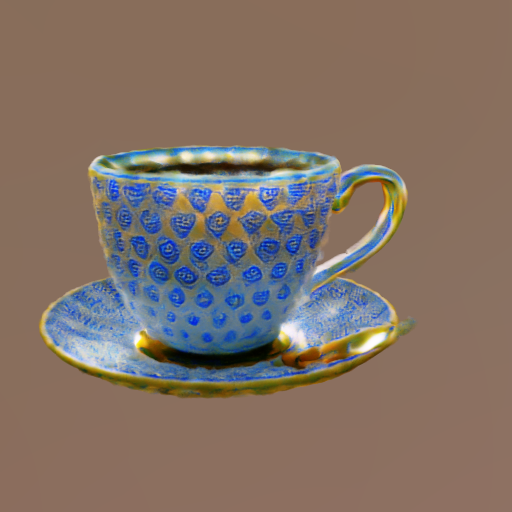} &
        \includegraphics[width=0.22\textwidth]{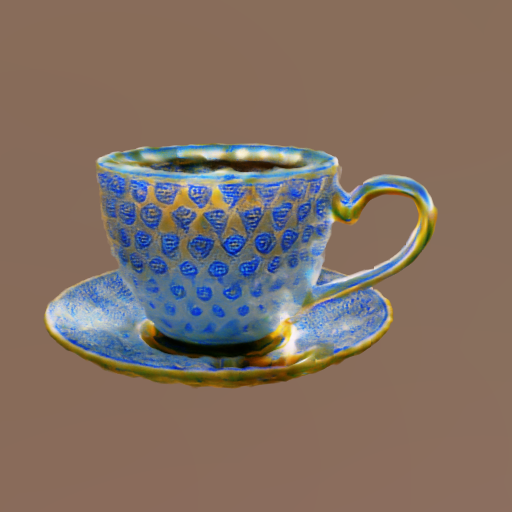} &
        \includegraphics[width=0.22\textwidth]{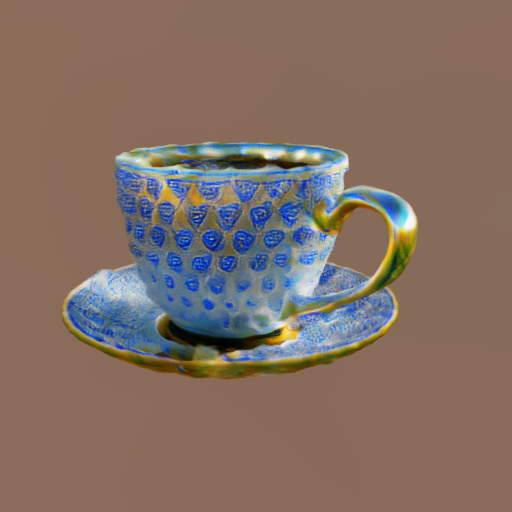} &
        \includegraphics[width=0.22\textwidth]{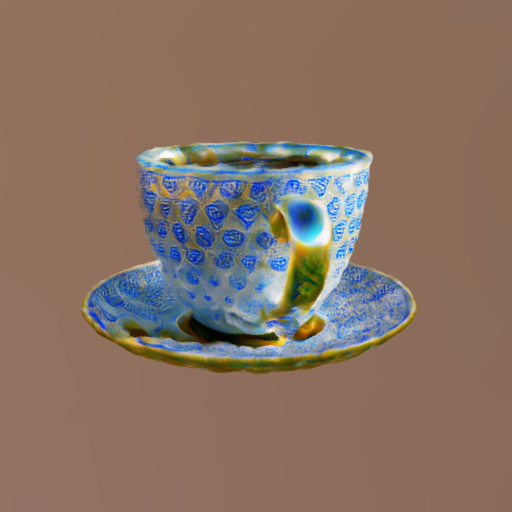} &
        \includegraphics[width=0.22\textwidth]{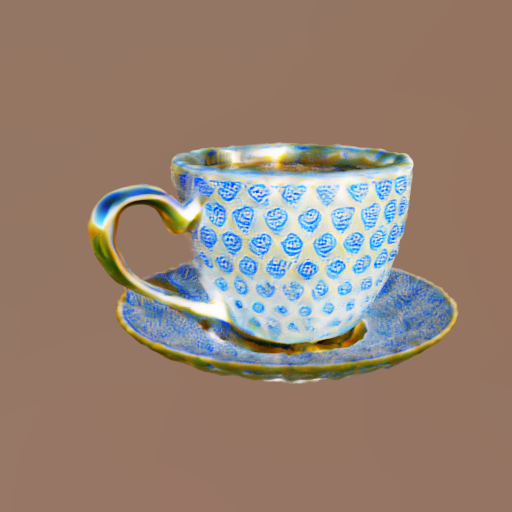} 
        \vspace{-0.1em}
        \\
        \begin{turn}{90} \,\,\,\,\,\,\,\,{\textbf{VLM3D (ours)}} \end{turn} &
        \includegraphics[width=0.22\textwidth]{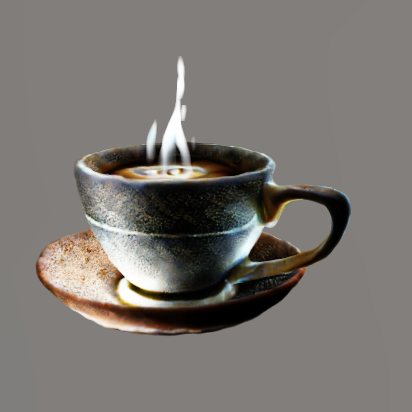} &
        \includegraphics[width=0.22\textwidth]{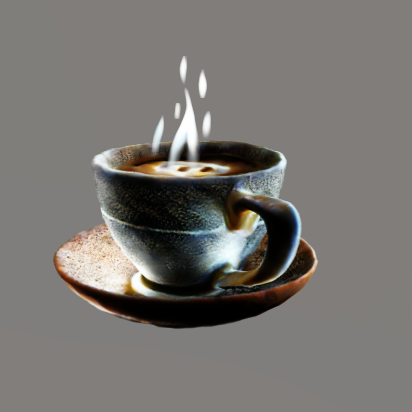} &
        \includegraphics[width=0.22\textwidth]{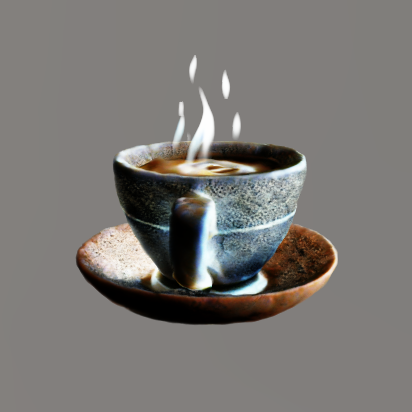} &
        \includegraphics[width=0.22\textwidth]{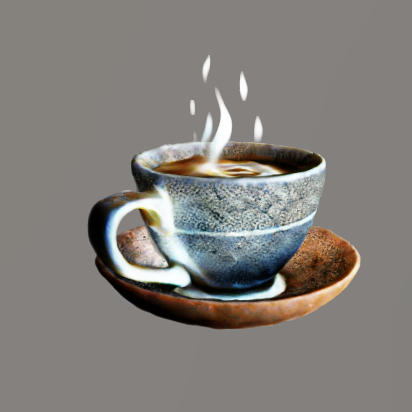} &
        \includegraphics[width=0.22\textwidth]{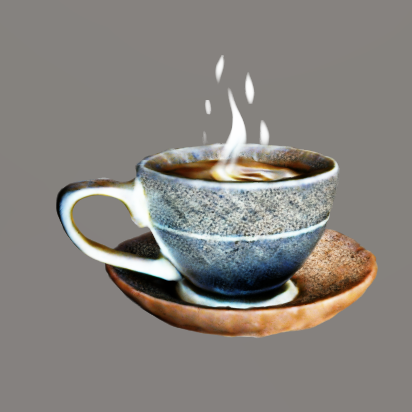} &
        \includegraphics[width=0.22\textwidth]{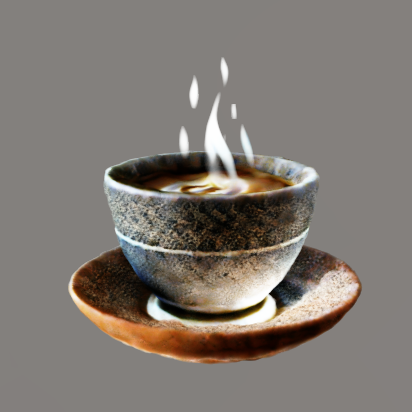} 
        \vspace{-0.1em}
        \\
        \multicolumn{7}{c}{{{A mug filled with \textcolor{red}{steaming} coffee}}}
        \\
        \begin{turn}{90} \,\,\,\,\,\,\,\,\,{MVDream~\cite{shi2023mvdream}} \end{turn} &
        \includegraphics[width=0.22\textwidth]{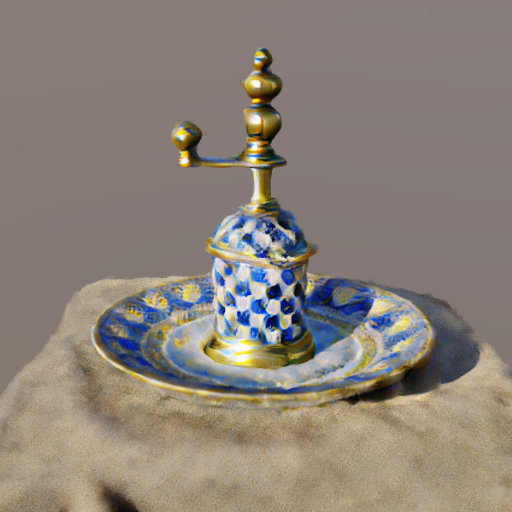} &
        \includegraphics[width=0.22\textwidth]{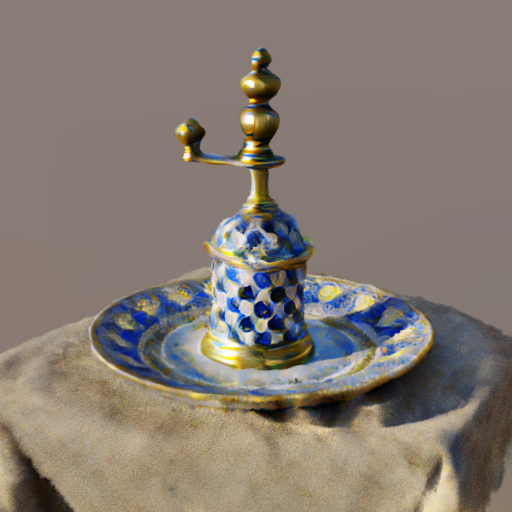} &
        \includegraphics[width=0.22\textwidth]{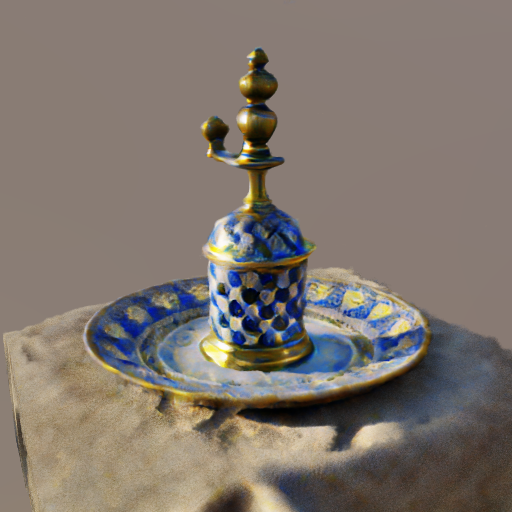} &
        \includegraphics[width=0.22\textwidth]{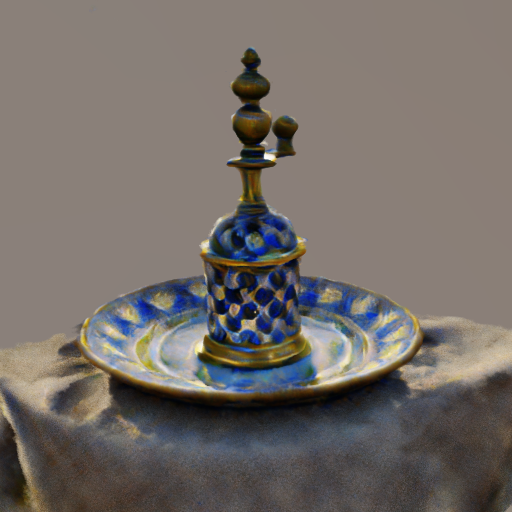} &
        \includegraphics[width=0.22\textwidth]{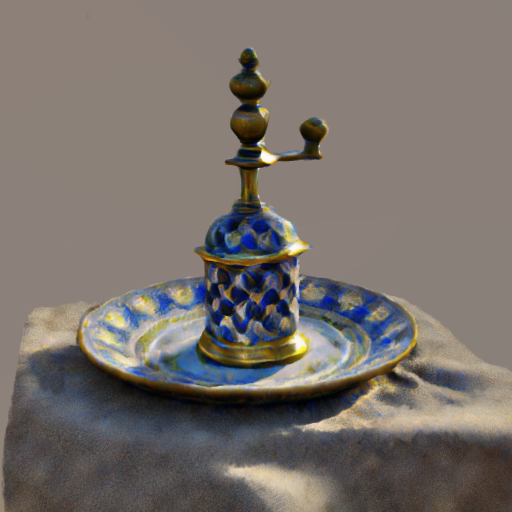} &
        \includegraphics[width=0.22\textwidth]{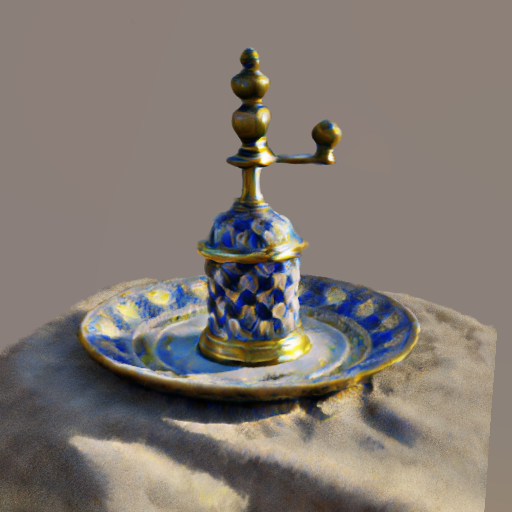} 
        \vspace{-0.1em}
        \\
        \begin{turn}{90} \,\,\,\,\,\,\,\,{\textbf{VLM3D (ours)}} \end{turn} &
        \includegraphics[width=0.22\textwidth]{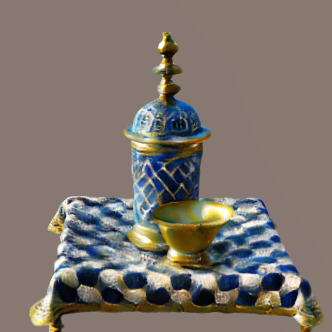} &
        \includegraphics[width=0.22\textwidth]{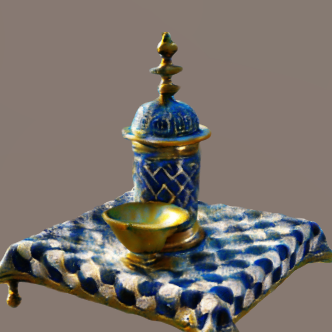} &
        \includegraphics[width=0.22\textwidth]{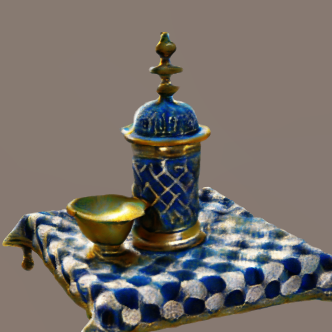} &
        \includegraphics[width=0.22\textwidth]{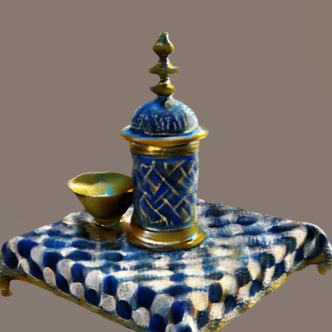} &
        \includegraphics[width=0.22\textwidth]{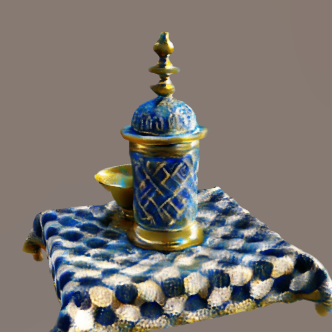} &
        \includegraphics[width=0.22\textwidth]{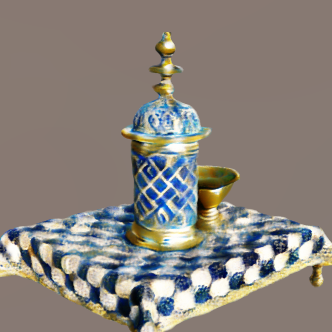} 
        \vspace{-0.1em}
        \\
        \multicolumn{7}{c}{{\prompts{A compact cylindrical vintage pepper mill, with a polished, ornate brass body, slightly worn from use,}}} \\
        \multicolumn{7}{c}{{\prompts{placed \textcolor{red}{beside} a porcelain plate on a \textcolor{red}{checkered tablecloth}}}}
        \\
        \begin{turn}{90} \,\,\,\,\,\,\,\,\,{MVDream~\cite{shi2023mvdream}} \end{turn} &
        \includegraphics[width=0.22\textwidth]{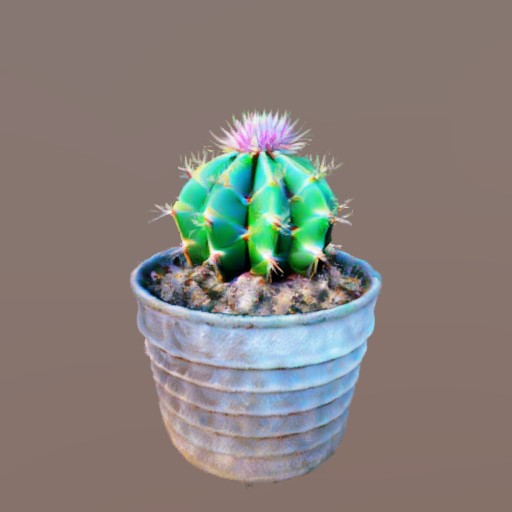} &
        \includegraphics[width=0.22\textwidth]{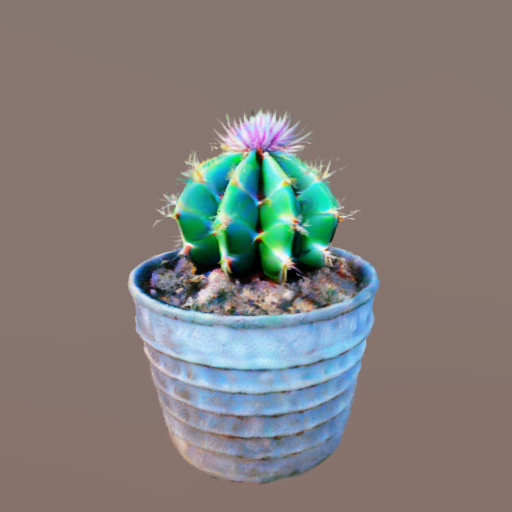} &
        \includegraphics[width=0.22\textwidth]{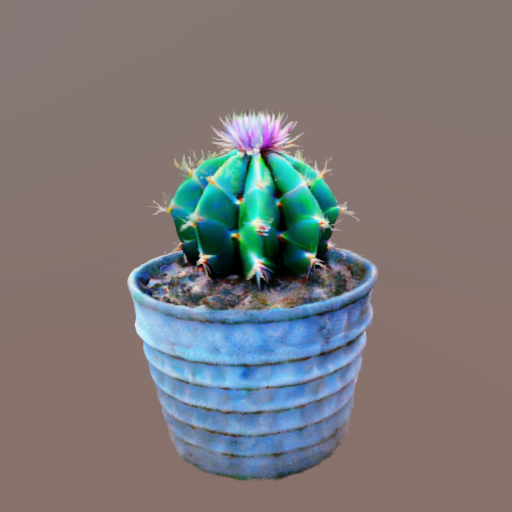} &
        \includegraphics[width=0.22\textwidth]{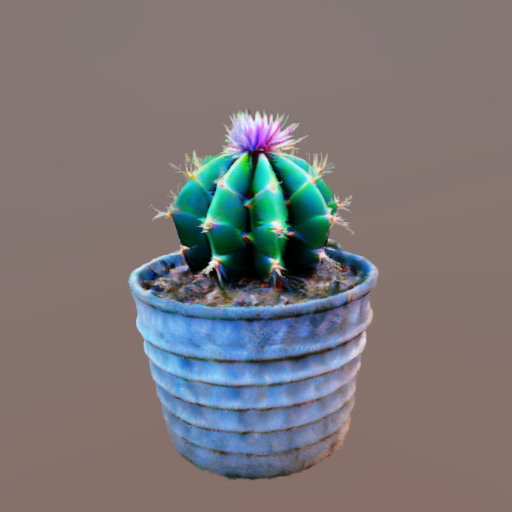} &
        \includegraphics[width=0.22\textwidth]{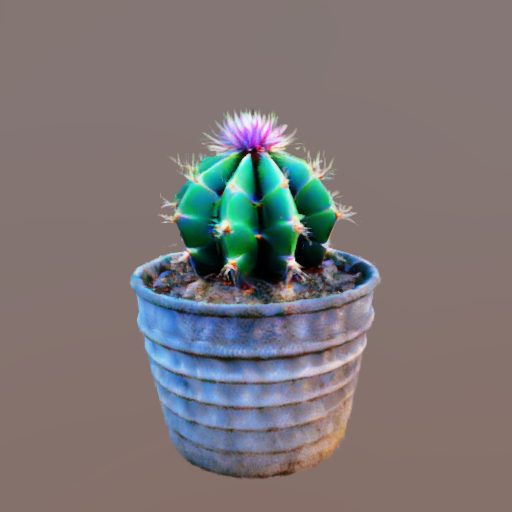} &
        \includegraphics[width=0.22\textwidth]{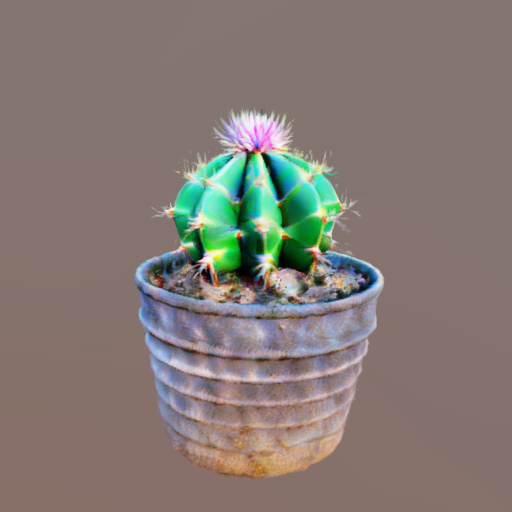} 
        \vspace{-0.1em}
        \\
        \begin{turn}{90} \,\,\,\,\,\,\,\,{\textbf{VLM3D (ours)}} \end{turn} &
        \includegraphics[width=0.22\textwidth]{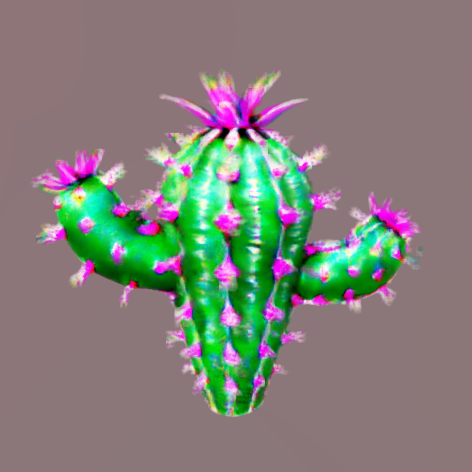} &
        \includegraphics[width=0.22\textwidth]{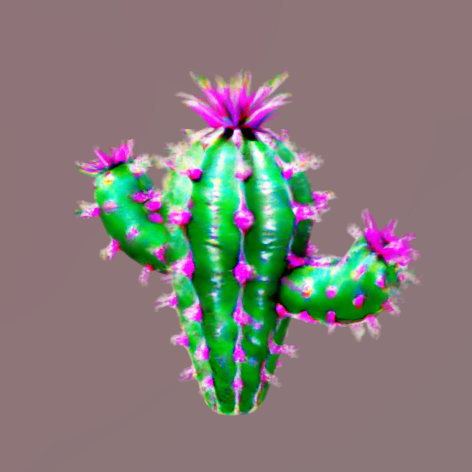} &
        \includegraphics[width=0.22\textwidth]{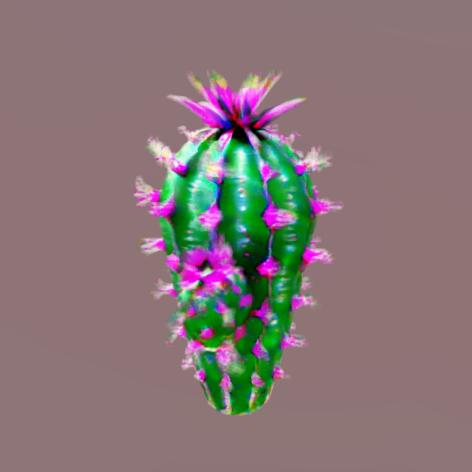} &
        \includegraphics[width=0.22\textwidth]{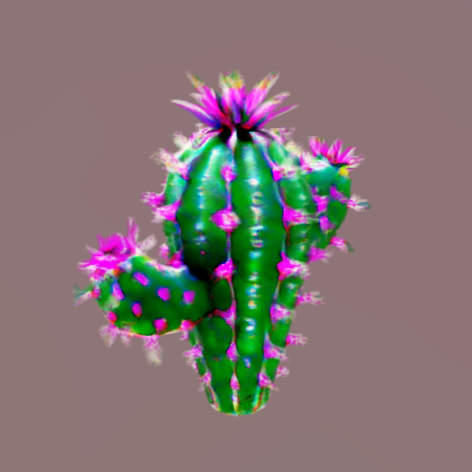} &
        \includegraphics[width=0.22\textwidth]{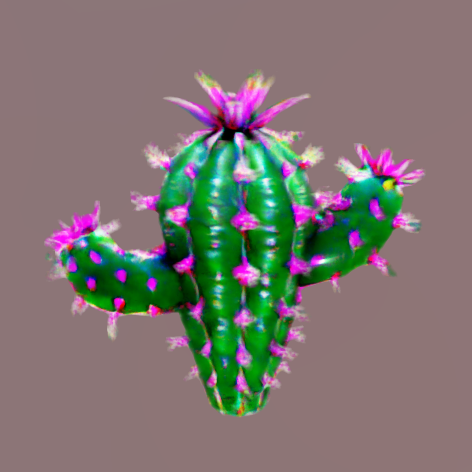} &
        \includegraphics[width=0.22\textwidth]{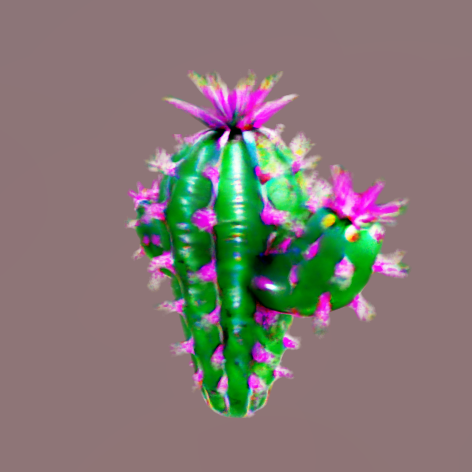} 
        \vspace{-0.1em}
        \\
        \multicolumn{7}{c}{{\prompts{A \textcolor{red}{rubbery} cactus}}}
        \\
        \begin{turn}{90} \,\,\,\,\,\,\,\,\,{MVDream~\cite{shi2023mvdream}} \end{turn} &
        \includegraphics[width=0.22\textwidth]{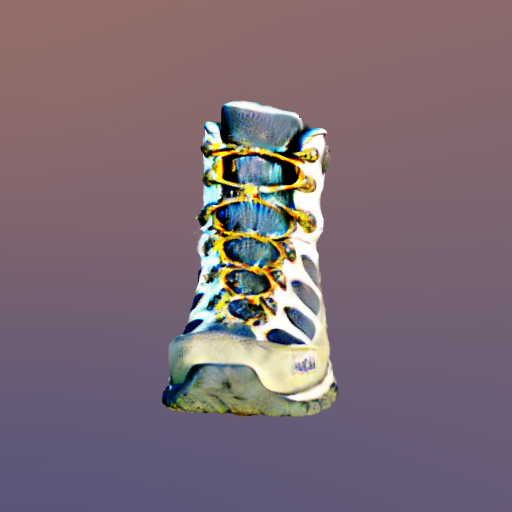} &
        \includegraphics[width=0.22\textwidth]{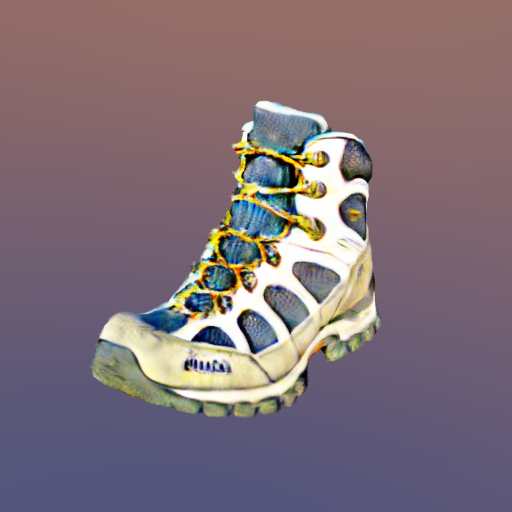} &
        \includegraphics[width=0.22\textwidth]{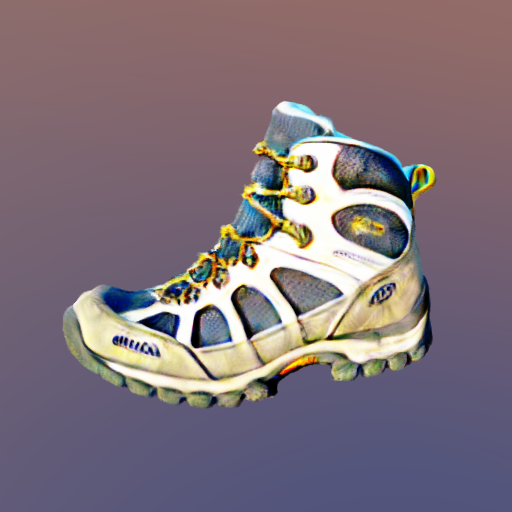} &
        \includegraphics[width=0.22\textwidth]{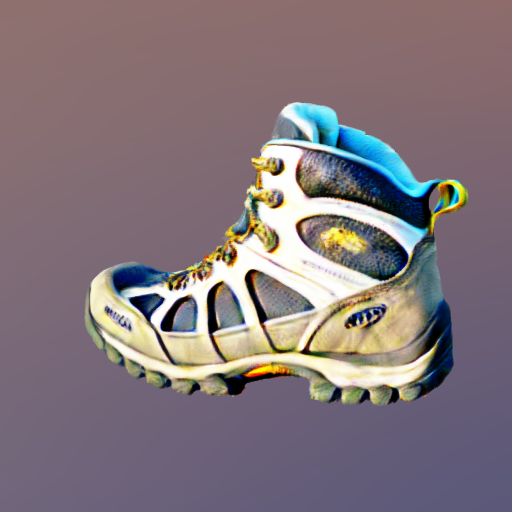} &
        \includegraphics[width=0.22\textwidth]{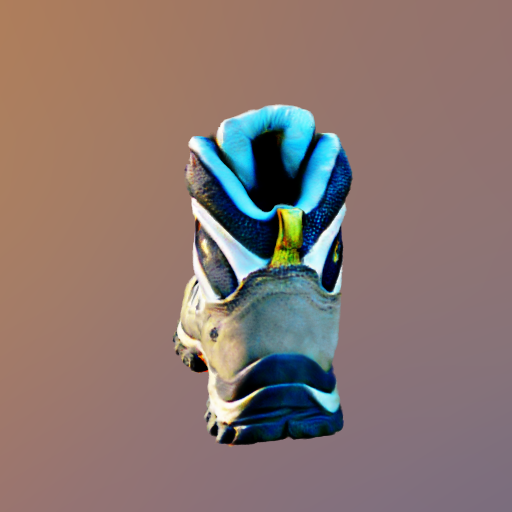} &
        \includegraphics[width=0.22\textwidth]{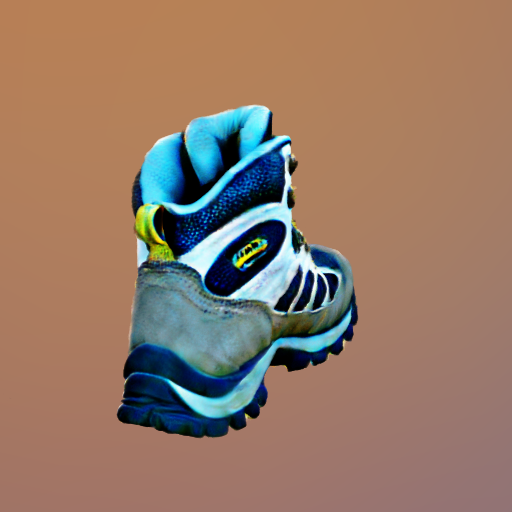} 
        \vspace{-0.1em}
        \\
        \begin{turn}{90} \,\,\,\,\,\,\,\,{\textbf{VLM3D (ours)}} \end{turn} &
        \includegraphics[width=0.22\textwidth]{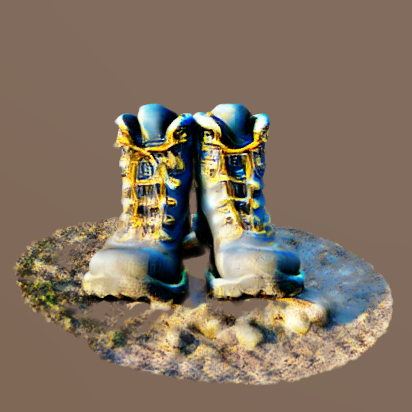} &
        \includegraphics[width=0.22\textwidth]{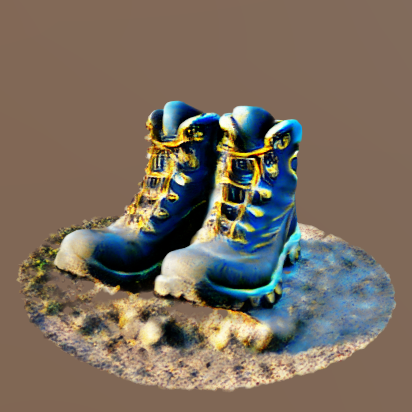} &
        \includegraphics[width=0.22\textwidth]{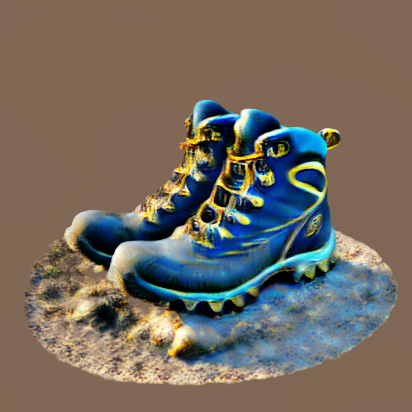} &
        \includegraphics[width=0.22\textwidth]{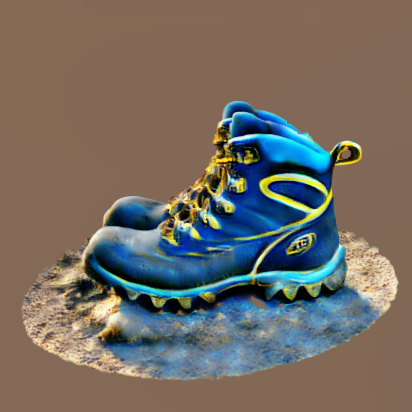} &
        \includegraphics[width=0.22\textwidth]{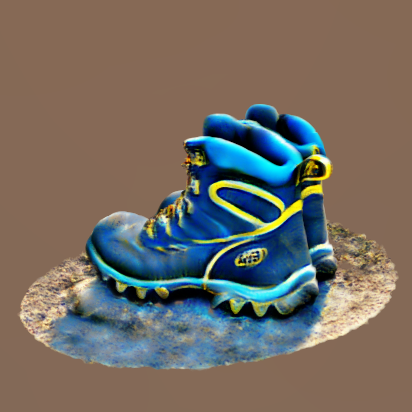} &
        \includegraphics[width=0.22\textwidth]{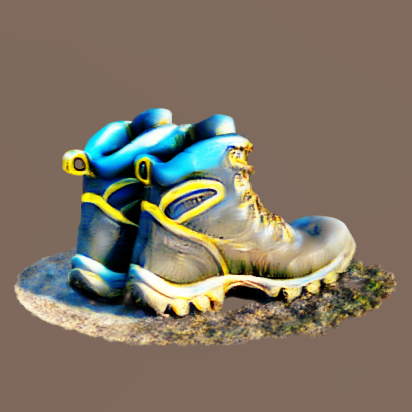} 
        \vspace{-0.1em}
        \\
        \multicolumn{7}{c}{{\prompts{\textcolor{red}{A pair of} hiking boots caked with mud at the doorstep of a cabin}}}
    \end{tabular}
    \end{tabular}}
    \vspace{-0.1em}
    \caption{
    \textbf{Additional results generated by our VLM3D.} 
    }
    \label{fig:appendix-additional-results2}
     \vspace{-0.3cm}
\end{figure}

More visual examples are provided in Figure~\ref{fig:appendix-additional-results1} and Figure~\ref{fig:appendix-additional-results2}. These results show that our method creates high-quality and human-preferred 3D assets. 
Additionally, these assets more accurately match the given text descriptions and display finer details in their shape and surface textures.

\newpage

\newpage

\end{document}